%% file: main.tex
\pdfoutput=1
\documentclass[11pt]{article}

\usepackage[final]{acl}

\usepackage{times}
\usepackage{latexsym}

\usepackage[T1]{fontenc}
\usepackage[utf8]{inputenc}

\usepackage{microtype}

\usepackage{inconsolata}

\usepackage{graphicx}
\usepackage{subcaption} % For subfigures
\usepackage{xspace}
\usepackage{booktabs}
\usepackage{multirow}
\usepackage{amsmath}
\usepackage{amsfonts}

\usepackage[capitalize,noabbrev,nameinlink]{cleveref}

\usepackage{todonotes}
\usepackage{soul}
\usepackage[normalem]{ulem}
\usepackage{bm}
\usepackage{dsfont}
\usepackage{mathtools}
\usepackage{pifont}

\input{preamble}

\hypersetup{colorlinks, citecolor=chart_blue, linkcolor=chart_blue}
\usepackage{booktabs}
\usepackage{tabularx}

\usepackage{tikz}
\usepackage{wrapfig}

\usepackage{tcolorbox}
\usepackage{float}
\usepackage[framemethod=tikz]{mdframed}

\usepackage{amsthm}
\newtheorem{observation}{Observation}

\makeatletter

\title{In-Place Feedback: Reliable Refinement for Multi-Turn \\ Expert-LLM Collaboration}

\author{
 \textbf{Youngbin Choi\textsuperscript{1,*}},
 \textbf{Minjong Lee\textsuperscript{2,*}},
\textbf{Saemi Moon\textsuperscript{2}},
\textbf{Seunghyuk Cho\textsuperscript{1}},\\
 \textbf{Chaehyeon Chung\textsuperscript{2}},
 \textbf{MoonJeong Park\textsuperscript{1}}, 
 \textbf{Dongwoo Kim\textsuperscript{1,2,\textdagger}}
\\
 \textsuperscript{1}Graduate School of Artificial Intelligence, POSTECH,
\\
 \textsuperscript{2}Department of Computer Science and Engineering, POSTECH, 
\\
 \small{
    \{choi.youngbin, minjong.lee, saemi, shhj1998, chchung, mjeongp, dongwoo.kim\}@postech.ac.kr
 }
}

\begin{document}
\maketitle
\begingroup
\renewcommand\thefootnote{}\footnotetext{{$^{*}$}Equal contribution.}
\renewcommand\thefootnote{}\footnotetext{$^\text{\textdagger}$Corresponding author.}

\endgroup

%%%%%%%%%%% prompt box %%%%%%%%%%%%%%%%%%
\newtcolorbox{prompt}[1][]{
  colback=chart_blue!5!white,
  colframe=chart_blue,
  floatplacement=floating,
  title=\centering \textsf{\small #1}
}
\newtcolorbox{inplace}[1][]{
  colback=chart_grey!5!white,
  colframe=chart_grey,
  floatplacement=floating,
  title=\centering \textsf{\small #1}
}
\newtcolorbox{failure}[1][]{
  colback=chart_red!5!white,
  colframe=chart_red,
  floatplacement=floating,
  title=\centering \textsf{\small #1}
}

\gdef\Sepline{%
  \par\noindent\makebox[\linewidth][l]{%
  \hspace*{-\mdflength{innerleftmargin}}%
   \tikz\draw[thick,dashed,gray!60] (0,0) --%
        (\textwidth+\the\mdflength{innerleftmargin}+\the\mdflength{innerrightmargin},0);
  }\par\nobreak}

%%%%%%%%%%%%%%%%%%%%%%%%%%%%%%%%%%%%%%%%

\input{sections/0_abstract}
\input{sections/1_introduction}

\input{sections/6_related_work}

\input{sections/2_inplace_feedback}

\input{sections/3_experimental_results}

\input{sections/4_analysis}

\input{sections/5_user_study}

\input{sections/7_conclusion}

% Bibliography entries for the entire Anthology, followed by custom entries
%\bibliography{anthology,custom}
% Custom bibliography entries only
\bibliography{custom}

\clearpage
\appendix

\section*{Appendix}
\input{sections/a_appendix}

\end{document}

%% file: preamble.tex
\definecolor{BrickRed}{rgb}{0.6,0,0}
\definecolor{RoyalBlue}{rgb}{0,0,0.8}
\definecolor{Tdgreen}{rgb}{0,0.4,0.7}
\definecolor{cadmiumgreen}{rgb}{0.0, 0.42, 0.24}
\definecolor{mybrown}{RGB}{128,64,0}
\definecolor{red}{RGB}{235, 37, 64}
\definecolor{blue}{RGB}{64, 123, 211}
\definecolor{chart_blue}{HTML}{0C7BDC}
\definecolor{chart_grey}{HTML}{525252}
\definecolor{chart_red}{rgb}{0.87, 0.36, 0.51}

\makeatletter 
\DeclareRobustCommand\onedot{\futurelet\@let@token\@onedot}
\def\@onedot{\ifx\@let@token.\else.\null\fi\xspace}

\def\eg{\emph{e.g}\onedot}

\makeatother

%% file: sections/0_abstract.tex
\begin{abstract}

LLM-generated drafts often contain subtle factual or logical errors, yet prior work shows that models struggle to reliably integrate multi-turn feedback aimed at fixing them. We propose \emph{in-place feedback}, an interaction paradigm in which the user directly edits the model's previous response and the model continues generation from the edited context. In-place feedback consistently outperforms standard multi-turn feedback across five reasoning-intensive benchmarks while requiring fewer tokens, and our fine-grained analysis shows that it applies corrections more reliably and propagates them to subsequent reasoning. A user study with domain experts refining LLM-generated summaries corroborates these findings: participants report higher final-output satisfaction and substantially lower fatigue with in-place feedback, and a mixed strategy combining in-place and multi-turn feedback improves every dimension we measure. These results suggest that editing errors in place is a more reliable way to incorporate expert corrections than multi-turn feedback, while combining the two acts complementarily to form a more effective collaboration than either mechanism alone.
%These results suggest that editing errors directly is a more effective paradigm for expert-LLM collaboration. %The code is available at \href{https://github.com/ml-postech/In-Place-Feedback}{https://github.com/ml-postech/In-Place-Feedback}.
\end{abstract}

%% file: sections/1_introduction.tex
\begin{figure*}[t]
    \centering
    \includegraphics[width=\linewidth]{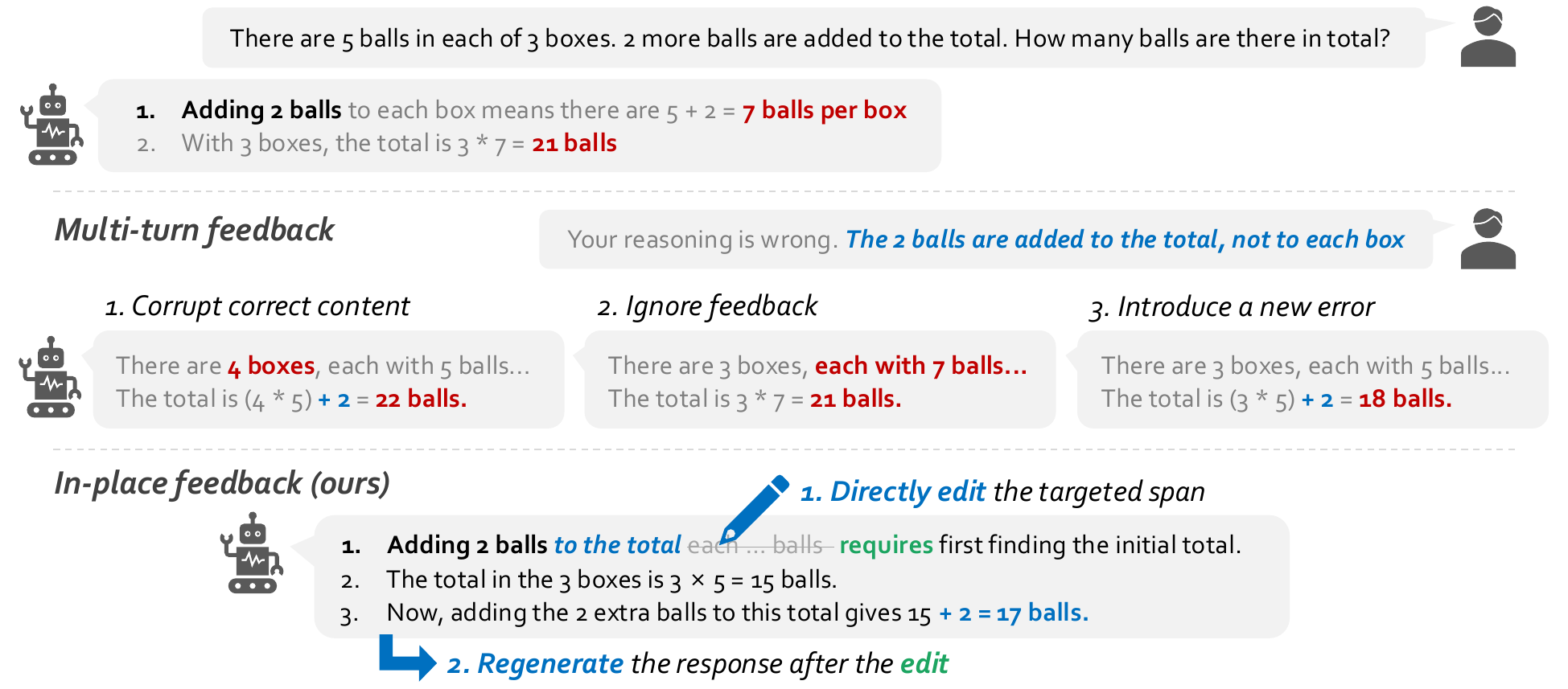} 
    % \vskip 0.5 em
    \caption{Illustration of common failure cases in multi-turn conversational feedback and in-place feedback. After in-place feedback, the LLM continues generation from the green word ``\texttt{requires}''.}
    % \vskip 1.em
    \label{fig:in_place_failure_case}
\end{figure*}

\section{Introduction}
\label{sec:intro}

Large language models (LLMs) are increasingly used as collaborative assistants in expert workflows such as software engineering, legal drafting, and clinical documentation~\citep{lee2022coauthor, wang2025training, zhou2024archer}. Since expert-level tasks involve complex reasoning, mistakes can cascade and disrupt the workflow, so experts typically review outputs and provide fine-grained feedback that the model uses to produce a revised response~\citep{he2025let, chi2026editbench, Lam2023ApplyingLL, satheakeerthy2025llm}. This refinement cycle repeats until the output meets the expert's requirements, forming an expert-in-the-loop workflow.

The productivity of this workflow hinges on whether the model reliably incorporates feedback~\citep{chidambaram2024socratic, vaccaro2024combinations}. Prior work shows that LLMs frequently fail to do so in multi-turn interactions~\citep{wang2024mint, deshpande2025multichallenge, jiang2025feedback, laban2025llms}, and the problem compounds as more turns accumulate: the model must integrate the latest correction while still preserving earlier feedback and leaving unrelated content untouched. When this integration fails, experts must re-check the output and re-state corrections, undermining productivity.

We hypothesize that these failures stem from the accumulation of erroneous context across turns, which obscures what the model should discard and what it should retain. To mitigate this context pollution, we propose \emph{in-place feedback}: instead of appending a new feedback message, the user directly edits the erroneous span within the LLM's previous response and prunes the downstream portion that depends on it, and the LLM regenerates only the remainder from the corrected context. By design, in-place feedback both prevents incorrect context from accumulating and keeps the effective input shorter, reducing token usage.

We validate in-place feedback in three complementary ways. 
\begin{itemize}
    \item First, across five reasoning-intensive benchmarks spanning mathematics, science, coding, and logic (MATH-hard~\citep{hendrycks2021measuring}, MMLU-pro~\citep{wang2024mmlu}, GPQA~\citep{rein2024gpqa}, LiveCodeBench~\citep{jain2025livecodebench}, and ZebraLogic~\citep{lin2025zebralogic}), LLMs follow in-place feedback more reliably than multi-turn feedback, with gains in both performance and token efficiency that widen across turns. 
    \item Second, we utilize the step-by-step reasoning framework of ZebraLogic to conduct a fine-grained analysis of feedback integration. Our controlled experiments reveal how standard multi-turn feedback often corrupts prior reasoning or fails to incorporate new corrections. In-place feedback mitigates these failures by regenerating directly from the context corrected by the expert, which reduces interference from stale reasoning and robustly propagates corrections.
    % \item Third, a user study with domain experts refining LLM-generated summaries corroborates these findings, where participants report higher satisfaction and lower fatigue with in-place feedback, and a mixed strategy that combines both paradigms scores highest on every measured dimension, suggesting that in-place feedback is a sensible default for expert-LLM interfaces while exposing both paradigms lets experts choose adaptively.
    \item Third, to validate these empirical gains in a practical setting, we conduct a user study involving domain experts editing LLM-generated summaries. The participants report significantly reduced fatigue and greater satisfaction when utilizing in-place feedback. Furthermore, a hybrid workflow integrating both interaction styles achieves the highest ratings across all evaluated dimensions.
\end{itemize}

%% file: sections/6_related_work.tex
\begin{figure*}[t]
  \centering    
    \includegraphics[width=\linewidth]{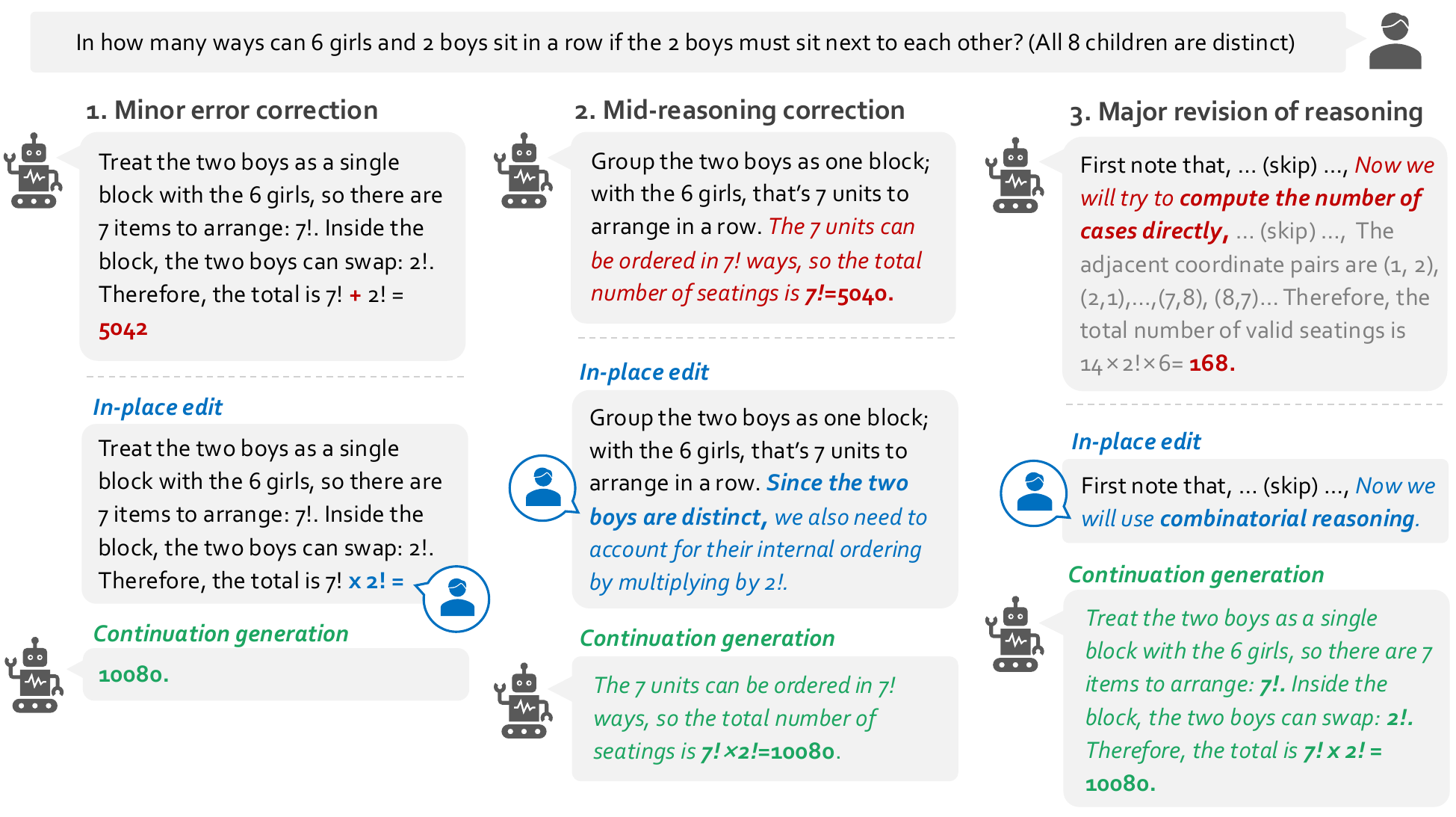}
    % \vskip -0.8 em
    \caption{Representative examples of in-place feedback on a toy problem. Red marks incorrect reasoning in the original response, blue indicates user corrections with in-place feedback, and green shows the subsequent reasoning based on the corrected context. Additional examples are provided in \cref{app.in_place}.}
    \label{fig:common_cases}
    % \vskip -1.3 em
\end{figure*}

\section{Related Work}
\label{sec:related_work}

\paragraph{Multi-turn interaction with LLMs.} Two complementary directions improve how LLMs behave across turns. Architectural approaches insert clarifying questions when user input is ambiguous~\citep{zhang2023clarify,  zamani2020generating, aliannejadi2019asking}. Training-based approaches fine-tune models to better follow user intent across turns~\citep{wu2025collabllm, zhou2024archer, shani2024multi, zhang2024modeling}. Our work is orthogonal to both: we change the paradigm through which feedback enters the model, without changing the model or its training.

\paragraph{Refinement of LLMs.} Self-refinement methods iterate a critique-and-revise loop with internal or external feedback~\citep{madaan2023self, dhuliawala2024chain, shinn2023reflexion,nathani2023maf, welleck2022generating}. Memory-augmented variants retain feedback or summarize past turns in an external buffer~\citep{madaan2022memory, packer2023memgpt}. These methods differ from ours along two axes. On the \emph{input} side, the erroneous response is retained, so generation still conditions on incorrect content. On the \emph{output} side, the entire response is regenerated at every turn, which inflates token cost and can overwrite spans that were already correct. In-place feedback removes the erroneous span from the conditioning context and continues generation from the corrected context.

\paragraph{Analysis of multi-turn LLM behavior.} A growing body of work analyzes how LLMs behave across turns, reporting degraded accuracy, instruction loss, and unreliable feedback uptake~\citep{wang2024mint, jiang2025feedback, laban2025llms, deshpande2025multichallenge}. These analyses report aggregate or task-level metrics. We complement them with a turn-level decomposition (CPR, FAR, CTRR) that separates \emph{retention} of correct content, \emph{acceptance} of explicit feedback, and \emph{autonomous correction} beyond stated feedback.
% Recent work analyzes LLM behavior in multi-turn conversations, identifying issues such as unreliable feedback incorporation and performance degradation across turns~\citep{jiang2025feedback, laban2025llms, sirdeshmukh2025multichallenge}. However, these studies mainly report high-level metrics such as task performance and do not examine turn-level dynamics or error propagation. In contrast, we conduct a turn-level analysis of feedback-driven refinement and introduce diagnostic metrics that quantify how errors propagate across turns. 

\paragraph{UI-level interactive editing.} A line of HCI work proposes systems that let users directly edit problematic spans of model outputs~\citep{Masson_2024, Xie2024wait,pang2025interactivereasoningvisualizingcontrolling}. These systems offer locality at the \emph{interface} level, but the model still conditions on the full prior response, so locality in the UI does not translate to locality in the conditioning context. We extend localized editing from the interface to the model's input, so that the corrected span replaces the original output in the actual context used for continuation.

We provide a more detailed comparison with prior work in \cref{app:related_work}.

%% file: sections/2_inplace_feedback.tex
\section{In-Place Feedback}
\label{sec:in_place_feedback}

\subsection{Multi-Turn Refinement with Feedback}
\label{sec:problem_formulation}

We describe how expert feedback is incorporated into LLMs in interactive settings. Let $\mathcal{M}$ be an LLM and $\mathcal{F}$ a feedback function. Given a prompt $x$, the LLM produces an initial response $y_0 = \mathcal{M}(x)$, and at each subsequent turn $t \geq 1$ generates a refined response by conditioning on the prompt together with the full history of previous responses and feedback,
$y_{t} = \mathcal{M}\left(x, y_{0}, f_{0}, \dots, y_{t-1}, f_{t-1}\right),$
where $f_i = \mathcal{F}(x, y_i)$ denotes the feedback associated with the $i$-th response. We refer to this process as refinement with standard multi-turn feedback, which we hereafter simply call \emph{multi-turn feedback}.

\subsection{Motivation: Three Failure Modes of Multi-Turn Feedback}
\label{subsec:motivation}

Recent work demonstrates that LLMs often fail to reliably integrate user feedback~\citep{laban2025llms, jiang2025feedback}. As illustrated in \cref{fig:in_place_failure_case}, we identify three recurring failure modes in multi-turn refinement: (i) previously correct content becomes corrupted after receiving feedback, (ii) the model ignores the feedback and reproduces its prior mistakes, and (iii) the model incorporates the feedback locally but generates new errors in subsequent reasoning. We present real-world failure cases of multi-turn refinement in \cref{app.multi_turn_failure_example}.

Failures of types (i) and (ii) can substantially reduce the efficiency of expert-in-the-loop refinement by forcing experts to repeatedly re-check content or re-state feedback that is already provided. Moreover, when such failures recur, the interaction may fail to converge, meaning that sustained multi-turn feedback does not reliably yield an error-free final response. We hypothesize that these failures result from two properties of multi-turn feedback: incorrect context accumulates in the dialogue history, and feedback arrives as a separate instruction that the model should interpret and locate. While failures of type (iii) may appear to stem from the limited reasoning capability of LLMs, our analysis in \cref{sec:analysis} reveals that accumulated dialogue history also plays a role. 

This hypothesis highlights three requirements for effective refinement. Edits from user feedback should target only the relevant reasoning step, preserve previously correct content outside the target span, and guide future reasoning from the corrected state rather than an outdated one. These considerations naturally lead us to ask: \emph{Can we mitigate the above failures by letting the user directly edit the targeted span and constraining the model to continue generation from that point?}

\subsection{In-Place Feedback}
\label{subsec:inplace_feedback}

To address this question, we propose \emph{in-place feedback}, a new interaction paradigm that treats feedback as a repair to the model's state rather than as a new instruction. As illustrated in \cref{fig:in_place_failure_case}, our method proceeds in two stages. The first, \emph{in-place edit}, allows the expert to directly modify the model's previous response. The expert then prunes the context that depends on the corrected span, while leaving the rest unchanged. In our setting, we assume the expert identifies one or two mistakes in the response and corrects only those parts. The second, \emph{continuation generation}, regenerates only what is necessary to continue from the updated context. Together, these stages limit unintended changes and rebuild the response from the correction. By retaining only the corrected span and continuing from it rather than regenerating the full response, in-place feedback also keeps the dialogue history compact, an efficiency benefit we quantify in \cref{sec:experimental_results}.

To illustrate how this method works in practice, \cref{fig:common_cases} presents representative cases of in-place feedback. In math problems, for example, in-place feedback can fix simple arithmetic mistakes or adjust flawed intermediate steps. When the error is more substantive, it can realign an incorrect reasoning path by revising larger portions of the solution.

%% file: sections/3_experimental_results.tex
\begin{figure*}[t]
  \centering
  \includegraphics[width=\linewidth]{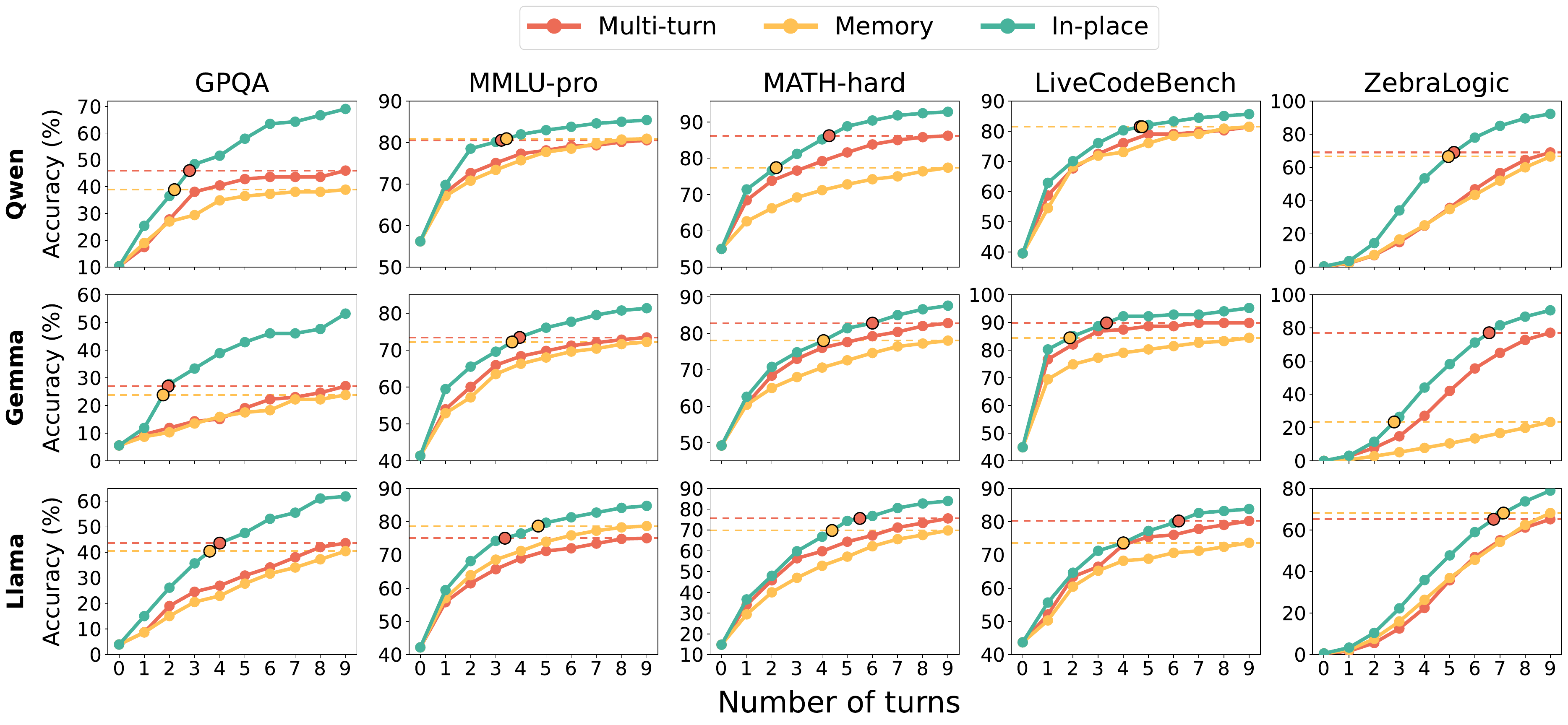}
  % \vskip -0.8 em
  \caption{Performance comparison across five datasets. Across all datasets and LLMs, in-place feedback consistently outperforms the multi-turn and memory-based feedback approaches. The points with black borders represent the baselines' performance at the last turn.}
  \label{fig:main_results}
  % \vskip -1.3 em
\end{figure*}

\section{Experiments}
\label{sec:experimental_results}

\subsection{Experimental Setup}
\label{sec:exp_set}

\paragraph{Datasets.} 
We evaluate in-place feedback on three reasoning tasks, MATH-hard~\citep{hendrycks2021measuring}, MMLU-pro~\citep{wang2024mmlu}, and GPQA~\citep{rein2024gpqa}, one coding task, LiveCodeBench~\citep{jain2025livecodebench}, and one logical reasoning task, ZebraLogic~\citep{lin2025zebralogic}. ZebraLogic is a collection of 573 Sudoku-like puzzles in which $N$ houses and $M$ attributes (\eg, \texttt{Name, Drink, Hobby}) form an $N\times M$ grid of cells. The values of these cells are uniquely determined by natural-language clues and uniqueness constraints. We use 500 level-5 problems from MATH-hard, the free-form subsets of MMLU-pro and GPQA~\citep{chandak2025answer}, and the code-generation problems from LiveCodeBench. Details are provided in \cref{app.experimental_details}.

\paragraph{Evaluation.}
We tailor the evaluation to each task. For MATH-hard, MMLU-pro, and GPQA, since each problem has a single ground-truth answer, we evaluate a response by checking whether its final answer matches the ground-truth. For LiveCodeBench, we execute the generated code on the provided test cases and mark a response correct only if all tests pass. For ZebraLogic, each puzzle admits a unique solution that fills every cell of the grid, so we mark a response correct only when all cells match the ground-truth solution. Further details on the evaluation protocols are provided in \cref{app.experimental_details}.

\paragraph{Baselines.}
We compare in-place feedback against two baselines: \emph{multi-turn feedback} (\cref{sec:problem_formulation}) and \emph{memory-based feedback}, which avoids retaining the full dialogue history by conditioning only on the latest response and the $k$ most recent feedback items~\citep{shinn2023reflexion, packer2023memgpt, madaan2022memory}: $y_t = \mathcal{M}(P(x, y_{t-1}, f_{t-1}, \ldots, f_{t-k}))$. Here, $P$ is a prompt template that combines the question, the previous response, and the $k$ most recent feedback items into a single revision prompt (see \cref{fig.memory}). We set $k=3$.

\paragraph{Feedback agents.} To run experiments at scale without human experts in the loop, we obtain feedback from a \emph{feedback agent} that stands in for the human experts across all methods. For benchmarks other than ZebraLogic, the agent has access to the ground-truth answer and returns a natural-language critique that points to the earliest critical error and a suggested fix, without revealing the correct final answer. We additionally post-process each critique to remove any unintended disclosure of the answer. For ZebraLogic, we instead obtain feedback systematically from a Z3 solver~\citep{de2008z3}, which identifies all conflicting cells. For each, the feedback names the cell and its correct value (\eg, \texttt{``the name of House~1 is Eric, not Alice''}); we provide the top-4 most critical violations per turn by default, with a top-2 ablation in \cref{app.zebralogic_extended_results}.

In standard multi-turn feedback, the feedback is passed directly as the input for the next turn. In-place feedback, by contrast, requires the targeted error to be located within the previous response and edited in place.
To automate this, in-place feedback additionally uses an \emph{in-place agent}. The agent takes the feedback produced by the feedback agent and the model's previous response as input. It then locates the sentence targeted by the feedback and replaces it with a corrected span. Finally, it removes subsequent text that may depend on the edit. We implement both agents with GPT-5-mini~\citep{OpenAI2025GPT5}. Details such as post-processing and prompt templates are provided in \cref{app.experimental_details}.

\paragraph{LLMs.} We experiment with five LLMs: \texttt{Qwen2.5-7B-Instruct}~\citep{qwen2025qwen25technicalreport}, \texttt{Gemma-3-4b-it}~\citep{gemmateam2025gemma3technicalreport}, \texttt{Llama-3.1-8B-Instruct}~\citep{grattafiori2024llama}, \texttt{Gemma-3-27b-it}, and \texttt{Llama-3.1-70B-Instruct}. The latter two are larger models that we use to examine scalability. We conduct ten refinement turns for the three smaller models and five for the two larger ones. Experimental settings are provided in \cref{app.experimental_details}.

\begin{figure}[t]
  \centering    
    \includegraphics[width=\columnwidth]{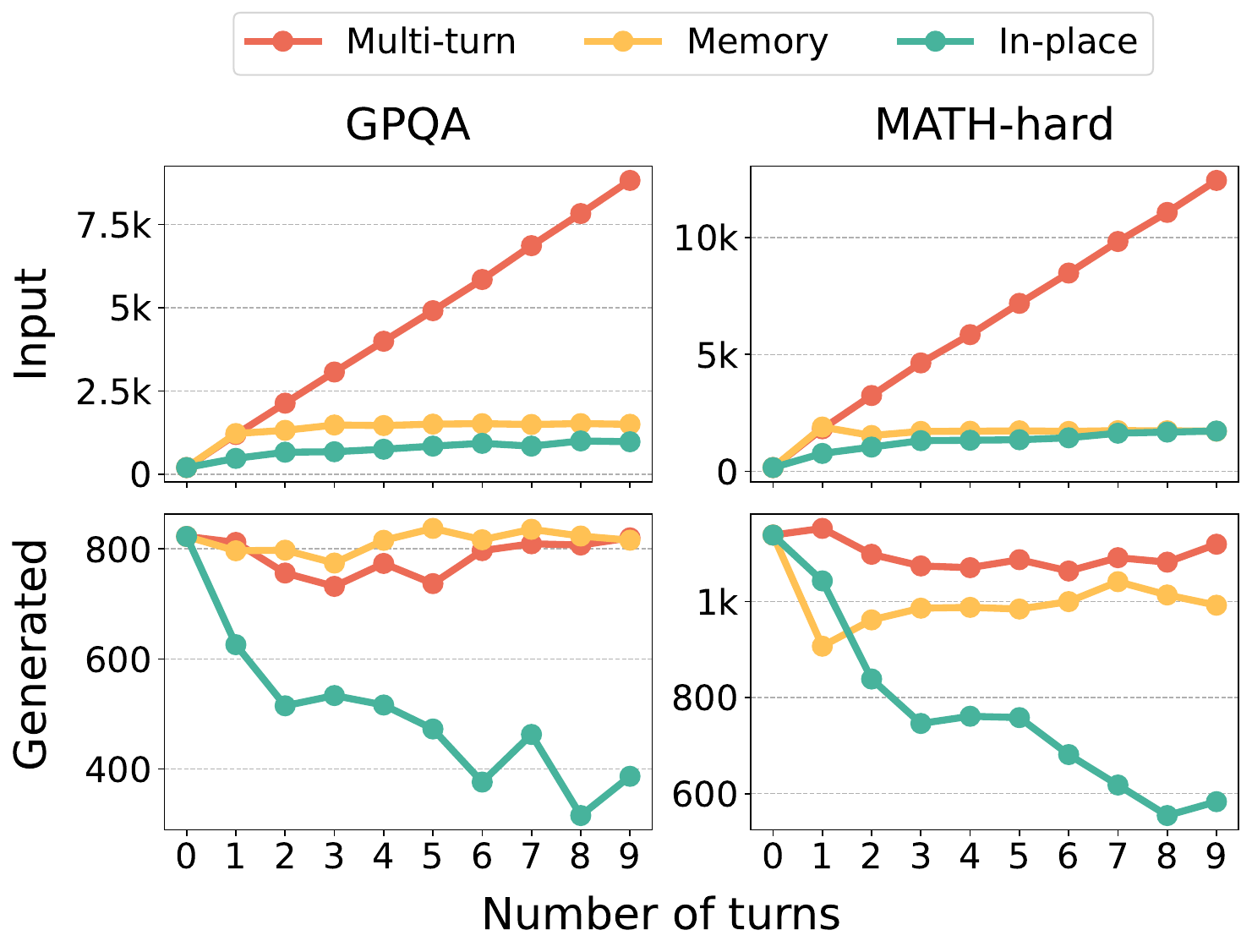}
    % \vskip -0.8 em
    \caption{The number of input and generated tokens across multiple turns for \texttt{Gemma-3-4b-it}. In-place feedback consistently requires fewer tokens than multi-turn feedback. The token counts for the other models and datasets can be found in \cref{app.tabulated_results}.}
    \label{fig:token_efficiency}
    % \vskip -1.3 em
\end{figure}

\begin{table*}[t]
    \centering
    \resizebox{\textwidth}{!}{%
    \begin{tabular}{lll cccccccccc}
        \toprule
        \multirow{2}{*}{Dataset} & \multirow{2}{*}{Model} & \multirow{2}{*}{Method}
        & \multicolumn{10}{c}{Number of turns} \\
        \cmidrule(lr){4-13}
        & & & $0$ & $1$ & $2$ & $3$ & $4$ & $5$ & $6$ & $7$ & $8$ & $9$ \\
        \midrule

        \multirow{6}{*}{MATH-hard}
        & \multirow{3}{*}{Qwen}
            & Multi-turn
            & $0.550$ & $0.684$ & $0.738$ & $0.766$ & $0.792$ & $0.816$ & $0.838$ & $0.850$ & $0.858$ & $0.862$ \\
        &   & Localized MT
            & $0.550$ & $0.682$ & $0.758$ & $0.790$ & $0.808$ & $0.830$ & $0.838$ & $0.848$ & $0.852$ & $0.858$ \\
        &   & In-place
            & $0.550$ & $\mathbf{0.714}$ & $\mathbf{0.766}$ & $\mathbf{0.812}$ & $\mathbf{0.852}$ & $\mathbf{0.888}$ & $\mathbf{0.904}$ & $\mathbf{0.918}$ & $\mathbf{0.924}$ & $\mathbf{0.928}$ \\
        \cmidrule(lr){2-13}

        & \multirow{3}{*}{Gemma}
            & Multi-turn
            & $0.492$ & $0.606$ & $0.684$ & $0.730$ & $0.760$ & $0.776$ & $0.792$ & $0.804$ & $0.820$ & $0.828$ \\
        &   & Localized MT
            & $0.492$ & $0.612$ & $0.692$ & $\mathbf{0.750}$ & $0.776$ & $0.794$ & $0.808$ & $0.824$ & $0.838$ & $0.846$ \\
        &   & In-place
            & $0.492$ & $\mathbf{0.626}$ & $\mathbf{0.708}$ & $0.748$ & $\mathbf{0.778}$ & $\mathbf{0.814}$ & $\mathbf{0.828}$ & $\mathbf{0.850}$ & $\mathbf{0.866}$ & $\mathbf{0.876}$ \\

        \midrule

        \multirow{6}{*}{GPQA}
        & \multirow{3}{*}{Qwen}
            & Multi-turn
            & $0.103$ & $0.175$ & $0.278$ & $0.381$ & $0.405$ & $0.429$ & $0.437$ & $0.437$ & $0.437$ & $0.460$ \\
        &   & Localized MT
            & $0.103$ & $0.238$ & $0.341$ & $0.381$ & $0.444$ & $0.484$ & $0.532$ & $0.564$ & $0.579$ & $0.611$ \\
        &   & In-place
            & $0.103$ & $\mathbf{0.254}$ & $\mathbf{0.365}$ & $\mathbf{0.484}$ & $\mathbf{0.516}$ & $\mathbf{0.579}$ & $\mathbf{0.635}$ & $\mathbf{0.643}$ & $\mathbf{0.667}$ & $\mathbf{0.691}$ \\
        \cmidrule(lr){2-13}

        & \multirow{3}{*}{Gemma}
            & Multi-turn
            & $0.056$ & $0.095$ & $0.119$ & $0.143$ & $0.151$ & $0.191$ & $0.222$ & $0.230$ & $0.246$ & $0.270$ \\
        &   & Localized MT
            & $0.056$ & $\mathbf{0.119}$ & $0.222$ & $0.270$ & $0.357$ & $0.421$ & $0.444$ & $\mathbf{0.468}$ & $\mathbf{0.492}$ & $0.516$ \\
        &   & In-place
            & $0.056$ & $\mathbf{0.119}$ & $\mathbf{0.278}$ & $\mathbf{0.333}$ & $\mathbf{0.389}$ & $\mathbf{0.429}$ & $\mathbf{0.460}$ & $0.460$ & $0.476$ & $\mathbf{0.532}$ \\

        \bottomrule
    \end{tabular}%
    }
    \caption{Cumulative accuracy on MATH-hard and GPQA with \texttt{Qwen-2.5-7B-Instruct} and \texttt{Gemma-3-4B-it}. \textit{Localized MT} uses the same in-place agent as in-place feedback, but the edited response is delivered as a multi-turn user message with the full dialogue history.}
    \label{tab:rebuttal-localized-mt}
\end{table*}

\subsection{Results}
\label{sec:inplace_feedback_ex}

\paragraph{Task performance.}
\cref{fig:main_results} shows how accuracy changes as the number of turns increases. Across all five datasets and models, in-place feedback consistently achieves higher accuracy and exhibits faster improvement over turns. On GPQA with \texttt{Gemma}, in-place feedback improves accuracy by 47.6 percentage points over the initial response after 10 turns, an improvement more than twice that of multi-turn feedback. On MMLU-pro, its accuracy at turn 5 already surpasses the final-turn performance of multi-turn feedback across all models. Similar trends hold on ZebraLogic. These results demonstrate that in-place feedback more effectively integrates external corrections into the reasoning of LLMs. 
Notably, these gains hold despite the in-place agent producing appropriate edits only 40.6\% of the time on MATH-hard with \texttt{Gemma} (\cref{app:robustness}). The same trend holds on larger-scale LLMs (\cref{app.large_scale_llms}).
Qualitative examples and tabulated results are provided in \cref{app.in_place} and \cref{app.tabulated_results}, respectively.

\paragraph{Token efficiency.}
Beyond task performance, in-place feedback is also more token-efficient. \cref{fig:token_efficiency} reports input and generated token counts. Multi-turn feedback appends each turn to the dialogue history, causing input tokens to grow linearly with turns. Both memory-based and in-place feedback avoid this accumulation and keep input length stable. The key difference emerges in generated tokens: memory-based feedback still regenerates the full response from scratch like multi-turn, whereas in-place feedback continues from the corrected span and produces substantially shorter outputs while preserving correct reasoning.

\subsection{Ablation Study} 

\paragraph{Effect of localization and continuation generation.} In-place feedback differs from multi-turn feedback in two aspects: it localizes and directly incorporates the correction into the previous response, and then continues generation from this corrected context. We conduct an ablation to disentangle these effects.
We introduce \textit{localized multi-turn (MT)}, which uses the same in-place agent to edit the previous response, but presents the edited response together with the same continuation prompt through a new user message while retaining the full dialogue history, rather than continuing directly from the edited context. Thus, localized MT isolates the benefit of localization while preserving the multi-turn interaction structure.

We conduct experiments on MATH-hard and GPQA with \texttt{Qwen-2.5-7B-Instruct} and \texttt{Gemma-3-4B-it}. As shown in \cref{tab:rebuttal-localized-mt}, localized MT generally improves over standard multi-turn feedback, while in-place feedback further improves over localized MT. This suggests that both localization and continuation from a clean corrected context contribute to the gains of in-place feedback.

\paragraph{Incomplete feedback.}
In practice, feedback sometimes only points out that a response is wrong, without indicating how to correct it. We evaluate in-place feedback under this weaker signal on MATH-hard, GPQA, and ZebraLogic, replacing the feedback (\eg, \texttt{``the name of House~1 is Eric, not Alice''}) with its incomplete counterpart (\texttt{``the name of House~1 is not Alice''}). In-place feedback still outperforms the multi-turn baseline across benchmarks, since removing the erroneous span allows the model to reason from a clean state rather than anchoring to prior mistakes. Full results and discussion are provided in \cref{app:robustness}.

%% file: sections/4_analysis.tex
\section{Failure Mode Analysis Under Controlled Experiment}
\label{sec:analysis}

\cref{sec:experimental_results} shows that in-place feedback consistently improves task performance over multi-turn feedback, but how these gains arise remains unclear. Task performance does not reveal which of the failure modes in \cref{fig:in_place_failure_case} are actually mitigated, specifically whether it prevents (i) corruption of previously correct content, (ii) ignoring of explicit feedback, or (iii) propagation of new errors in subsequent reasoning.
It also does not show how these effects evolve as the dialogue progresses. To address this gap, we conduct a turn-level analysis on ZebraLogic, whose rule-based feedback and discrete grid structure enable us to match intermediate states against ground truth at the cell level. 

\begin{figure}[t]
    \centering
    \includegraphics[width=\columnwidth]{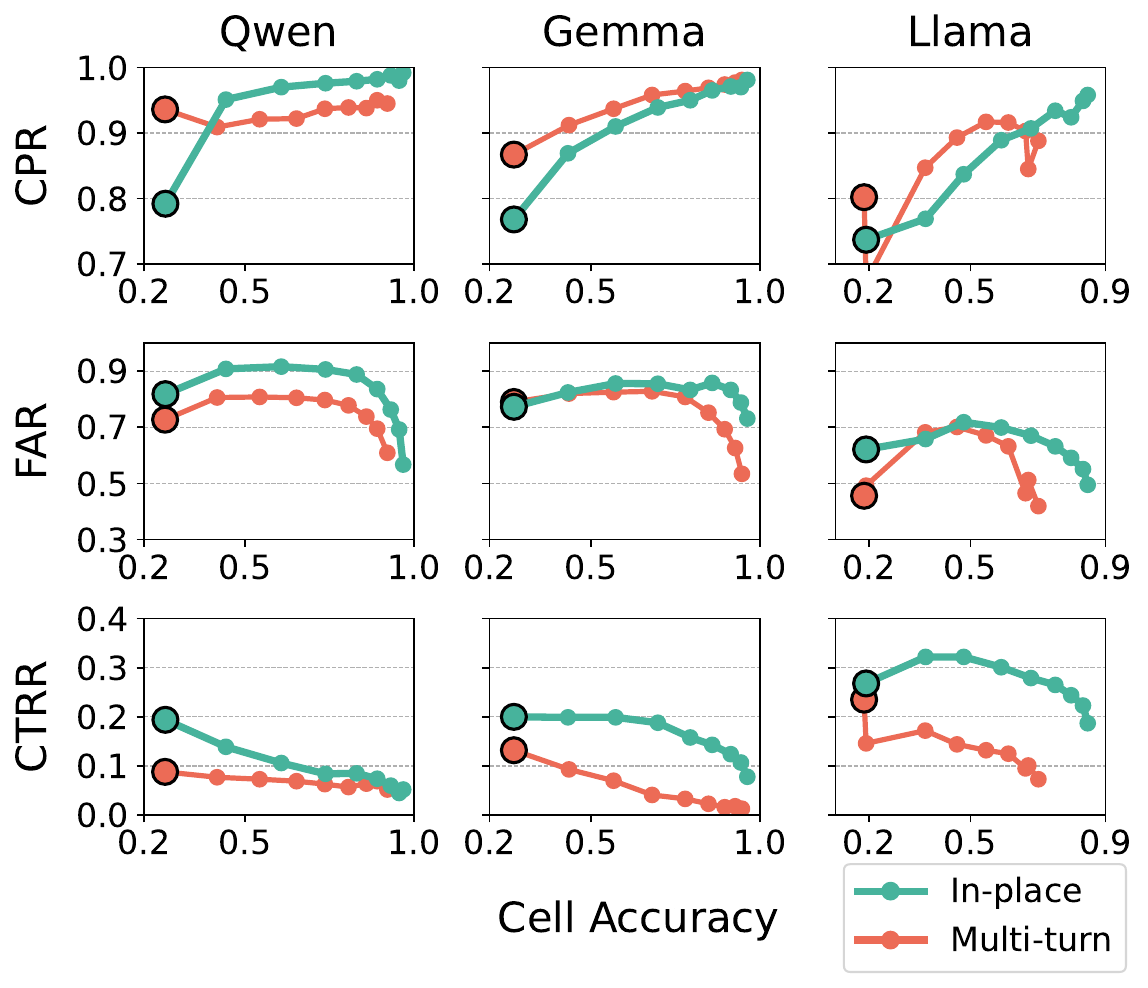}
    % \vskip -0.5em
    \caption{CPR, FAR, and CTRR for 10-turn. The points with a black border represent the second response of the LLMs (i.e., $y_1$), and the subsequent responses across turns are connected by lines.}
    % \vskip -1 em
    \label{fig:zebra_reasoning}
\end{figure}

\subsection{Metrics for Diagnosis of Failure Modes}

Each ZebraLogic puzzle is structured as a grid of cells, with one cell per entity-attribute pair. \cref{sec:experimental_results} scores a response correct only when every cell matches the solution. As a finer-grained complement, we also consider \emph{cell accuracy}, the average proportion of cells matching the solution, with full results reported in \cref{app.memory_based_feedback_performance}. Both, however, collapse the three failure modes from \cref{fig:in_place_failure_case} into a single number and do not reveal which mode dominates at which turn. To separate them, we introduce three cell-level metrics that compare consecutive responses $y_t$ and $y_{t+1}$, each targeting one of the failure modes.

\begin{itemize}
    \item \textbf{Correctness-Preserving Ratio (CPR).} The proportion of cells that are correct in $y_t$ and remain correct in $y_{t+1}$, relative to the total number of cells correct in $y_t$. \emph{This metric captures retention of valid reasoning and targets failure mode (i).}
    
    \item \textbf{Feedback Acceptance Ratio (FAR).} Among cells flagged by feedback $f_t$, the proportion corrected in $y_{t+1}$. \emph{This metric captures uptake of the explicit corrective signal and targets failure mode (ii).}
    
    \item \textbf{Correction Through Reasoning Ratio (CTRR).} Among cells incorrect in $y_t$ but not indicated by $f_t$, the proportion corrected in $y_{t+1}$. \emph{This metric captures whether the model can extend the correction beyond what feedback states, and targets failure mode (iii).}
\end{itemize}

% \begin{figure}[t]
%     \centering
%     \includegraphics[width=\columnwidth]{figures/analysis/analysis_top4-2.pdf}
%     \vskip -0.5em
%     \caption{CPR, FAR, and CTRR for 10-turn. The points with a black border represent the second response of the LLMs (i.e., $y_1$), and the subsequent responses across turns are connected by lines.}
%     \vskip -1 em
%     \label{fig:zebra_reasoning}
% \end{figure}

% \begin{figure}[t]
%     \centering
%     \includegraphics[width=\columnwidth]{figures/analysis/inplace_vs_multiturn.pdf}
%     \vskip -0.5em
%     \caption{CPR, FAR, and CTRR for 10-turn. The points with a black border represent the second response of the LLMs (i.e., $y_1$), and the subsequent responses across turns are connected by lines.}
%     \vskip -1 em
%     \label{fig:zebra_reasoning}
% \end{figure}

\subsection{Refinement Dynamics Across Turns}
\label{sec:analysis_feedback_utilization}

\begin{observation}
With multi-turn feedback, models in extended dialogues reliably preserve correct content but struggle to incorporate new feedback and reason further.
\end{observation}
\cref{fig:zebra_reasoning} reports CPR, FAR, and CTRR across turns. CPR remains above 0.9 throughout, and rises with turn depth for \texttt{Gemma} and \texttt{Llama}, indicating that models retain previously correct answers more reliably as history accumulates.
FAR and CTRR, however, decline steadily: FAR drops most sharply between turns 5 and 9, and CTRR eventually falls below 0.1. The models thus become less receptive to feedback and less able to reason beyond it as the dialogue lengthens. \texttt{Llama} sits at one extreme of this trade-off, attaining higher CTRR but lower CPR and FAR than the other models.

% \begin{figure}[t]
%     \centering
%     \includegraphics[width=\linewidth]{figures/controlled.pdf}
%     % \vskip -0.65 em
%     \caption{Impact of accumulated dialogue history. For CPR, FAR, and CTRR, we report the difference between the multi-turn feedback rates and the rates under pruned-history settings. Negative values indicate performance degradation due to accumulated history.}
%     \label{fig:zebra_controlled}
%     % \vskip -1.0 em
% \end{figure}

\begin{figure}[t]
    \centering
    \includegraphics[width=\linewidth]{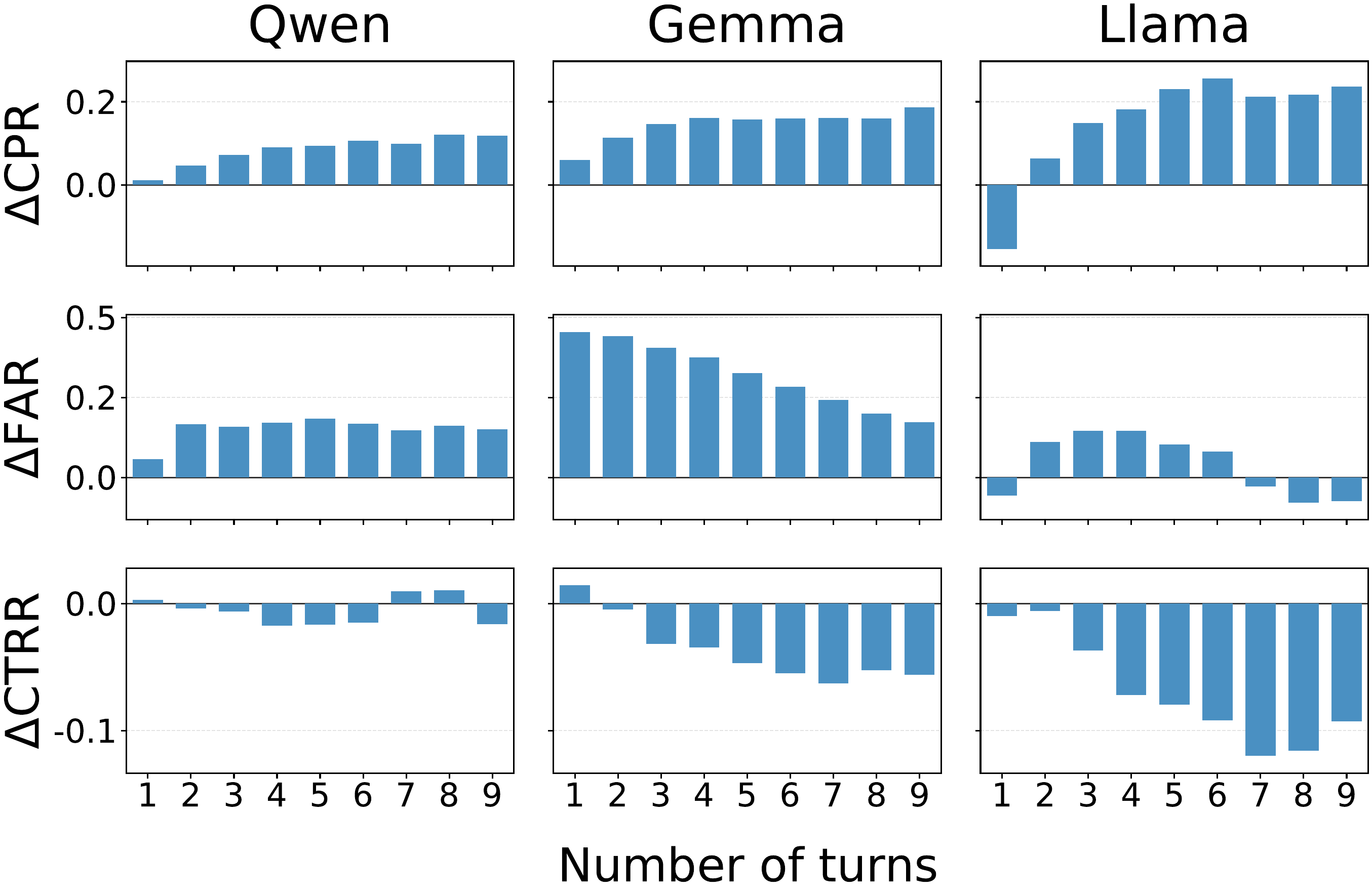}
    % \vskip -0.65 em
    \caption{Impact of accumulated dialogue history. For CPR, FAR, and CTRR, we report the difference between the multi-turn feedback rates and the rates under pruned-history settings. Negative values indicate performance degradation due to accumulated history.}
    \label{fig:zebra_controlled}
    % \vskip -1.0 em
\end{figure}

\begin{observation}
    In-place feedback is more receptive to corrections and propagates them better than multi-turn feedback.
\end{observation}
In-place feedback achieves a higher FAR than multi-turn feedback in later turns and consistently higher CTRR across all turns, accounting for the larger gap in accuracy. We attribute both gains to a single mechanism: the in-place edit applies the feedback directly to the response, and the model then continues from this corrected span rather than the full prior response. This reduces interference from stale reasoning and lets the model propagate the correction to cells the feedback never names explicitly. Nevertheless, even with in-place feedback, adoption remains imperfect since ZebraLogic's iterative reasoning often overwrites integrated corrections when revisiting cells.

% \paragraph{Suboptimality of in-place feedback in CPR and FAR.}
% One might expect CPR and FAR for in-place feedback to reach 1.0, since the edit applies the correction directly to the response. Two factors explain the gap.
% 1) \emph{Imperfect in-place agent.} As reported in \cref{tab:agent_reliability}, the agent produces appropriate edits in only 53.9\% of cases on ZebraLogic, with under-pruning of dependent reasoning as the dominant failure mode, leaving the targeted cell unfixed. 
% 2) \emph{Re-derivation during continuation.} Even when the edit correctly enforces the feedback, the LLM may revisit the same cell during generation. Since ZebraLogic involves parallel reasoning, a cell is referenced across several steps before being written into the final answer grid, and the continuation can re-derive a different value, especially when top-$k$ feedback addresses only the $k$ most critical violations per turn and leaves other incorrect cells in the edited context. 

% This is a trade-off rather than a failure: the same active re-reasoning lets in-place feedback fix cells the feedback never points to, yielding consistently higher CTRR than multi-turn feedback. 
% Under oracle feedback, where edits can be error-free and no residual conflicts remain, both CPR and FAR are typically higher for in-place than for multi-turn feedback, except for CPR with \texttt{Qwen} in the first turn (see \cref{fig:zebra_analysis_oracle}).

\subsection{Effect of Dialogue History for Multi-turn Feedback}
\label{sec:effect_history}

While multi-turn feedback accumulates all prior interactions, in-place feedback conditions generation solely on the edited context. To isolate the effect of accumulated dialogue history, we implement an ablation: at each turn $t$, we re-generate the response with the context pruned to the problem, the most recent response, and the immediate feedback, $\tilde{y}_t = \mathcal{M}(P(x, y_{t-1}, f_{t-1}))$. \cref{fig:zebra_controlled} reports the gap between $\tilde{y}_t$ and the original $y_t$ across CPR, FAR, and CTRR; any difference is attributable to the presence of earlier dialogue history.

\begin{observation}
    While accumulating history helps models reliably reproduce correct content, it simultaneously stifles their ability to reason beyond explicit feedback.
\end{observation}
Accumulated dialogue history has opposing effects on retention and revision. As shown in \cref{fig:zebra_controlled}, the CPR gap widens with turn depth, so the full history helps the model preserve previously correct content. The CTRR gap moves in the opposite direction and turns negative for \texttt{Gemma} and \texttt{Llama}, so the same history that aids retention also obstructs the extension of reasoning beyond what feedback states. We conjecture this trade-off reflects induction-head-style copying~\citep{olsson2022incontextlearninginductionheads}, which raises CPR by reproducing prior content but lowers CTRR by suppressing new reasoning. A deeper mechanistic analysis is left to future work. 

\begin{observation}
    FAR trends under multi-turn feedback vary across models, unlike the uniform trends observed in CPR and CTRR.
\end{observation}

\texttt{Qwen} maintains consistently high FAR, \texttt{Gemma} shows diminishing gains in later turns, and \texttt{Llama} benefits initially but reverses in later turns. Since FAR is closely tied to instruction-following capabilities, which are heavily shaped by model-specific post-training, we attribute these patterns to a complex interplay between post-training differences and shared mechanistic inductive biases. 

In-place feedback sidesteps this heterogeneity, reaching the highest FAR across all three models in later turns (\Cref{fig:zebra_reasoning}). We attribute this to its mechanism: in-place feedback places the correction directly into the response, rather than sending it as a separate instruction for the model to interpret and apply. This removes the need to treat feedback as an external command, yielding consistently high FAR regardless of model-specific instruction-following behavior.
% We further discuss FAR in \cref{app.extend_analysis}.

%% file: sections/5_user_study.tex
\begin{table*}[t]
\centering
% \vskip -0.7em

\small
\begin{tabularx}{\textwidth}{@{}l X X@{}}
\toprule
\textbf{Method} & \textbf{Strengths} & \textbf{Weaknesses} \\
\midrule
Multi-turn &
\textit{``Generates natural sentences that reflect my intended nuance when I know the direction but not the exact wording.'', ``Good at revising the overall flow.''} &
\textit{``Fatigue becomes severe when the model repeats the same mistake after correction.'', ``Untargeted parts get modified, and previously accepted edits are undone.'', ``Waiting for full regeneration every turn is tiring.''} \\
\addlinespace
In-place &
\textit{``I can quickly convey my intent without making the model understand it first.'', ``Convenient for fixing a word or phrase directly.'', ``Useful when the opening of a sentence is fixed and I want the model to continue naturally.''} &
\textit{``The model's drafting ability is hard to leverage when I don't know how to fix it myself.'', ``Sentences after the edited region sometimes change slightly.'', ``Tedious when many small items need to be removed at once.''} \\
\addlinespace
Mixed &
\textit{``Restructuring with multi-turn first and then polishing details with in-place was the most efficient.'', ``Combines the advantages of both methods.'', ``Lowest fatigue and highest satisfaction overall.''} &
\textit{``Choosing which mode to use for each edit is itself a small burden.'', ``I tended to use in-place as the default and multi-turn only as a fallback.''} \\
\bottomrule
\end{tabularx}
\caption{Summary of user experience across interaction modes. Representative strengths and weaknesses of three interaction modes from open-ended responses collected in our user study.}
\label{tab:feedback-methods}
% \vskip -1.2em
\end{table*}

\section{User Study}
\label{sec:user_study}

To complement the automated evaluations in \cref{sec:experimental_results,sec:analysis}, which use LLM-based agents in place of human experts, we conduct a user study in which expert participants directly refine LLM-generated text under different feedback paradigms. The study was approved by our institutional review board.

\paragraph{Setup.}
We recruit 15 participants, each of whom has authored at least one paper in English so that they can act as domain experts on a scientific summarization task. Prior to the session, each participant nominates papers they are familiar with. During the session, the system prompts \texttt{Gemma-3-27b-it} to generate a summary of a designated portion of one such paper at a participant-specified length. The participant iteratively refines that summary across three interaction modes: (i) \emph{multi-turn} feedback, (ii) \emph{in-place} feedback, and (iii) a \emph{mixed} mode in which the participant freely chooses between the two mechanisms for each correction. Each participant completes up to three topics in a 90-minute session, with the order of the three modes counterbalanced across topics to mitigate ordering effects. Procedural details and the participant interface are deferred to \cref{app.user_study} and \cref{app.user_study_interface}, respectively.

\begin{figure}[t]
    \centering
    \includegraphics[width=1.0\linewidth]{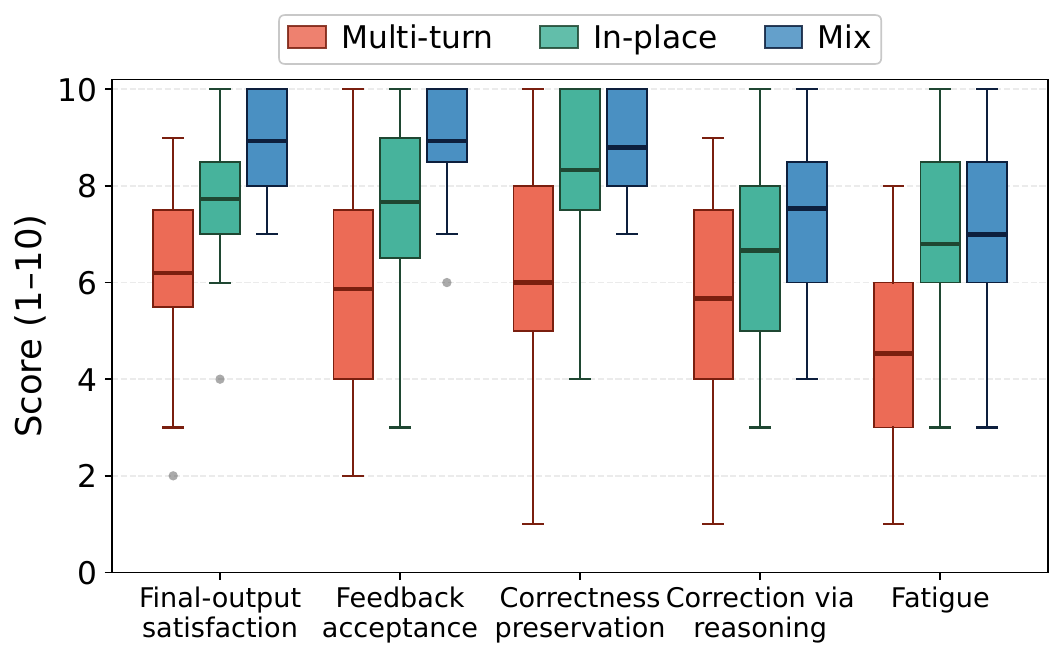}
    % \vskip -0.8 em
    \caption{User study results across the five questionnaire items, scored on a 1--10 scale (higher is better; for \emph{fatigue}, 10 corresponds to \emph{not fatigued}). In-place feedback outperforms standard multi-turn feedback on every dimension, and a mixed strategy that combines both paradigms is rated highest overall.}
    \label{fig:user_study}
    % \vskip -1.0 em
\end{figure}

\paragraph{Metrics.}
After each session, participants rate the session on a 1--10 scale along five dimensions: final-output satisfaction, feedback acceptance, correctness preservation, correction through reasoning, and fatigue. The middle three items are the human-perceived counterparts of FAR, CPR, and CTRR (\cref{sec:analysis}), allowing us to examine whether the turn-level dynamics observed in our automated experiments are also reflected in expert perception. A semi-structured interview at the end of each session collects open-ended impressions; representative quotes are summarized in \cref{tab:feedback-methods}.

\paragraph{In-place versus multi-turn.}

\Cref{fig:user_study} shows that in-place feedback is rated above multi-turn feedback on every item, with the largest gap appearing on \emph{fatigue}. As shown in \cref{tab:feedback-methods}, interview responses attribute the satisfaction gap to reliability rather than expressivity: participants praise multi-turn for generating natural sentences that reflect their intended nuance, but its weaknesses dominate the open-ended responses, as the model repeats already-corrected mistakes and modifies untargeted parts while undoing previously accepted edits. The fatigue gap follows from the re-verification cost imposed by full regeneration: each turn produces a fresh draft that must be re-read end-to-end, a burden amplified by the same failure modes captured by low FAR in \cref{sec:analysis}. In-place feedback bounds this re-verification region by localizing edits and resuming generation only from the corrected span.

\paragraph{Mixed mode feedback.}
The mixed mode is rated highest on every item, and the interviews suggest that it succeeds by exploiting the complementarity of the two paradigms rather than by introducing a new mechanism. Participants describe a common workflow in which they restructure with multi-turn and then polish details with in-place, which they report as the most efficient and as yielding the lowest fatigue and highest satisfaction. The only reported downside is mild and structural: deciding which mode to use for each edit is a small burden, with several participants noting that they tend to use in-place as the default and multi-turn as a fallback.

%% file: sections/7_conclusion.tex
\section{Conclusion}
\label{sec:conclusion}

We introduce in-place feedback, an interaction method in which the expert directly edits the model's prior response and generation continues from the edited context. Across reasoning benchmarks, controlled experiments on ZebraLogic, and a user study with domain experts refining scientific summaries, in-place feedback applies corrections more reliably across turns, lowers expert fatigue, and uses fewer tokens than standard multi-turn feedback. The user study further finds that a mixed strategy combining both paradigms scores highest on every measured dimension. We therefore recommend in-place feedback as a sensible default for expert-LLM collaboration, while exposing standard multi-turn feedback alongside it lets experts choose the most effective tool for each correction.

% \newpage
\section*{Limitations}

Due to limited access to large-scale GPU clusters and commercial LLM APIs, we are unable to run experiments with substantially larger models or closed-source models. Moreover, in-place feedback depends on a text-continuation API, which most commercial providers do not currently offer, limiting our ability to evaluate closed-source models. As a result, our empirical evaluation focuses on models that are feasible to serve within an academic computing budget. We believe, however, that in-place feedback is not tied to a specific model size or provider, and we expect the main conclusions to continue to hold as more powerful models become available in future work.

Our automated evaluation relies on an LLM-based in-place agent to simulate expert edits at scale. As reported in \cref{app:robustness}, this agent produces appropriate edits in only 40.6\% of cases on MATH-hard and 53.9\% on ZebraLogic, with the dominant failure modes being incomplete repair and imperfect pruning of dependent reasoning. While in-place feedback consistently outperforms multi-turn feedback even under this noisy editor, the absolute gains reported in \cref{sec:experimental_results,sec:analysis} could underestimate what is achievable in real expert–LLM collaboration, where expert edits are typically cleaner and more targeted.

Our user study (\cref{sec:user_study}) involves 15 expert participants in a scientific paper summarization task, with a 90-minute session per participant. Although the participants are domain experts who have authored English-language papers, and the three interaction modes are counterbalanced to mitigate ordering effects, broader claims about expert-LLM collaboration would benefit from larger-scale studies spanning more diverse task domains (\eg, legal drafting, clinical documentation, code review), participant populations, and longer-term usage patterns that capture how preferences and fatigue evolve over repeated sessions. We leave such broader human-centered evaluation to future work.

% Although our work focuses on expert–LLM interaction, obtaining experts' judgments for every benchmark instance would be prohibitively costly and difficult to scale. We therefore adopt LLM-based automatic evaluation as a practical and scalable proxy for human assessment in our experiments. This choice allows us to systematically compare different configurations of in-place feedback over a large set of examples, while keeping the evaluation procedure consistent and reproducible.

% We analyze turn-level dynamics of the feedback mechanism using ZebraLogic in \cref{sec:analysis}. ZebraLogic enables controlled experiments by automatically generating feedback via the Z3 solver, without relying on a feedback agent. While correcting an early step in sequential reasoning tasks would naturally propagate improvements to subsequent reasoning, ZebraLogic involves parallel reasoning where cells can be determined independently, meaning that errors in unedited portions can persist after in-place correction. This issue can be amplified by imperfect edits from the in-place agent, which may fail to remove all dependent reasoning. As a result, our analysis may understate the practical gains of in-place feedback. Nonetheless, the controlled nature of ZebraLogic enables clear attribution of effects, yielding useful insights into refinement dynamics and exposing fundamental constraints of multi-turn feedback.

\section*{Ethical Considerations}

Our method allows users to edit an LLM response and then continue generating from that edit. However, such edits can be misused to bypass safety mechanisms. For example, a user might insert harmful instructions or unsafe text, similar to jailbreak attacks. A straightforward defense is to run user edits through a safety filter before resuming the generation process. We leave the design of defenses against such attacks to future work.

\section*{Acknowledgements}
This work was supported by the National Research Foundation of Korea (NRF) grant funded by the Korea government (MSIT) (RS-2024-00337955); by the Institute of Information \& Communications Technology Planning \& Evaluation (IITP) grants funded by the Korea government (MSIT) (RS-2024-00457882, Artificial Intelligence Research Hub Project; RS-2019-II191906, Artificial Intelligence Graduate School Program (POSTECH)); and by IITP under the Leading Generative AI Human Resources Development (IITP-2026-RS-2026-25546560) grant funded by the Korea government (MSIT).

% We utilize four open-sourced LLMs and one closed-sourced LLM. All experimental settings and prompts are provided in {%
%   \crefname{section}{Section}{Sections}%
%   \Crefname{section}{Section}{Sections}%
%   \cref{sec:exp_set}, \cref{sec:exp_set2}%
% }, and \cref{appendix.experimental_details} to ensure reproducibility. For the closed-source LLM, we use the \texttt{gpt-5-mini-2025-08-07} version.

%% file: sections/a_appendix.tex
% Appendix section references

\crefalias{section}{appendix}
\crefalias{subsection}{appendix}

\numberwithin{figure}{section}
\numberwithin{table}{section}

\crefname{figure}{Figure}{Figures}
\crefname{table}{Table}{Tables}

\section{Experimental Details}
\label{app.experimental_details}

\subsection{Datasets}

\paragraph{MMLU-pro and GPQA free-form.}

The MMLU-pro and GPQA-Diamond datasets are employed to evaluate the knowledge-intensive reasoning abilities of LLMs in free-form question answering~\citep{chandak2025answer}. Both datasets originate from multiple-choice benchmarks but are adapted for generative evaluation via answer matching.

\paragraph{Motivation for free-form evaluation.}
Multiple-choice evaluation has intrinsic limitations: models can exploit statistical patterns in the answer choices without engaging in critical reasoning. As a result, multiple-choice accuracy overestimates the model's ability to generate correct answers. When the choices are removed and free-form responses are required, models are forced to generate an answer directly, aligning the evaluation more closely with the capabilities that matter in real-world use cases.

\paragraph{Selection of questions.}
MMLU-pro is derived from the MMLU benchmark and initially includes approximately 12,000 questions across various domains. To ensure that the questions are answerable without the provided answer choices, an automatic filtering process is employed. This process uses a rubric-based grader to narrow down the dataset to about 5,500 questions. From this reduced pool, questions that are sufficiently specific and have a unique correct answer are selected. After this, a manual filtering step yields a final set of 493 questions. This dataset offers comprehensive coverage across diverse domains, making it a useful resource for evaluating general knowledge and reasoning skills.

The GPQA-Diamond dataset comprises 198 graduate-level science questions, designed to be challenging and to test in-depth knowledge and critical reasoning. As with MMLU-pro, a filtering process is applied to select questions with clear and specific correct answers, resulting in a final set of 126 questions. This dataset is more focused and rigorous, emphasizing high-quality scientific questions that demand deep reasoning.

These preparation steps yield two complementary free-form datasets: MMLU-pro, which provides broader coverage across domains with 493 carefully filtered and annotated items, and GPQA-Diamond, which offers a smaller but more rigorous collection of 126 scientific questions. This ensures that free-form evaluations are conducted only on questions with clear, unambiguous solutions.

\paragraph{Evaluation.}
We evaluate model answers in two stages on MATH-hard, MMLU-pro, and GPQA. We first attempt exact matching. If it fails, we then apply an LLM judge to identify semantically equivalent answers expressed in different forms. In LiveCodeBench, we evaluate correctness by executing the provided test cases. We use GPT-oss-20b~\citep{openai2025gptoss120bgptoss20bmodel} as the judge model.

\subsection{ZebraLogic}
\label{app.zebralogic_explanation}

The ZebraLogic dataset comprises logic grid puzzles designed to assess the reasoning capabilities of LLMs. Each puzzle is structured around a grid with a certain number of houses and attributes. Specifically, each puzzle involves $N$ houses and $M$ attributes, creating an $N \times M$ grid that needs to be filled. The attributes in these puzzles are distinct for each house, where each attribute has $N$ unique values corresponding to the houses.

The puzzles come with a set of clues that impose logical constraints on the grid. For example, one clue might specify that ``\texttt{The person who likes milk is Eric}'', while another might state, ``\texttt{The person who drinks water is Arnold}''. These clues help guide the reasoning process to fill in the grid, ensuring that all constraints are satisfied. Importantly, each puzzle has a unique solution, which guarantees that any feedback provided for solving the puzzle is definitive and accurate. This structure ensures that ZebraLogic provides a rigorous framework for evaluating logical reasoning in models, where the set of clues provided uniquely determines each puzzle's solution.

\paragraph{Puzzle generation.}

Puzzles are generated by first sampling a complete solution, then constructing a superset of consistent clues from a fixed inventory. The clue types are as follows: \texttt{FOUNDAT}, \texttt{SAMEHOUSE}, \texttt{NOTAT}, \texttt{DIRECTLEFT/RIGHT}, \texttt{SIDEBYSIDE}, \texttt{LEFT/RIGHTOF}, and \texttt{ONE/TWOBETWEEN}. Each clue type provides a constraint by capturing a specific relationship between variables. A minimal subset of clues is retained through iterative pruning while preserving the uniqueness of the solution. This guarantees that puzzles are neither under-specified nor trivially over-constrained.

\paragraph{Dataset filtering.}

The ZebraLogic dataset contains 1{,}000 puzzles spanning all combinations of $N, M \in \{2,\ldots,6\}$. We filter the dataset by puzzle difficulty based on the search space size, which is defined as the total number of possible configurations that satisfy only the uniqueness constraints of the puzzle; for a puzzle with $N$ houses and $M$ attributes, the search space size is $|\mathcal{S}| = (N!)^M$.
Puzzles with small search spaces ($|\mathcal{S}| < 10^3$) are excluded, along with those of size $3\times4$ and $3\times5$, as well as invalid puzzles (\eg, cases where distinct categories share identical attribute values). This yields a controlled, reasoning-intensive testbed for analyzing the feedback acceptance of LLMs in multi-turn settings.

\paragraph{Input and output format.}
Puzzles are presented in natural language, followed by an instruction asking the model to fill the $N\times M$ grid. The input template prompt is provided in \cref{fig.input_zebra}. The expected output is a structured JSON table. This format enables automatic cell-level evaluation and the application of fine-grained feedback. We attempt up to 30 re-generations to obtain a syntactically valid JSON output. If all attempts fail, we treat the instance as wrong and omit it from CPR, FAR, and CTRR calculations. 

For JSON-parsed predictions, we employ a fuzz score from the Python rapidfuzz library to perform approximate string matching. Specifically, we compute the highest similarity score among the candidate attributes, and if the score exceeds 50, we adopt the corresponding attribute as the predicted value.

\paragraph{Feedback functions.} We generate rule-based feedback using a fixed template, as illustrated in \cref{fig:feedback_zebra}. For \texttt{Llama}, we additionally append reasoning guidance to the feedback, since the model frequently produces only the final JSON-formatted answer without including the corrected reasoning process.

\paragraph{In-place agent.}

The in-place agent simulates a human editor by directly modifying the LLM's response during evaluation. In math problems, each reasoning step depends on the previous one, so a correction usually requires discarding subsequent steps. Zebra puzzles, however, follow a different structure. They involve parallel reasoning, in which multiple constraints must be satisfied simultaneously. As a result, later reasoning steps can remain valid even if earlier ones are incorrect. To handle this, we retain subsequent reasoning steps after in-place feedback is applied.

We first segment the model's response into reasoning steps. The agent checks each step against the received feedback and edits the response when a directly mispredicted attribute value is specified (\eg, changing \texttt{Name of House 2 = Alice} to \texttt{Name of House 2 = Eric}). If there are reasoning steps that depend on the mispredicted value identified by the feedback (\eg, reasoning built on \texttt{Alice in House 2}), those dependent steps are removed to prevent error propagation. After applying these edits or deletions, the final solution is removed, and the prompt \texttt{Further reasoning:} is appended to encourage continuation of the reasoning process. We employ GPT-5-mini as the in-place feedback agent, following the rule prompt in \cref{appendix.intervention_zebra}.

\subsection{Prompt and Post-Processing}

\paragraph{Empirical experiments.}
We define answer leakage as any explicit revelation of the ground-truth answer within feedback or intervention outputs. Following \citet{jiang2025feedback}, we prevent it by using the prompts in \cref{fig.feedback,fig.intervention_math,fig.intervention_code} and applying post-processing for each agent: (i) after generating feedback, we scan the message and mask any span that reveals the ground truth; (ii) for the in-place feedback agent, if any part of the message exposes the solution, we prune those spans before presenting the message to the model.

\paragraph{ZebraLogic.} We constrain the in-place feedback agent to avoid introducing reasoning beyond the scope of the provided feedback. The model's output is segmented into discrete reasoning steps, which are then checked for consistency with the provided feedback. In cases of conflict, the corresponding step is minimally revised, following the prompt in \cref{appendix.intervention_zebra}. Notably, the agent does not have direct access to the puzzle itself, which limits its ability to extend reasoning beyond the explicitly given feedback.

\subsection{LLMs.}
We experiment with five LLMs: \texttt{Qwen2.5-7B-Instruct}~\citep{qwen2025qwen25technicalreport}, \texttt{Gemma-3-4b-it}~\citep{gemmateam2025gemma3technicalreport}, \texttt{Llama-3.1-8B-Instruct}~\citep{grattafiori2024llama}, \texttt{Gemma-3-27b-it}, and \texttt{Llama-3.1-70B-Instruct}. The latter two are larger models that we use to examine scalability. We use publicly available models released under their respective licenses.

\subsection{Hyperparameters}
\label{app.hyperparams}
All models use temperature 0. We observed that the \texttt{Llama} family can be overly verbose, resulting in degradation of generation quality. To stabilize decoding, we set its repetition penalty to 1.15. All other models use a repetition penalty of 1.0. The maximum generation length is 2048 tokens for all models and experiments. We use vLLM~\citep{kwon2023efficient} for efficient inference.

\subsection{Memory-based Feedback}

Following \citet{shinn2023reflexion}, we retain at most the three most recent feedback items. \cref{fig.memory} shows the prompt template for memory-based feedback.

\section{User Study Details}
\label{app.user_study}

\subsection{Study Setup}

\paragraph{Participants.} We recruit 15 participants. To qualify, each participant must have authored at least one paper written in English. This criterion ensures that participants have the reading comprehension and academic writing skills required to act as domain experts: they can reliably judge whether a model-generated summary faithfully reflects a source paper and can issue targeted, well-formed corrections. All participants provide informed consent, take part voluntarily, and may withdraw at any time. Each participant receives \$20 as compensation for the 90-minute session. The study was approved by our institutional review board. All participant data is anonymized prior to analysis.

\paragraph{Task.} We conduct a user study on a scientific summarization task. Prior to the session, each participant nominates papers they are already familiar with. At the start of each topic, the participant specifies the desired summary length, which is then incorporated into the system prompt provided to the LLM. The system subsequently prompts the LLM to generate a summary of the designated portion of the paper at the requested length. Since the participant knows the source material well, they can refine the generated summary, not only correcting factual or logical errors but also improving its wording, structure, and completeness, mirroring the expert-in-the-loop refinement setting. We place no restriction on the kinds of edits a participant may make, though we recommend against vague, sweeping revisions (\eg, paraphrasing the entire summary at once). Within a 90-minute session, each participant completes up to three summarization topics, each corresponding to a distinct paper or section of a paper.

\paragraph{Interaction modes.} We compare three interaction modes: (i) \emph{standard multi-turn feedback}, where the participant appends natural-language feedback messages and the model regenerates the full
response; (ii) \emph{in-place feedback}, where the participant directly edits the erroneous span and the model continues generation from the edited context; and (iii) \emph{mixed mode}, where the participant may freely choose, for each correction, between the two mechanisms. For each summarization topic, the participant carries out the refinement process under all three interaction modes, so that the conditions can be compared on the same source material. The order of the three modes is counterbalanced across topics and participants to mitigate ordering effects.

\paragraph{Model.} We use \texttt{Gemma-3-27b-it} as the underlying LLM for all conditions with a temperature of 0.0, consistent with the large-scale model setting in our main experiments.

\paragraph{Procedure.} Each session begins with a short onboarding in which participants are introduced to the interface and the three feedback mechanisms, followed by a practice trial on a warm-up paper to familiarize them with the workflow. Participants then complete their summarization topics, each under all three interaction modes. We provide examples of the user study interface in \cref{app.user_study_interface}. After the session, participants complete a post-study survey to evaluate each of the three feedback mechanisms.

\paragraph{Metrics.} After each interaction mode, participants answer five questions, each rated on a 1--10 scale:
\begin{itemize}
    \item \textbf{Final-output satisfaction.} How satisfied the participant is with the final generated summary produced under this method.
    \item \textbf{Feedback acceptance.} Whether the next-turn response sufficiently incorporates the feedback provided by the participant.
    \item \textbf{Correctness preservation.} Whether the next-turn response preserves the parts the participant is already satisfied with and does not want changed.
    \item \textbf{Correction through reasoning.} Whether the next-turn response also satisfactorily improves parts that are not explicitly mentioned in the participant's feedback.
    \item \textbf{Fatigue.} The participant's overall fatigue while refining the output with this mode, taking into account all relevant factors such as task efficiency and the difficulty of the feedback mechanism.
\end{itemize}
The feedback acceptance, correctness preservation, and correction through reasoning items are the human-perceived counterparts of the Feedback Acceptance Ratio (FAR), Correctness-Preserving Ratio (CPR), and Correction Through Reasoning Ratio (CTRR) defined in \cref{sec:analysis}, respectively, allowing us to examine whether the turn-level dynamics observed in our automated experiments are also reflected in expert perception. In addition to the questionnaire, we conduct a semi-structured interview at the end of the session to collect open-ended impressions of each feedback paradigm, including perceived control, effort, and trust.

\subsection{Representative User Comments}
\cref{tab:feedback-methods} summarizes representative strengths and weaknesses of each interaction mode, drawn from the open-ended responses collected in our post-study survey. Non-English comments were translated into English using an LLM, and all resulting entries were manually reviewed and verified by the authors to ensure that the original intent and nuance were faithfully preserved.

\section{The Use of Large Language Models}

We used an LLM assistant as a writing tool for grammar checking and paraphrasing. In addition, since our experiments required carefully designed prompts, we employed the assistant to refine prompts.

\section{Extended ZebraLogic Results}
\label{app.zebralogic_extended_results}

\subsection{Grid Accuracy and Cell Accuracy}
\label{app.memory_based_feedback_performance}

\begin{figure}[H]
    \centering
    \includegraphics[width=\linewidth]{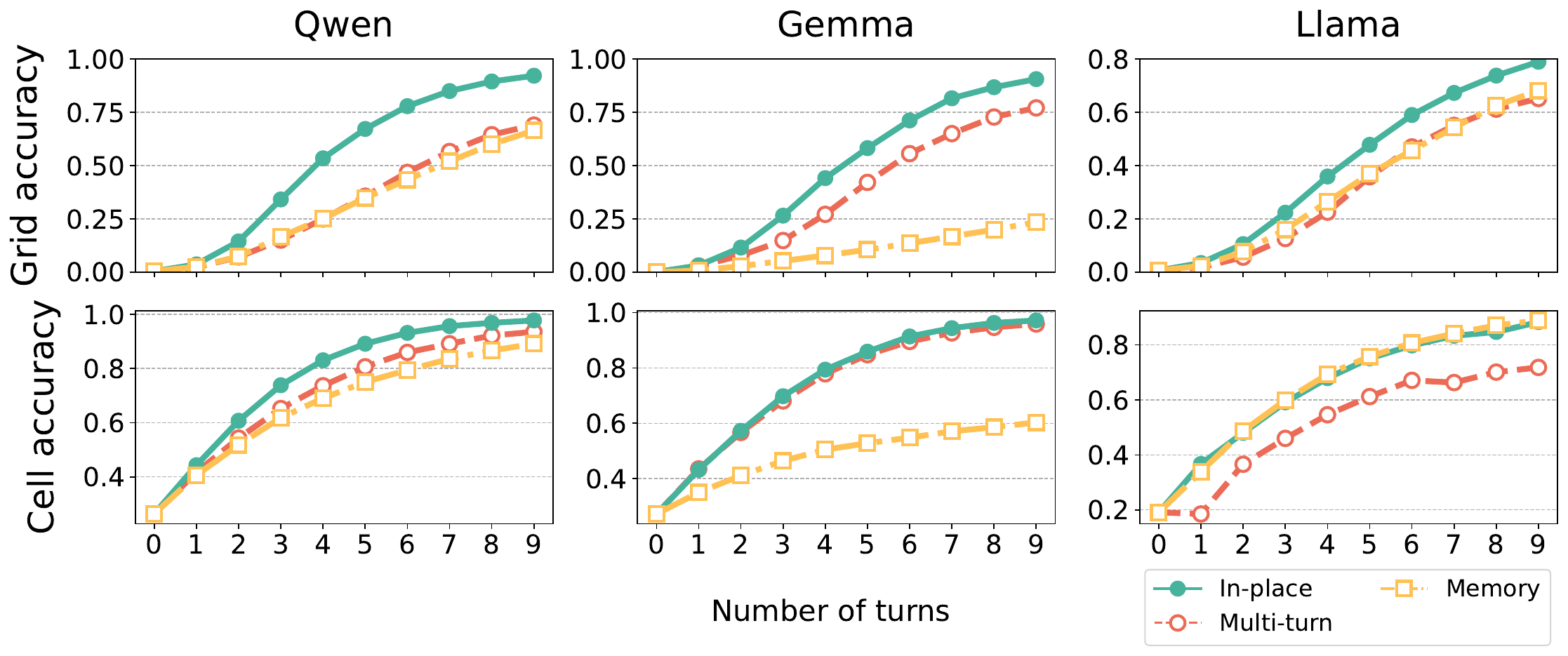}
    \caption{Grid and cell accuracy on the ZebraLogic dataset. Top-$4$ feedback is used.}
    \label{fig:zebra_acc_memory_top4}
\end{figure}

\begin{figure}[H]
    \centering
    \includegraphics[width=\linewidth]{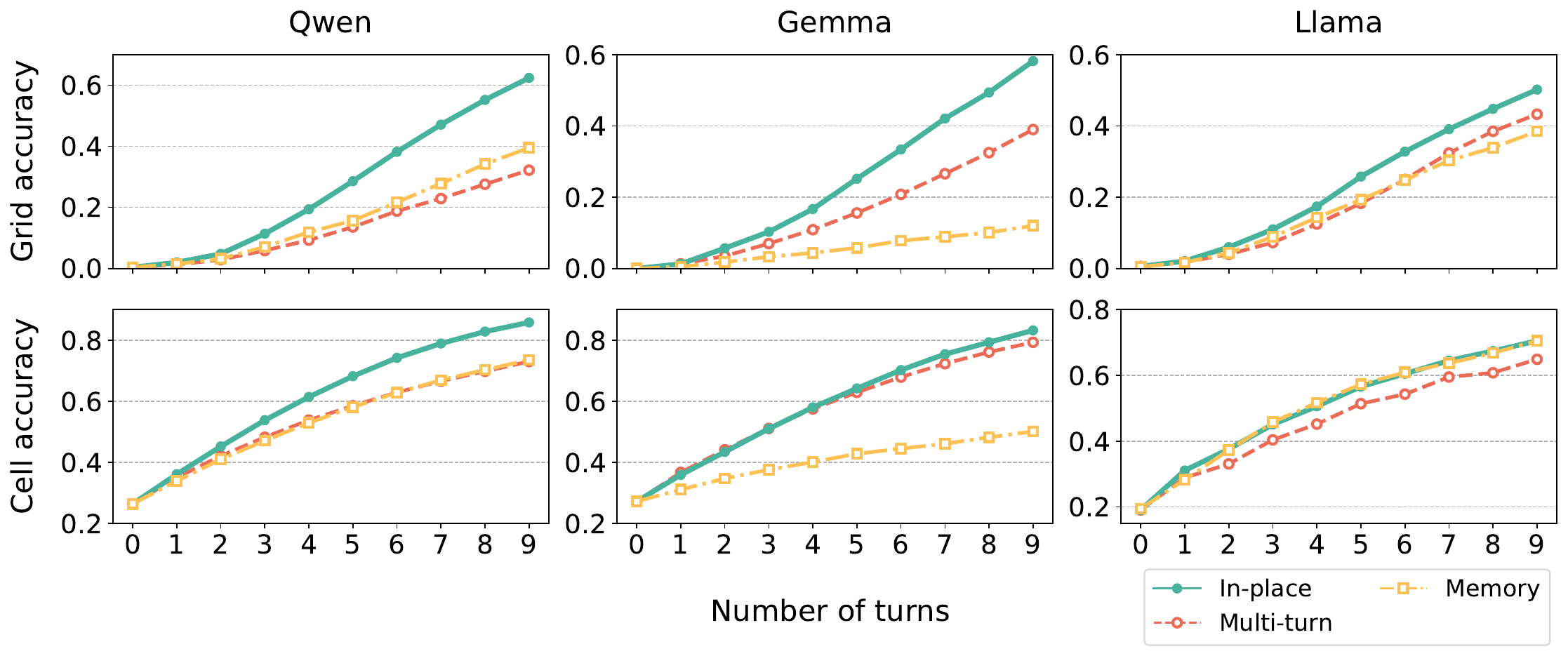}
    \caption{Grid and cell accuracy on the ZebraLogic dataset. Top-$2$ feedback is used.}
    \label{fig:zebra_acc_memory_top2}
\end{figure}

\begin{figure}[H]
    \centering
    \includegraphics[width=\linewidth]{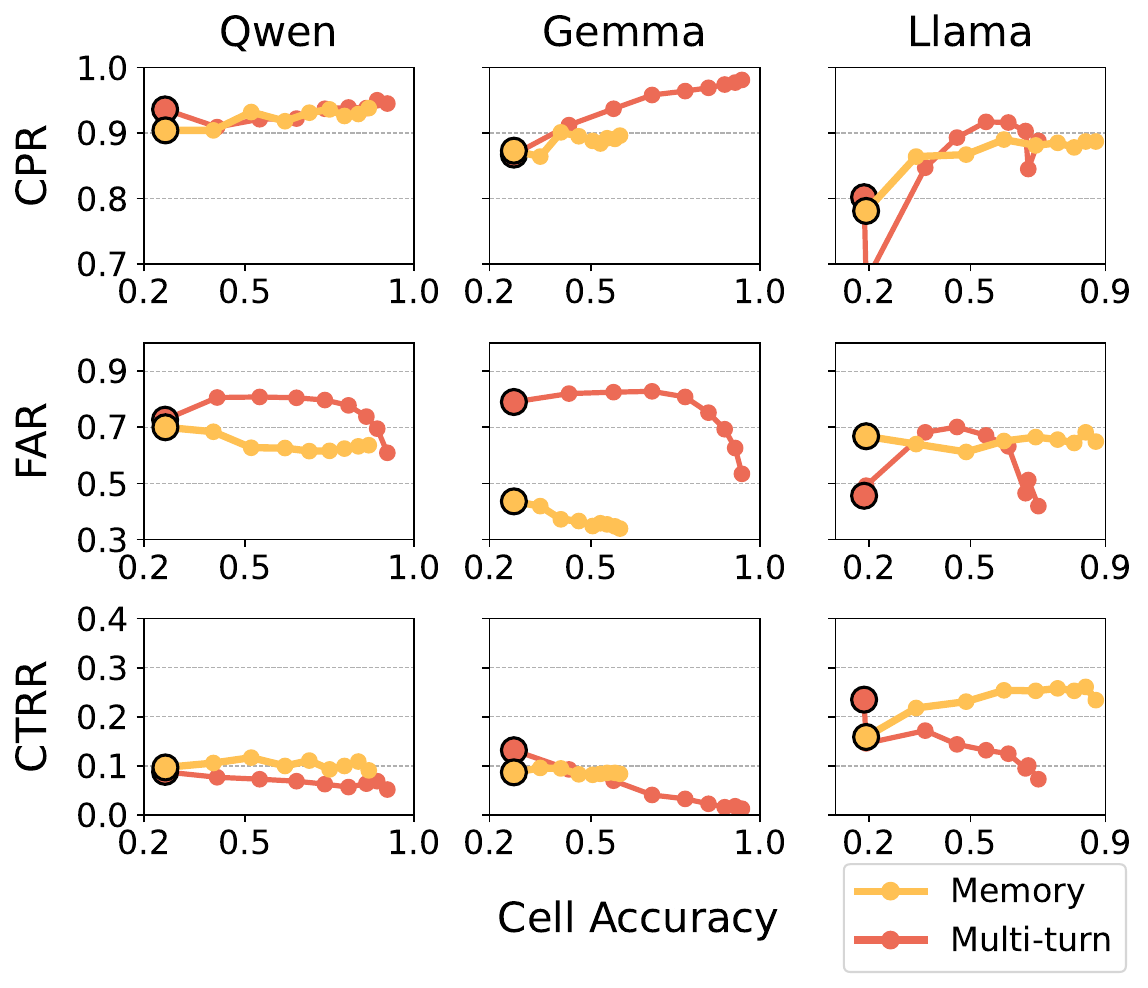}
    \caption{CPR, FAR, and CTRR for 10-turn conversations of LLMs on the ZebraLogic dataset. Top-$4$ feedback is used. The points with a black border represent the second response of the LLMs (i.e., $y_1$), and the subsequent responses across turns are connected by lines.}
    \label{fig:app_zebra_memory_top4}
\end{figure}

\begin{figure}[H]
    \centering
    \includegraphics[width=\linewidth]{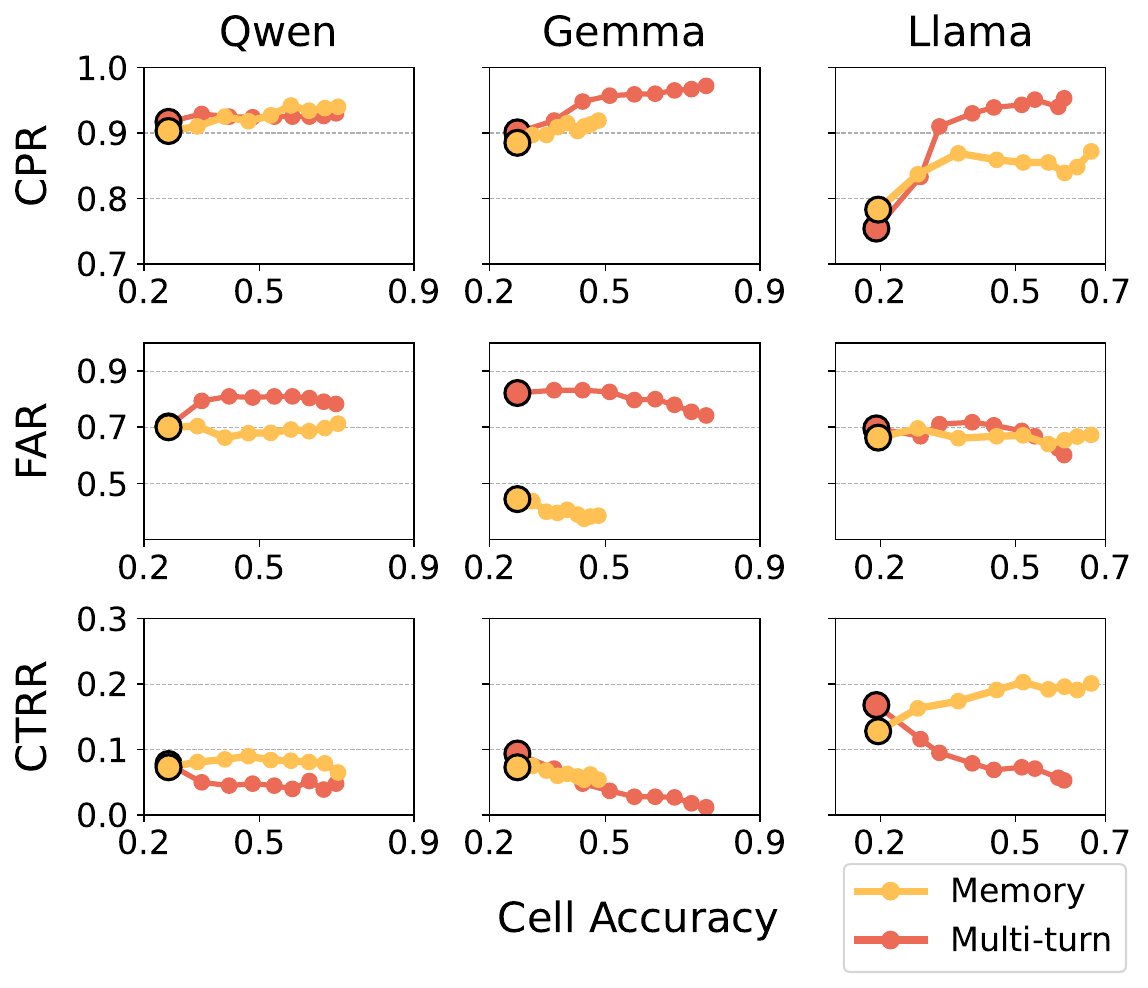}
    \caption{CPR, FAR, and CTRR for 10-turn conversations of LLMs on the ZebraLogic dataset. Top-$2$ feedback is used. The points with a black border represent the second response of the LLMs (i.e., $y_1$), and the subsequent responses across turns are connected by lines.}
    \label{fig:app_zebra_memory_top2}
\end{figure}

\cref{fig:zebra_acc_memory_top4,fig:zebra_acc_memory_top2} show performance of methods on grid show grid accuracy and cell accuracy, respectively. Memory-based feedback performs the worst among the three methods. \cref{fig:app_zebra_memory_top4,fig:app_zebra_memory_top2} show that, compared with multi-turn feedback, memory-based feedback has smaller declines in FAR and CTRR as turns evolve. This pattern may partly reflect its lower overall performance. However, our controlled experiments in \cref{sec:effect_history} further indicate that the accumulation of dialogue history can contribute to the decreases in FAR and CTRR observed under multi-turn feedback.

\begin{figure}[H]
    \centering
    \includegraphics[width=\columnwidth]{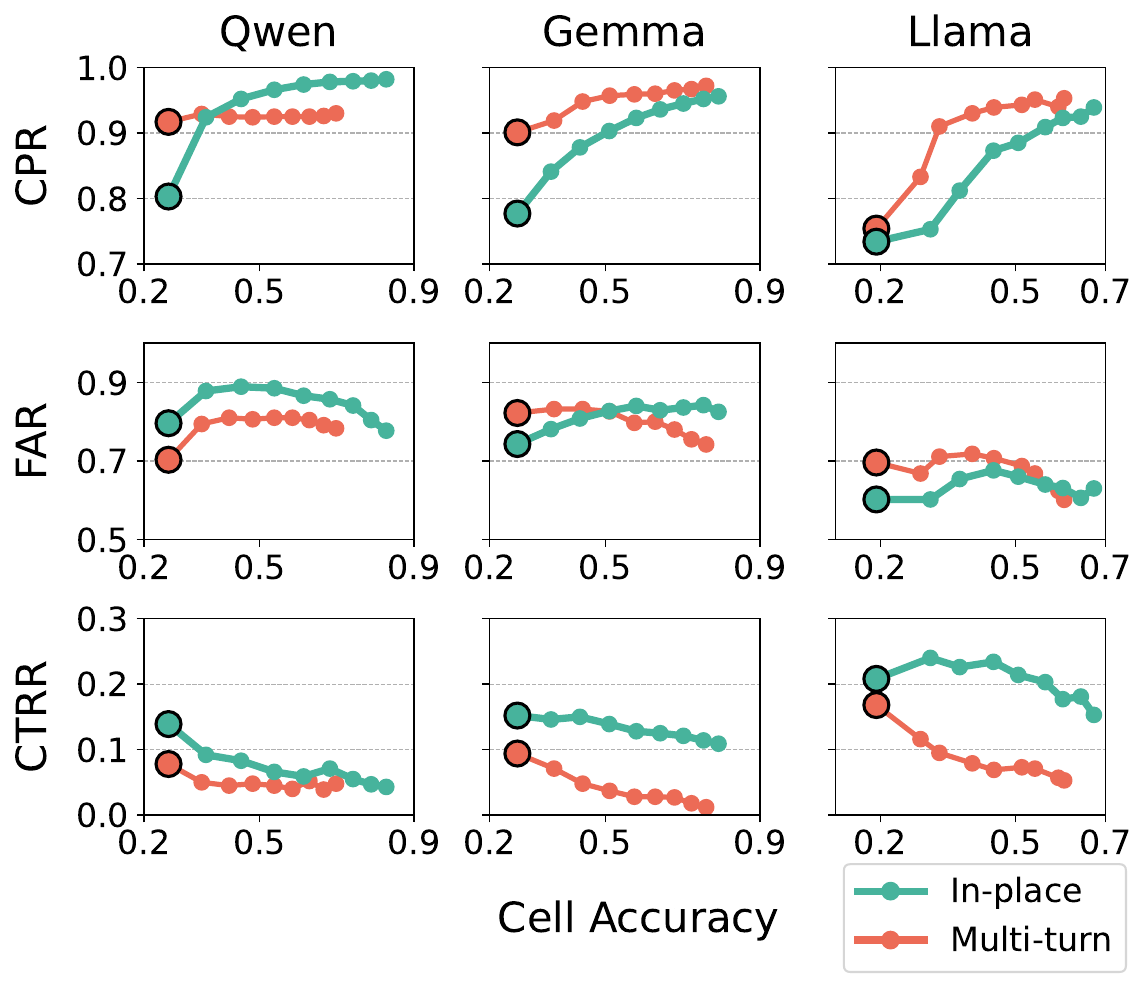}
    % \vskip -0.6em
    \caption{CPR, FAR, and CTRR for 10-turn with top-2 feedback. Filled points denote in-place feedback, and hollow points denote multi-turn feedback. The points with a black border represent the second response of the LLMs (i.e., $y_1$), and the subsequent responses across turns are connected by lines.}
    % \vskip -1 em
    \label{fig:zebra_reasoning_top2}
\end{figure}

\subsection{Performance under Oracle Feedback}
\label{app.zebralogic_oracle_feedback}

\begin{figure}[H]
    \centering
    \includegraphics[width=1.0\linewidth]{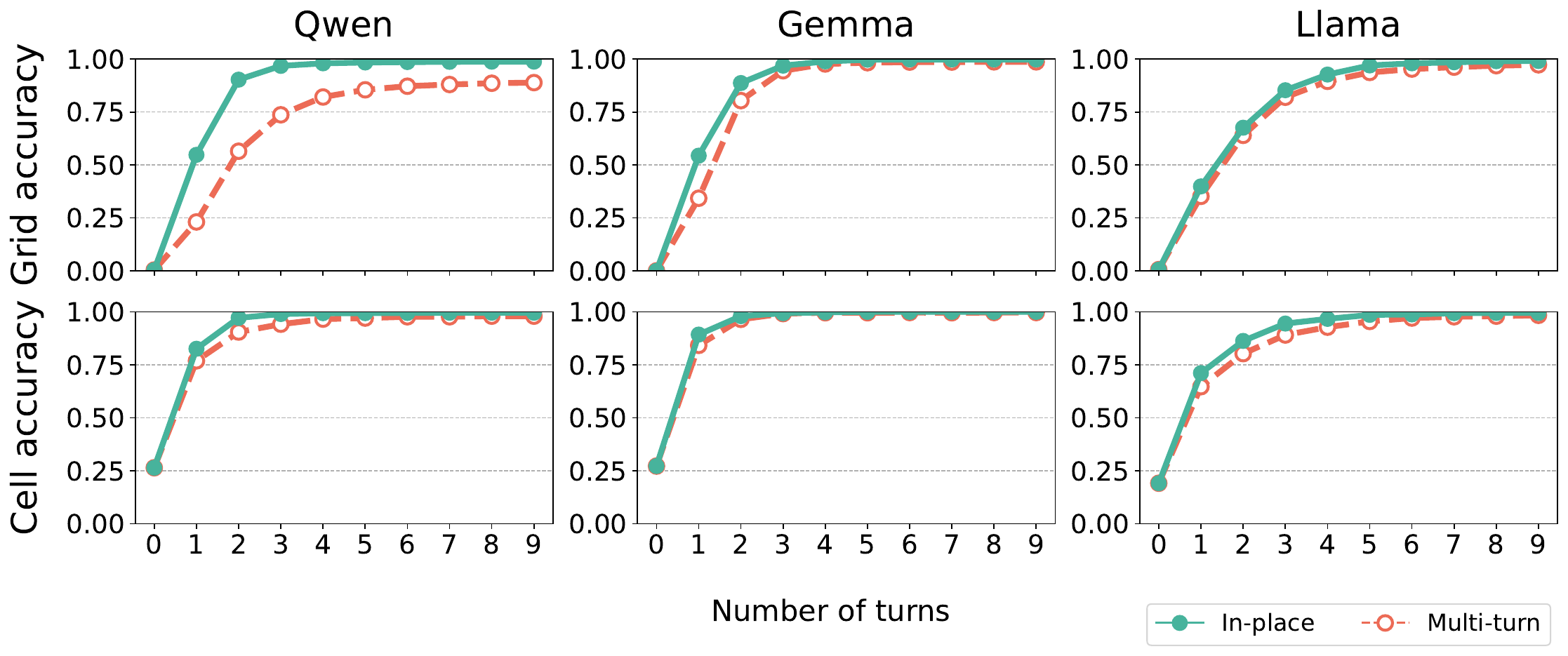}
    \caption{Grid and cell accuracy of LLMs on the ZebraLogic dataset. In-place feedback consistently outperforms multi-turn feedback under an oracle setting.}
    \label{fig:zebra_acc_oracle}
\end{figure}

\begin{figure}[H]
    \centering
    \includegraphics[width=1.0\linewidth]{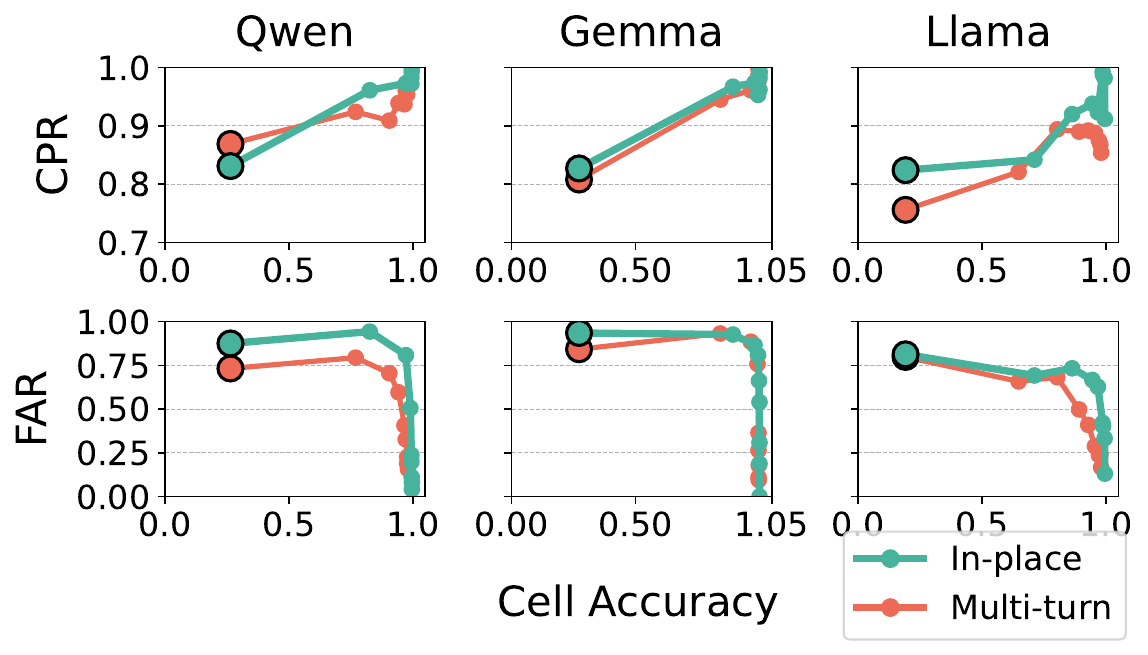}
    \caption{CPR, FAR, and CTRR for 10-turn conversations of LLMs on the ZebraLogic dataset. The oracle feedback function is used. Filled points denote in-place feedback, and hollow points denote multi-turn feedback. The points with a black border represent the second response of the LLMs (i.e., $y_1$), and the subsequent responses across turns are connected by lines.}
    \label{fig:zebra_analysis_oracle}
\end{figure}

In addition to top-$k$ feedback, we conduct an experiment with \emph{oracle feedback}, in which feedback is provided for every incorrect cell rather than only the top-$k$ most critical ones. \cref{fig:zebra_acc_oracle} presents grid and cell accuracy as a function of the number of turns. In-place feedback converges substantially faster than multi-turn feedback, indicating that it enables the model to incorporate feedback more efficiently. \cref{fig:zebra_analysis_oracle} illustrates the turn-level behavior of LLMs under oracle feedback. In most cases, in-place feedback achieves higher CPR and FAR than multi-turn feedback; the only exception is the CPR of \texttt{Qwen} in the first turn.

% \subsection{Extended Analysis on Dialog History}
% \label{app.extend_analysis}

% We extend the discussion in \cref{sec:effect_history}.

% \begin{observation}
%     FAR trends under multi-turn feedback vary across models, unlike the uniform trends observed in CPR and CTRR.
% \end{observation}

% \texttt{Qwen} maintains consistently high FAR, \texttt{Gemma} shows diminishing gains in later turns, and \texttt{Llama} benefits initially but reverses in later turns. Since FAR is closely tied to instruction-following capabilities, which are heavily shaped by model-specific post-training, we attribute these patterns to a complex interplay between post-training differences and shared mechanistic inductive biases. 

% In-place feedback sidesteps this heterogeneity, reaching the highest FAR across all three models in later turns (\Cref{fig:zebra_reasoning}). We attribute this to its mechanism: in-place feedback places the correction directly into the response, rather than sending it as a separate instruction for the model to interpret and apply. This removes the need to treat feedback as an external command, yielding consistently high FAR regardless of model-specific instruction-following behavior.

\section{Robustness to Imperfect Feedback}
\label{app:robustness}

To assess the robustness of in-place feedback under realistic conditions, we conduct two complementary analyses of the quality of (i)~the feedback content and (ii)~the editing process.

\subsection{Incomplete Feedback} 

We examine whether the advantage of in-place feedback persists when the feedback itself is incomplete. In this setting, feedback identifies what is wrong but does not provide the information needed to directly correct it, requiring the model to infer the appropriate revision on its own.

We first construct a controlled experiment on ZebraLogic. Specifically, the standard top-$k$ feedback template \emph{``$\langle$category$\rangle$ of $\langle$house$\rangle$ is $\langle$gt$\rangle$, not $\langle$pred$\rangle$''} is replaced with \emph{``$\langle$category$\rangle$ of $\langle$house$\rangle$ is not $\langle$pred$\rangle$''}. We provide the top-4 incomplete feedback signals per instance and run 10~turns with \texttt{Qwen-2.5-7B-Instruct} and \texttt{Gemma-3-4B-it}.

\cref{tab:incomplete_grid,tab:incomplete_cell} report grid and cell accuracy. As expected, absolute performance is substantially lower than under complete feedback because the model must independently derive the correct value rather than simply incorporate it from the feedback. Nevertheless, in-place feedback consistently outperforms multi-turn feedback across both models and both metrics. This suggests that removing the erroneous span from the conditioning context remains beneficial even when the correction itself must be inferred.

To test whether this effect is specific to the structured ZebraLogic setting, we further extend the incomplete-feedback ablation to MATH-hard and GPQA using \texttt{Qwen-2.5-7B-Instruct} and \texttt{Gemma-3-4B-it}. Here, the feedback identifies the erroneous part of the previous response but does not specify how it should be fixed. As shown in \cref{tab:rebuttal-incomplete-feedback-math,tab:rebuttal-incomplete-feedback-gpqa}, in-place feedback again consistently outperforms multi-turn feedback under this weaker signal.

Across both the structured ZebraLogic setting and the more open-ended reasoning tasks, the same pattern emerges. Multi-turn feedback retains the erroneous response in the dialogue history, which can anchor subsequent reasoning to the model's prior mistake when the correct revision is not explicitly supplied. In-place feedback instead removes the erroneous span and lets the model continue from a cleaner state. Thus, its advantage does not depend on feedback explicitly specifying the correction, but persists even when the model must infer the fix through subsequent reasoning.

\begin{table}[h]
\centering
% \small
% \setlength{\tabcolsep}{4pt}
\resizebox{\linewidth}{!}{
\begin{tabular}{ll cccccccccc}
\toprule
Model & Method & 0 & 1 & 2 & 3 & 4 & 5 & 6 & 7 & 8 & 9 \\
\midrule
\multirow{2}{*}{Qwen}  & Multi-turn & 0.005 & 0.009 & 0.010 & 0.014 & 0.016 & 0.017 & 0.017 & 0.019 & 0.021 & 0.024 \\
                       & In-place   & 0.005 & 0.009 & \textbf{0.017} & \textbf{0.028} & \textbf{0.037} & \textbf{0.049} & \textbf{0.059} & \textbf{0.068} & \textbf{0.079} & \textbf{0.086} \\
\midrule
\multirow{2}{*}{Gemma} & Multi-turn & 0.000 & 0.002 & 0.003 & 0.005 & 0.009 & 0.010 & 0.017 & 0.019 & 0.021 & 0.023 \\
                       & In-place   & 0.000 & \textbf{0.009} & \textbf{0.018} & \textbf{0.034} & \textbf{0.042} & \textbf{0.044} & \textbf{0.055} & \textbf{0.067} & \textbf{0.076} & \textbf{0.090} \\
% \midrule
% \multirow{2}{*}{Llama} & Multi-turn & 0.007 & 0.014 & 0.017 & 0.019 & 0.021 & 0.026 & 0.031 & 0.033 & 0.037 & 0.038 \\
%                        & In-place   & 0.007 & \textbf{0.032} & \textbf{0.043} & \textbf{0.053} & \textbf{0.059} & \textbf{0.069} & \textbf{0.085} & \textbf{0.090} & \textbf{0.106} & \textbf{0.106} \\
\bottomrule
\end{tabular}
}
\caption{Grid accuracy on ZebraLogic under \emph{incomplete} top-4 feedback. Feedback indicates only that a cell is incorrect, without providing the correct value.}
\label{tab:incomplete_grid}
\end{table}

\begin{table}[h]
\centering
% \small
% \setlength{\tabcolsep}{4pt}
\resizebox{\linewidth}{!}{
\begin{tabular}{ll cccccccccc}
\toprule
Model & Method & 0 & 1 & 2 & 3 & 4 & 5 & 6 & 7 & 8 & 9 \\
\midrule
\multirow{2}{*}{Qwen}  & Multi-turn & 0.274 & 0.284 & 0.290 & 0.296 & 0.295 & 0.297 & 0.296 & 0.299 & 0.300 & 0.299 \\
                       & In-place   & 0.274 & \textbf{0.304} & \textbf{0.341} & \textbf{0.372} & \textbf{0.395} & \textbf{0.414} & \textbf{0.434} & \textbf{0.444} & \textbf{0.453} & \textbf{0.449} \\
\midrule
\multirow{2}{*}{Gemma} & Multi-turn & 0.273 & 0.283 & 0.307 & 0.319 & 0.333 & 0.344 & 0.346 & 0.346 & 0.354 & 0.361 \\
                       & In-place   & 0.273 & \textbf{0.309} & \textbf{0.336} & \textbf{0.371} & \textbf{0.392} & \textbf{0.414} & \textbf{0.424} & \textbf{0.449} & \textbf{0.461} & \textbf{0.472} \\
% \midrule
% \multirow{2}{*}{Llama} & Multi-turn & 0.188 & \textbf{0.208} & 0.218 & 0.234 & 0.238 & 0.254 & 0.292 & 0.312 & 0.324 & 0.338 \\
%                        & In-place   & 0.188 & 0.169 & \textbf{0.315} & \textbf{0.358} & \textbf{0.394} & \textbf{0.416} & \textbf{0.438} & \textbf{0.453} & \textbf{0.454} & \textbf{0.457} \\
\bottomrule
\end{tabular}
}
\caption{Cell accuracy on ZebraLogic under \emph{incomplete} top-4 feedback.}
\label{tab:incomplete_cell}
\end{table}

\begin{table}[h]
    \centering
    \resizebox{\linewidth}{!}{%
    \begin{tabular}{ll cccccccccc}
        \toprule
        \multirow{2}{*}{Model} & \multirow{2}{*}{Method} & \multicolumn{10}{c}{Number of turns} \\
        \cmidrule(lr){3-12}
        & & 0 & 1 & 2 & 3 & 4 & 5 & 6 & 7 & 8 & 9 \\
        \midrule
        \multirow{2}{*}{Qwen}
            & Multi-turn & 0.550 & 0.596 & 0.614 & 0.622 & 0.624 & 0.626 & 0.632 & 0.634 & 0.634 & 0.634 \\
            & In-place   & 0.550 & 0.606 & 0.632 & 0.646 & 0.650 & 0.660 & 0.666 & 0.668 & 0.676 & \textbf{0.686} \\
        \midrule
        \multirow{2}{*}{Gemma}
            & Multi-turn & 0.492 & 0.566 & 0.610 & 0.630 & 0.646 & 0.650 & 0.658 & 0.664 & 0.670 & 0.670 \\
            & In-place   & 0.492 & 0.600 & 0.638 & 0.658 & 0.684 & 0.694 & 0.698 & 0.710 & 0.714 & \textbf{0.722} \\
        \bottomrule
    \end{tabular}%
    }
    \caption{Incomplete-feedback ablation on MATH-hard. Feedback only identifies the erroneous part of the previous response, without specifying how to correct it.}
    \label{tab:rebuttal-incomplete-feedback-math}
\end{table}

\begin{table}[h]
    \centering
    \resizebox{\linewidth}{!}{%
    \begin{tabular}{ll cccccccccc}
        \toprule
        \multirow{2}{*}{Model} & \multirow{2}{*}{Method} & \multicolumn{10}{c}{Number of turns} \\
        \cmidrule(lr){3-12}
        & & 0 & 1 & 2 & 3 & 4 & 5 & 6 & 7 & 8 & 9 \\
        \midrule
        \multirow{2}{*}{Qwen}
            & Multi-turn & 0.103 & 0.119 & 0.119 & 0.143 & 0.143 & 0.151 & 0.151 & 0.151 & 0.159 & 0.159 \\
            & In-place   & 0.103 & 0.119 & 0.119 & 0.151 & 0.191 & 0.198 & 0.214 & 0.230 & 0.230 & \textbf{0.230} \\
        \midrule
        \multirow{2}{*}{Gemma}
            & Multi-turn & 0.056 & 0.079 & 0.079 & 0.087 & 0.103 & 0.103 & 0.103 & 0.103 & 0.111 & 0.111 \\
            & In-place   & 0.056 & 0.087 & 0.087 & 0.111 & 0.127 & 0.127 & 0.135 & 0.143 & 0.143 & \textbf{0.151} \\
        \bottomrule
    \end{tabular}%
    }
    \caption{Incomplete-feedback ablation on GPQA. Feedback only identifies the erroneous part of the previous response, without specifying how to correct it.}
    \label{tab:rebuttal-incomplete-feedback-gpqa}
\end{table}

\subsection{Reliability of the In-place Agent}

\begin{table*}[t]
\centering
\small
\begin{tabular}{ll cc l}
\toprule
Category & Sub-type & MATH (\%) & Zebra (\%) & Description \\
\midrule
Appropriate     & --              & 40.6 & 53.9 & Correct localization, minimal fix, proper pruning. \\
\midrule
\multirow{4}{*}{Incomplete}
                & Over-prune      & 18.2 &  0.7 & Valid reasoning discarded beyond what feedback necessitated. \\
                & Under-prune     &  3.8 & 39.9 & Erroneous downstream content retained. \\
                & Partial address & 33.4 & --   & Feedback partially incorporated. \\
                & Misapplied      &  2.4 &  4.2 & Feedback misunderstood or not reflected in the edit. \\
\midrule
Excessive       & --              &  1.6 &  1.3 & Rewriting valid content beyond feedback scope. \\
\bottomrule
\end{tabular}
\caption{Reliability analysis of the in-place agent on MATH-Hard (Gemma-3-4b-it) and ZebraLogic (top-4 feedback), evaluated by Claude~4.5~Sonnet as LLM-judge. Most edits are appropriate or incomplete; aggressive over-rewriting (excessive) is rare.}
\label{tab:agent_reliability}
\end{table*}

A second source of imperfection is the in-place agent itself, which must localize the conflicting span, apply a minimal edit, and prune dependent reasoning. To analyze its reliability, we classify edits 
from MATH-hard and ZebraLogic (top-4 feedback) with Gemma-3-4b-it, using Claude~4.5~Sonnet as an LLM-judge according to the taxonomy in \cref{tab:agent_reliability}.

The agent produces appropriate edits in only 40.6\% (MATH-hard) and 53.9\% (ZebraLogic) of cases. The dominant failure mode is incomplete repair: on MATH-hard, the agent often partially addresses the feedback (33.4\%) or over-prunes (18.2\%); on ZebraLogic, under-pruning is most common (39.9\%), reflecting the parallel reasoning structure in which dependencies between cells are harder to trace. Excessive rewriting is rare ($\leq$1.6\%) on both datasets. 
This is the failure mode most likely to inflate performance through unintended content changes, so its rarity suggests that the gains reported in \cref{sec:experimental_results,sec:analysis} are not artifacts of the agent rewriting beyond the targeted span.

\paragraph{Takeaway.} Together, these analyses show that in-place feedback is robust to imperfection from \emph{both} the feedback content (incomplete signals) and the editing process (imperfect agent edits). Both factors degrade in-place feedback, yet it consistently outperforms multi-turn feedback. We therefore expect the gap to be \emph{larger} in real expert-LLM collaboration, where human edits are typically cleaner and more targeted than those produced by an automatic agent.

\section{Extended Related Work}
\label{app:related_work}

\paragraph{Multi-turn interaction of LLMs.}
Several studies aim to improve the performance of LLMs in multi-turn interaction. One line of work focuses on clarifying questions, where the LLM generates follow-up questions when the user's input is ambiguous~\citep{zhang2023clarify,zamani2020generating,aliannejadi2019asking}. \citet{zhang2023clarify} proposes a framework that integrates clarifying questions into the response generation process. Another line of work enhances multi-turn performance through additional training, often employing a user simulation model to generate multi-turn interaction scenarios~\citep{zhou2024archer, shani2024multi, zhang2024modeling, wu2025collabllm}. For example, \citet{wu2025collabllm} fine-tunes LLMs with reinforcement learning to improve user interactivity. Our work focuses on how LLMs can achieve more effective interaction with users without additional training, while ensuring token efficiency.

\paragraph{Refinement of LLMs.} 
Recent research explores self-refinement, an approach in which LLMs generate feedback on their own outputs and improve them accordingly~\citep{madaan2023self, dhuliawala2024chain, shinn2023reflexion, nathani2023maf}. Self-Refine~\citep{madaan2023self}, for instance, iterates a critique-then-revise loop in which the model regenerates the full response at each step while previous responses remain in the dialogue history. \citet{huang2024large} shows that LLMs often fail to self-correct reasoning errors without external signals. In response, \citet{welleck2022generating} trains a separate model to produce feedback, and \citet{han2025convcodeworld} uses an external LLM agent to provide feedback for evaluating model performance. A complementary line of work modifies the prompt or external memory based on user feedback to avoid recurring mistakes. MemPrompt~\citep{madaan2022memory} stores user clarifications in a memory and retrieves them to rewrite future prompts, and MemGPT~\citep{packer2023memgpt} selectively manages context by summarizing or discarding older interactions to fit a bounded working memory. These approaches differ from ours along two axes. First, on the input side, the erroneous response itself is retained, either verbatim in the dialogue history (Self-Refine) or in summarized form in memory (MemPrompt, MemGPT), so the model still conditions on incorrect outputs when producing its next response. Second, on the output side, all of these methods regenerate the entire response in every iteration, which is token-inefficient and can introduce new errors in spans that were already correct. In contrast, our method removes the erroneous span from the conditioning context and generates only the continuation after the corrected point, so that both the conditioning input and the regenerated output are restricted to the region affected by the error.

\paragraph{Analysis of multi-turn conversations.}
Recent studies analyze the performance of LLMs in multi-turn conversations. \citet{jiang2025feedback} shows that LLMs fail to reliably incorporate feedback, even when it is close to the correct answer, highlighting the need for stronger correction mechanisms. \citet{laban2025llms} finds that accuracy decreases when a single problem is divided into multiple parts and solved by LLMs in a multi-turn manner. \citet{deshpande2025multichallenge} analyzes the performance of LLMs in multi-turn conversations across four categories, including instruction retention and self-coherence. While these studies analyze the performance of LLMs in the multi-turn setting, they primarily report high-level metrics, such as overall task accuracy, without analyzing how reasoning evolves across turns or how errors propagate. Our turn-level diagnostic metrics (CPR, FAR, CTRR) directly address this gap by quantifying when and how erroneous context degrades subsequent reasoning.

\paragraph{UI-level interaction systems.}
Prior HCI research has investigated UI-level systems that provide localized, in-place interventions when interacting with LLMs~\citep{pang2025interactivereasoningvisualizingcontrolling, Masson_2024, Xie2024wait}. These systems allow users to directly edit problematic spans or to monitor intermediate reasoning steps and execution traces to diagnose and guide multi-step behavior. This line of work shows that UI-level support for localized, in-place editing helps users work more efficiently and with lower cognitive load, since they can precisely identify and correct errors. While these systems offer locality at the interaction level, at the implementation level, generation is still conditioned on the entire prior response, so locality in the UI does not translate into locality in the conditioning context. Motivated by these findings, we extend the idea of localized editing from the UI level to the model level, so that the corrected span replaces the original output in the actual context used for continuation.

\section{Performance of Large-Scale LLMs}
\label{app.large_scale_llms}

In this section, we present the results obtained with large-scale LLMs. We use two open-source large-scale LLMs, such as \texttt{Llama-3.1-70B-Instruct} and \texttt{Gemma-3-27b-it}.

\subsection{Task Performance}
\label{app.large_scale_llms_task_performance}

\begin{figure}[H]
    \centering
    \includegraphics[width=1.0\linewidth]{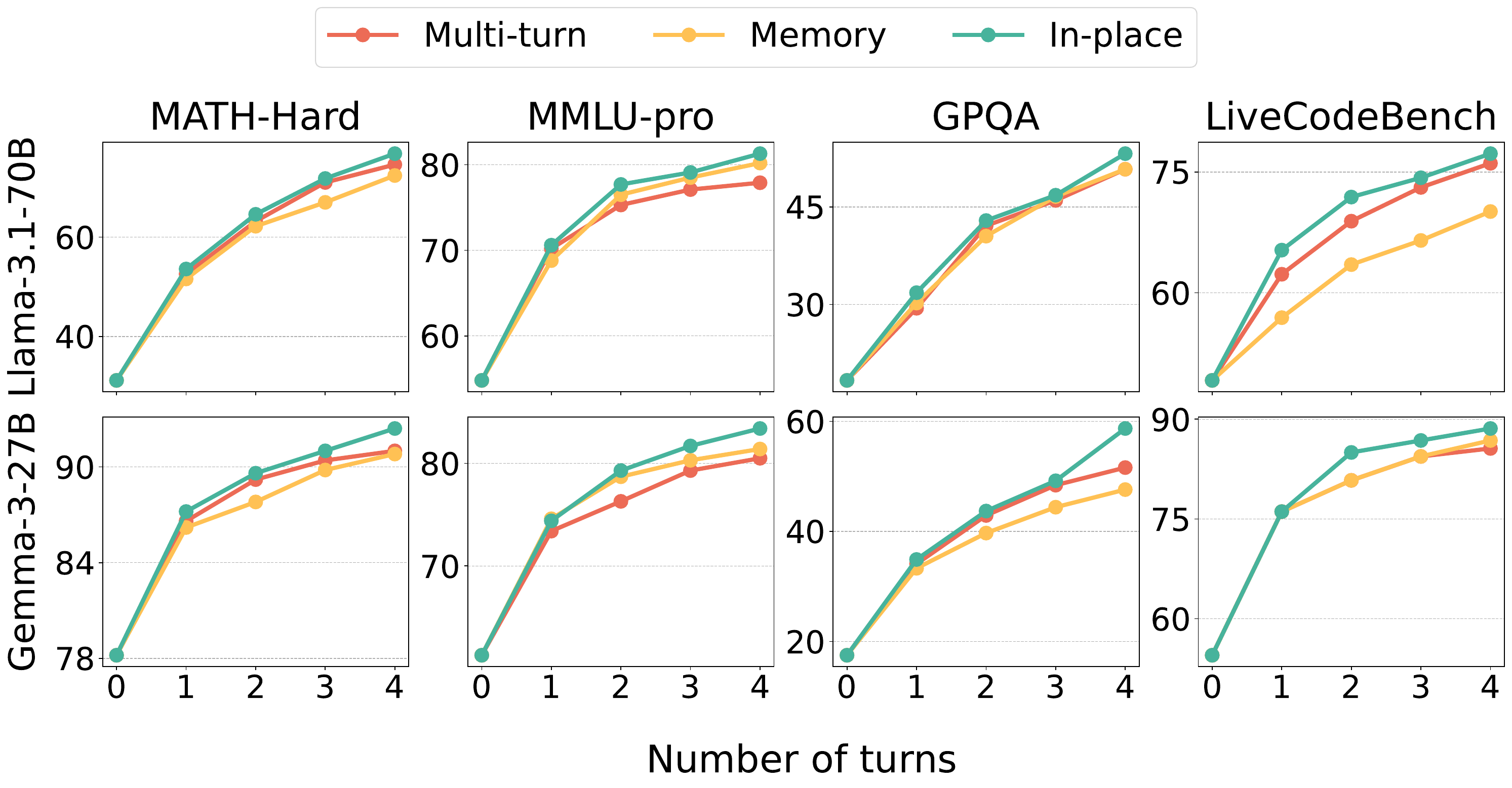}
    \caption{Performance comparison in MATH-hard, MMLU-pro, GPQA, and LiveCodeBench. Across all datasets and large-scale LLM models, in-place feedback consistently outperforms the multi-turn and memory-based feedback approaches. Tabulated results are provided in \cref{tab:app_performance_large_scale_MATH_Hard,tab:app_performance_large_scale_MMLU_Pro,tab:app_performance_large_scale_GPQA,tab:app_performance_large_scale_code}.}
    \label{fig:large_results_combined}
\end{figure}

\cref{fig:large_results_combined} shows the task performance, which follows a similar trend to \cref{fig:main_results}. While the gap between baselines and in-place feedback is smaller, this is largely because the absolute performance is already high, leaving limited room for improvement. Additionally, these LLMs have been extensively tuned via instruction-tuning to follow user instructions effectively, thereby benefiting baselines. Exploring whether dedicated post-training can further amplify the advantages of in-place feedback remains an interesting direction for future work. \cref{app.input_token_len_large_scale,app.gen_token_len_large_scale} show the input and generated token counts, respectively, confirming that in-place feedback maintains its token efficiency at larger scales. Overall, these results demonstrate that in-place feedback achieves the highest task performance and is the most token-efficient, even in large models.

\begin{table}[H]
\centering
\resizebox{\linewidth}{!}{
\begin{tabular}{llccccc}
\toprule
\multirow{2}{*}{\textbf{Model}} 
& \multirow{2}{*}{\textbf{Method}} 
& \multicolumn{5}{c}{\textbf{Number of turns}} \\
\cmidrule{3-7}
& & \textbf{0} & \textbf{1} & \textbf{2} & \textbf{3} & \textbf{4} \\
\midrule

% Llama 70B
\multirow{3}{*}{Llama-3.1-70B}
& Multi-turn & 31.2 & 52.6 & 63.2 & 71.0 & 74.6 \\
& Memory  & 31.2 & 51.6 & 62.2 & 67.0 & 72.4 \\
& In-place   & 31.2 & 53.6 & 64.6 & 71.8 & 76.8 \\
\midrule

% Gemma 27B
\multirow{3}{*}{Gemma-3-27B}
& Multi-turn & 78.2 & 86.6 & 89.2 & 90.4 & 91.0 \\
& Memory  & 78.2 & 86.2 & 87.8 & 89.8 & 90.8 \\
& In-place   & 78.2 & 87.2 & 89.6 & 91.0 & 92.4 \\
\bottomrule

\end{tabular}
}
\caption{Performance comparison on MATH-hard.}
\label{tab:app_performance_large_scale_MATH_Hard}
\end{table}

\begin{table}[H]
\centering
\resizebox{\linewidth}{!}{
\begin{tabular}{llccccc}
\toprule
\multirow{2}{*}{\textbf{Model}}
& \multirow{2}{*}{\textbf{Method}}
& \multicolumn{5}{c}{\textbf{Number of turns}} \\
\cmidrule{3-7}
& & \textbf{0} & \textbf{1} & \textbf{2} & \textbf{3} & \textbf{4} \\
\midrule

% Llama-70b
\multirow{3}{*}{Llama-3.1-70B}
& Multi-turn & 54.8 & 70.2 & 75.3 & 77.1 & 77.9 \\
& Memory  & 54.8 & 68.8 & 76.5 & 78.5 & 80.2 \\
& In-place   & 54.8 & 70.6 & 77.7 & 79.1 & 81.3 \\
\midrule

% Gemma-27b
\multirow{3}{*}{Gemma-3-27B}
& Multi-turn & 61.3 & 73.4 & 76.3 & 79.3 & 80.5 \\
& Memory  & 61.3 & 74.6 & 78.7 & 80.3 & 81.4 \\
& In-place   & 61.3 & 74.4 & 79.3 & 81.7 & 83.4 \\
\bottomrule

\end{tabular}
}

\caption{Performance comparison MMLU-pro.}
\label{tab:app_performance_large_scale_MMLU_Pro}
\end{table}

\begin{table}[H]
\centering
\resizebox{\linewidth}{!}{
\begin{tabular}{llccccc}
\toprule
\multirow{2}{*}{\textbf{Model}}
& \multirow{2}{*}{\textbf{Method}}
& \multicolumn{5}{c}{\textbf{Number of turns}} \\
\cmidrule{3-7}
& & \textbf{0} & \textbf{1} & \textbf{2} & \textbf{3} & \textbf{4} \\
\midrule

% Llama 70B
\multirow{3}{*}{Llama-3.1-70B}
& Multi-turn & 18.3 & 29.4 & 42.1 & 46.0 & 50.8 \\
& Memory  & 18.3 & 30.2 & 40.5 & 46.6 & 50.8 \\
& In-place   & 18.3 & 31.8 & 42.9 & 46.8 & 53.2 \\
\midrule

% Gemma 27B
\multirow{3}{*}{Gemma-3-27B}
& Multi-turn & 17.5 & 34.1 & 42.9 & 48.4 & 51.6 \\
& Memory  & 17.5 & 33.3 & 39.7 & 44.4 & 47.6 \\
& In-place   & 17.5 & 34.9 & 43.7 & 49.2 & 58.7 \\
\bottomrule

\end{tabular}
}

\caption{Performance comparison on GPQA.}
\label{tab:app_performance_large_scale_GPQA}
\end{table}

\begin{table}[H]
\centering
\resizebox{\linewidth}{!}{
\begin{tabular}{llccccc}
\toprule
\multirow{2}{*}{\textbf{Model}}
& \multirow{2}{*}{\textbf{Method}}
& \multicolumn{5}{c}{\textbf{Number of turns}} \\
\cmidrule{3-7}
& & \textbf{0} & \textbf{1} & \textbf{2} & \textbf{3} & \textbf{4} \\
\midrule

% Llama 70B
\multirow{3}{*}{Llama-3.1-70B}
& Multi-turn & 49.1 & 62.3 & 68.9 & 73.1 & 76.1 \\
& Memory     & 49.1 & 56.9 & 63.5 & 66.5 & 70.1 \\
& In-place   & 49.1 & 65.3 & 71.9 & 74.3 & 77.3 \\
\midrule

% Gemma 27B
\multirow{3}{*}{Gemma-3-27B}
& Multi-turn & 54.5 & 76.1 & 80.8 & 84.4 & 85.6 \\
& Memory     & 54.5 & 76.1 & 80.8 & 84.4 & 86.8 \\
& In-place   & 54.5 & 76.1 & 85.0 & 86.8 & 88.6 \\
\bottomrule

\end{tabular}
}

\caption{Performance comparison on LiveCodeBench.}
\label{tab:app_performance_large_scale_code}
\end{table}

\subsection{Number of Input Tokens}
\label{app.input_token_len_large_scale}

\begin{table}[H]
\centering
\resizebox{\linewidth}{!}{
\begin{tabular}{llrrrrr}
\toprule
\multirow{2}{*}{\textbf{Model}}
& \multirow{2}{*}{\textbf{Method}}
& \multicolumn{5}{c}{\textbf{Number of turns}} \\
\cmidrule{3-7}
& & \textbf{0} & \textbf{1} & \textbf{2} & \textbf{3} & \textbf{4} \\
\midrule

% ---------------- Llama ----------------------
\multirow{3}{*}{Llama-3.1-70B}
& Multi-turn & 231.0 & 715.1 & 1373.4 & 2070.4 & 2902.1 \\
& Memory  & 231.0 & 740.4 & 1059.7 & 1301.0 & 1340.6 \\
& In-place   & 231.0 & 378.0 & 472.7  & 534.1  & 610.2  \\
\midrule

% ---------------- Gemma ----------------------
\multirow{3}{*}{Gemma-3-27B}
& Multi-turn & 177.9 & 1833.3 & 3527.8 & 4934.1 & 6250.7 \\
& Memory  & 177.9 & 1897.3 & 1769.4 & 1958.0 & 2020.5 \\
& In-place   & 177.9 & 1008.8 & 1425.0 & 1540.9 & 1731.2 \\
\bottomrule

\end{tabular}
}
\caption{Number of input tokens on Math-hard.}
\label{tab:app_input_token_length_math_hard_large}
\end{table}

\begin{table}[H]
\centering
\resizebox{\linewidth}{!}{
\begin{tabular}{llrrrrr}
\toprule
\multirow{2}{*}{\textbf{Model}}
& \multirow{2}{*}{\textbf{Method}}
& \multicolumn{5}{c}{\textbf{Number of turns}} \\
\cmidrule{3-7}
& & \textbf{0} & \textbf{1} & \textbf{2} & \textbf{3} & \textbf{4} \\
\midrule

% ---------------- Llama ----------------------
\multirow{3}{*}{Llama-3.1-70B}
& Multi-turn & 165.1 & 747.6 & 1342.3 & 1987.7 & 2608.3 \\
& Memory  & 165.1 & 776.8 & 954.3  & 1187.8 & 1210.4 \\
& In-place   & 165.1 & 451.9 & 515.8  & 560.3  & 595.4  \\
\midrule

% ---------------- Gemma ----------------------
\multirow{3}{*}{Gemma-3-27B}
& Multi-turn & 149.0 & 894.6 & 1672.0 & 2454.1 & 3209.8 \\
& Memory  & 149.0 & 924.6 & 1117.7 & 1315.8 & 1314.0 \\
& In-place   & 149.0 & 473.6 & 553.6  & 642.9  & 649.5  \\
\bottomrule

\end{tabular}
}
\caption{Number of input tokens on MMLU-pro.}
\label{tab:app_input_token_length_mmlu_pro_large}
\end{table}

\begin{table}[H]
\centering
\resizebox{\linewidth}{!}{
\begin{tabular}{llrrrrr}
\toprule
\multirow{2}{*}{\textbf{Model}}
& \multirow{2}{*}{\textbf{Method}}
& \multicolumn{5}{c}{\textbf{Number of turns}} \\
\cmidrule{3-7}
& & \textbf{0} & \textbf{1} & \textbf{2} & \textbf{3} & \textbf{4} \\
\midrule

% ---------------- Llama ----------------------
\multirow{3}{*}{Llama-3.1-70B}
& Multi-turn & 224.1 & 998.2  & 1720.3 & 2405.4 & 3099.9 \\
& Memory  & 224.1 & 1027.5 & 1138.1 & 1309.9 & 1307.2 \\
& In-place   & 224.1 & 585.9  & 636.0  & 694.4  & 752.2  \\
\midrule

% ---------------- Gemma ----------------------
\multirow{3}{*}{Gemma-3-27B}
& Multi-turn & 205.0 & 1085.9 & 1903.6 & 2754.2 & 3606.0 \\
& Memory  & 205.0 & 1115.9 & 1230.8 & 1380.2 & 1414.2 \\
& In-place   & 205.0 & 555.0  & 692.9  & 756.2  & 783.1  \\
\bottomrule

\end{tabular}
}
\caption{Number of input tokens on GPQA.}
\label{tab:app_input_token_length_gpqa_large}
\end{table}

\begin{table}[H]
\centering
\resizebox{\linewidth}{!}{
\begin{tabular}{llrrrrr}
\toprule
\multirow{2}{*}{\textbf{Model}}
& \multirow{2}{*}{\textbf{Method}}
& \multicolumn{5}{c}{\textbf{Number of turns}} \\
\cmidrule{3-7}
& & \textbf{0} & \textbf{1} & \textbf{2} & \textbf{3} & \textbf{4} \\
\midrule

% ---------------- Llama ----------------------
\multirow{3}{*}{Llama-3.1-70B}
& Multi-turn & 584.3 & 1317.0 & 2055.4 & 2835.1 & 3598.0 \\
& Memory     & 584.3 & 1342.2 & 1439.8 & 1648.5 & 1662.7 \\
& In-place   & 584.3 & 931.8  & 1007.7 & 1080.3 & 1186.6 \\
\midrule

% ---------------- Gemma ----------------------
\multirow{3}{*}{Gemma-3-27B}
& Multi-turn & 594.3 & 1251.5 & 1922.7 & 2529.5 & 3266.9 \\
& Memory     & 594.3 & 1571.1 & 1914.6 & 2182.0 & 2254.8 \\
& In-place   & 594.3 & 1005.6 & 1139.5 & 1180.6 & 1263.5 \\
\bottomrule

\end{tabular}
}
\caption{Number of input tokens on LiveCodeBench.}
\label{tab:app_input_token_length_code_large}
\end{table}

\begin{table}[H]
\centering
\resizebox{\linewidth}{!}{
\begin{tabular}{llrrrrr}
\toprule
\multirow{2}{*}{\textbf{Model}}
& \multirow{2}{*}{\textbf{Method}}
& \multicolumn{5}{c}{\textbf{Number of turns}} \\
\cmidrule{3-7}
& & \textbf{0} & \textbf{1} & \textbf{2} & \textbf{3} & \textbf{4} \\
\midrule

% ---------------- Llama ----------------------
\multirow{3}{*}{Llama-3.1-70B}
& Multi-turn & 915.3 & 2488.4 & 4109.8 & 5844.1 & 7626.1 \\
& Memory     & 915.3 & 2454.8 & 2250.5 & 2460.7 & 2558.0 \\
& In-place   & 915.3 & 2147.0 & 3037.9 & 3612.3 & 4322.2 \\
\midrule
% ---------------- Gemma ----------------------
\multirow{3}{*}{Gemma-3-27B}
& Multi-turn & 961.4 & 2269.8 & 2971.6 & 3663.9 & 4401.4 \\
& Memory     & 961.4 & 2245.8 & 2133.8 & 2421.5 & 2599.9 \\
& In-place   & 961.4 & 2067.0 & 2411.5 & 2638.4 & 2848.3 \\
\bottomrule

\end{tabular}
}
\caption{Number of input tokens on ZebraLogic.}
\label{tab:app_input_token_length_zebralogic_large}
\end{table}

\subsection{Generated Token Length}
\label{app.gen_token_len_large_scale}

\begin{table}[H]
\centering
\resizebox{\linewidth}{!}{
\begin{tabular}{llrrrrr}
\toprule
\multirow{2}{*}{\textbf{Model}}
& \multirow{2}{*}{\textbf{Method}}
& \multicolumn{5}{c}{\textbf{Number of turns}} \\
\cmidrule{3-7}
& & \textbf{0} & \textbf{1} & \textbf{2} & \textbf{3} & \textbf{4} \\
\midrule

% ---------------- Llama-70B ----------------------
\multirow{3}{*}{Llama-3.1-70B}
& Multi-turn & 279.8 & 380.0 & 482.5 & 522.4 & 593.4 \\
& Memory  & 279.8 & 434.3 & 555.5 & 615.1 & 592.4 \\
& In-place   & 279.8 & 249.2 & 239.0 & 315.2 & 300.5 \\
\midrule

% ---------------- Gemma-27B ----------------------
\multirow{3}{*}{Gemma-3-27B}
& Multi-turn & 927.4 & 1134.5 & 1238.2 & 1154.2 & 1098.9 \\
& Memory  & 927.4 & 969.4 & 1134.5 & 1222.9 & 1204.4 \\
& In-place   & 927.4 & 807.4 & 731.0 & 879.4 & 756.5 \\
\bottomrule

\end{tabular}
}
\caption{Number of generated tokens on Math-hard.}
\label{tab:app_output_token_length_large_scale_math_hard}
\end{table}

\begin{table}[H]
\centering
\resizebox{\linewidth}{!}{
\begin{tabular}{llrrrrr}
\toprule
\multirow{2}{*}{\textbf{Model}}
& \multirow{2}{*}{\textbf{Method}}
& \multicolumn{5}{c}{\textbf{Number of turns}} \\
\cmidrule{3-7}
& & \textbf{0} & \textbf{1} & \textbf{2} & \textbf{3} & \textbf{4} \\
\midrule

% ---------------- Llama-70B ----------------------
\multirow{3}{*}{Llama-3.1-70B}
& Multi-turn & 310.1 & 385.4 & 423.6 & 467.4 & 477.6 \\
& Memory  & 310.1 & 432.1 & 515.1 & 575.7 & 623.8 \\
& In-place   & 310.1 & 153.9 & 160.8 & 166.7 & 147.7 \\
\midrule

% ---------------- Gemma-27B ----------------------
\multirow{3}{*}{Gemma-3-27B}
& Multi-turn & 420.3 & 511.5 & 561.6 & 561.3 & 605.3 \\
& Memory  & 420.3 & 586.8 & 680.9 & 712.1 & 723.7 \\
& In-place   & 420.3 & 291.5 & 297.5 & 327.8 & 324.0 \\
\bottomrule

\end{tabular}
}
\caption{Number of generated tokens on MMLU-pro.}
\label{tab:app_output_token_length_large_scale_MMLU-Pro}
\end{table}

\begin{table}[H]
\centering
\resizebox{\linewidth}{!}{
\begin{tabular}{llrrrrr}
\toprule
\multirow{2}{*}{\textbf{Model}}
& \multirow{2}{*}{\textbf{Method}}
& \multicolumn{5}{c}{\textbf{Number of turns}} \\
\cmidrule{3-7}
& & \textbf{0} & \textbf{1} & \textbf{2} & \textbf{3} & \textbf{4} \\
\midrule

% ---------------- Llama-70B ----------------------
\multirow{3}{*}{Llama-3.1-70B}
& Multi-turn & 597.8 & 549.2 & 508.2 & 515.6 & 571.9 \\
& Memory  & 597.8 & 574.9 & 607.8 & 600.5 & 620.3 \\
& In-place   & 597.8 & 280.1 & 229.7 & 232.3 & 259.7 \\
\midrule

% ---------------- Gemma-27B ----------------------
\multirow{3}{*}{Gemma-3-27B}
& Multi-turn & 738.9 & 641.1 & 643.1 & 666.4 & 679.0 \\
& Memory  & 738.9 & 683.3 & 728.2 & 751.3 & 749.8 \\
& In-place   & 738.9 & 417.1 & 330.4 & 292.9 & 315.6 \\
\bottomrule

\end{tabular}
}
\caption{Number of generated tokens on GPQA.}
\label{tab:app_output_token_length_large_scale_GPQA}
\end{table}

\begin{table}[H]
\centering
\resizebox{\linewidth}{!}{
\begin{tabular}{llrrrrr}
\toprule
\multirow{2}{*}{\textbf{Model}}
& \multirow{2}{*}{\textbf{Method}}
& \multicolumn{5}{c}{\textbf{Number of turns}} \\
\cmidrule{3-7}
& & \textbf{0} & \textbf{1} & \textbf{2} & \textbf{3} & \textbf{4} \\
\midrule

% ---------------- Llama-70B ----------------------
\multirow{3}{*}{Llama-3.1-70B}
& Multi-turn & 467.9 & 488.3 & 513.9 & 553.0 & 517.2 \\
& Memory     & 467.9 & 369.2 & 392.5 & 420.2 & 429.3 \\
& In-place   & 467.9 & 233.2 & 217.7 & 238.8 & 212.5 \\
\midrule

% ---------------- Gemma-27B ----------------------
\multirow{3}{*}{Gemma-3-27B}
& Multi-turn & 449.1 & 573.4 & 499.5 & 472.8 & 522.9 \\
& Memory     & 449.1 & 552.9 & 704.2 & 835.5 & 895.1 \\
& In-place   & 449.1 & 285.8 & 438.3 & 482.7 & 313.8 \\
\bottomrule

\end{tabular}
}
\caption{Number of generated tokens on LiveCodeBench.}
\label{tab:app_output_token_length_large_scale_code}
\end{table}

\begin{table}[H]
\centering
\resizebox{\linewidth}{!}{
\begin{tabular}{llrrrrr}
\toprule
\multirow{2}{*}{\textbf{Model}}
& \multirow{2}{*}{\textbf{Method}}
& \multicolumn{5}{c}{\textbf{Number of turns}} \\
\cmidrule{3-7}
& & \textbf{0} & \textbf{1} & \textbf{2} & \textbf{3} & \textbf{4} \\
\midrule

% ---------------- Llama ----------------------
\multirow{3}{*}{Llama-3.1-70B}
& Multi-turn & 1429.6 & 1444.6 & 1439.1 & 1431.2 & 1445.6 \\
& Memory     & 1429.6 & 1126.5 & 1200.4 & 1178.2 & 1216.5 \\
& In-place   & 1429.6 & 1069.6 & 791.1 & 762.5 & 666.8 \\
\midrule
% ---------------- Gemma ----------------------
\multirow{3}{*}{Gemma-3-27B}
& Multi-turn & 1191.3 & 575.1 & 590.1 & 585.2 & 605.5 \\
& Memory     & 1191.3 & 985.0 & 1153.2 & 1250.3 & 1313.4 \\
& In-place   & 1191.3 & 414.2 & 387.7 & 402.7 & 423.0 \\
\bottomrule

\end{tabular}
}
\caption{Number of generated tokens on ZebraLogic.}
\label{tab:app_output_token_length_zebralogic_large}
\end{table}

\subsection{Results on ZebraLogic}
\label{app:large_zebra}

We conduct experiments on ZebraLogic with Large-scale LLMs, such as \texttt{Gemma-3-27b-it} and \texttt{Llama-3.1-70B-Instruct}. Due to cost constraints, we run five turns and one seed.

\paragraph{Task performance.}

\begin{figure}[H]
    \centering
    \includegraphics[width=1.0\linewidth]{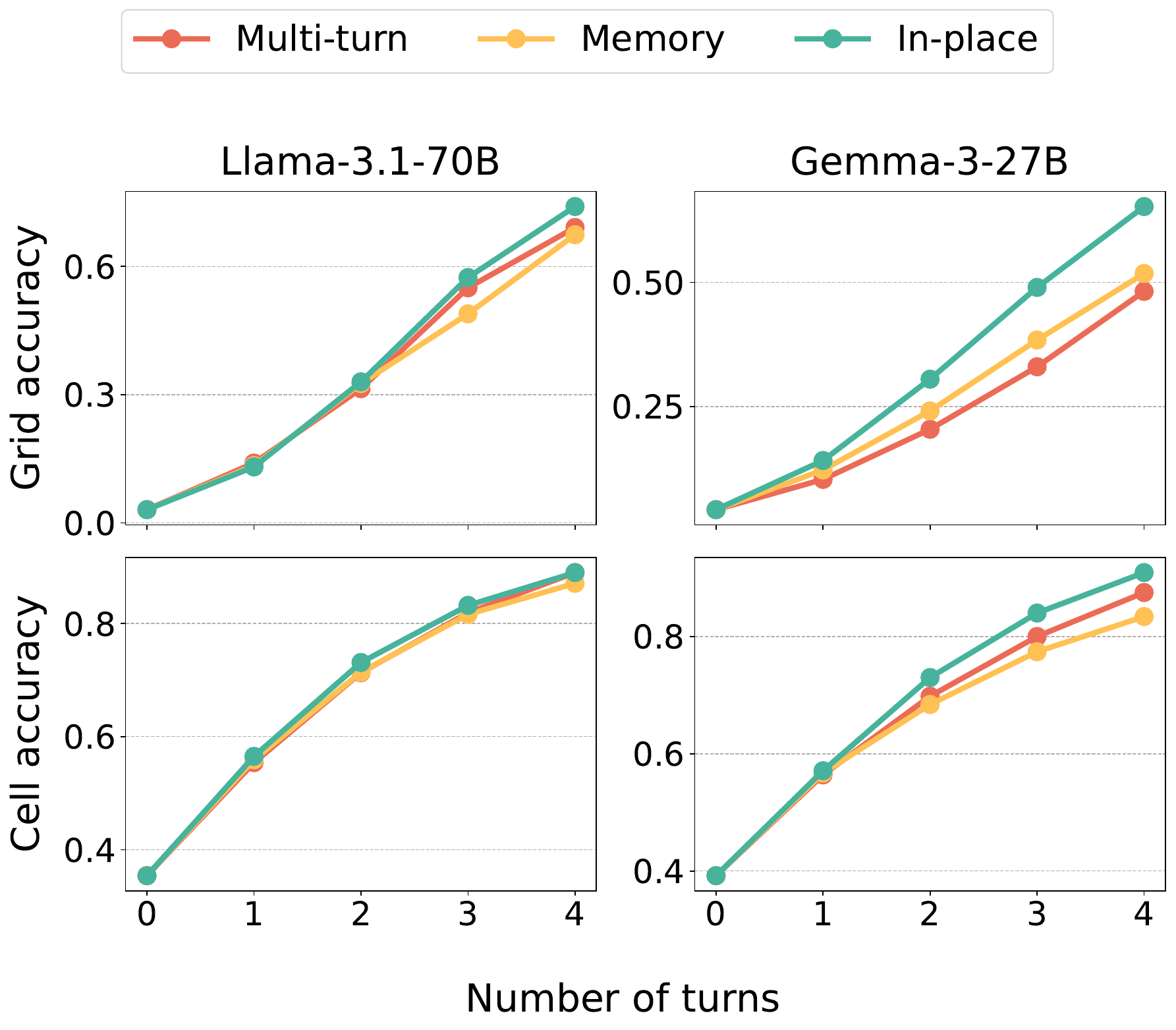}
    \caption{Grid and cell accuracy on the ZebraLogic dataset with large-scale LLMs. Tabulated results are provided in \cref{tab:app_large_scale_game_acc_top4,tab:app_large_scale_cell_acc_top4,tab:app_large_scale_game_acc_top2,tab:app_large_scale_cell_acc_top2}.}
    \label{fig:zebra_acc_combined_large}
\end{figure}

As shown in \cref{fig:zebra_acc_combined_large}, in-place feedback also outperforms other baselines, except for \texttt{Llama} with top-2 feedback. This gap arises from two factors: the in-place agent is less reliable than human experts at applying corrections, and due to the parallel reasoning structure of ZebraLogic, the edited span may inherit incorrect reasoning from earlier turns. This limitation could be exacerbated under top-2 feedback, in which fewer corrections lead to more errors that propagate through subsequent reasoning.

\paragraph{Turn-level dynamics.} 

\begin{figure}[H]
    \centering
    \includegraphics[width=0.8\linewidth]{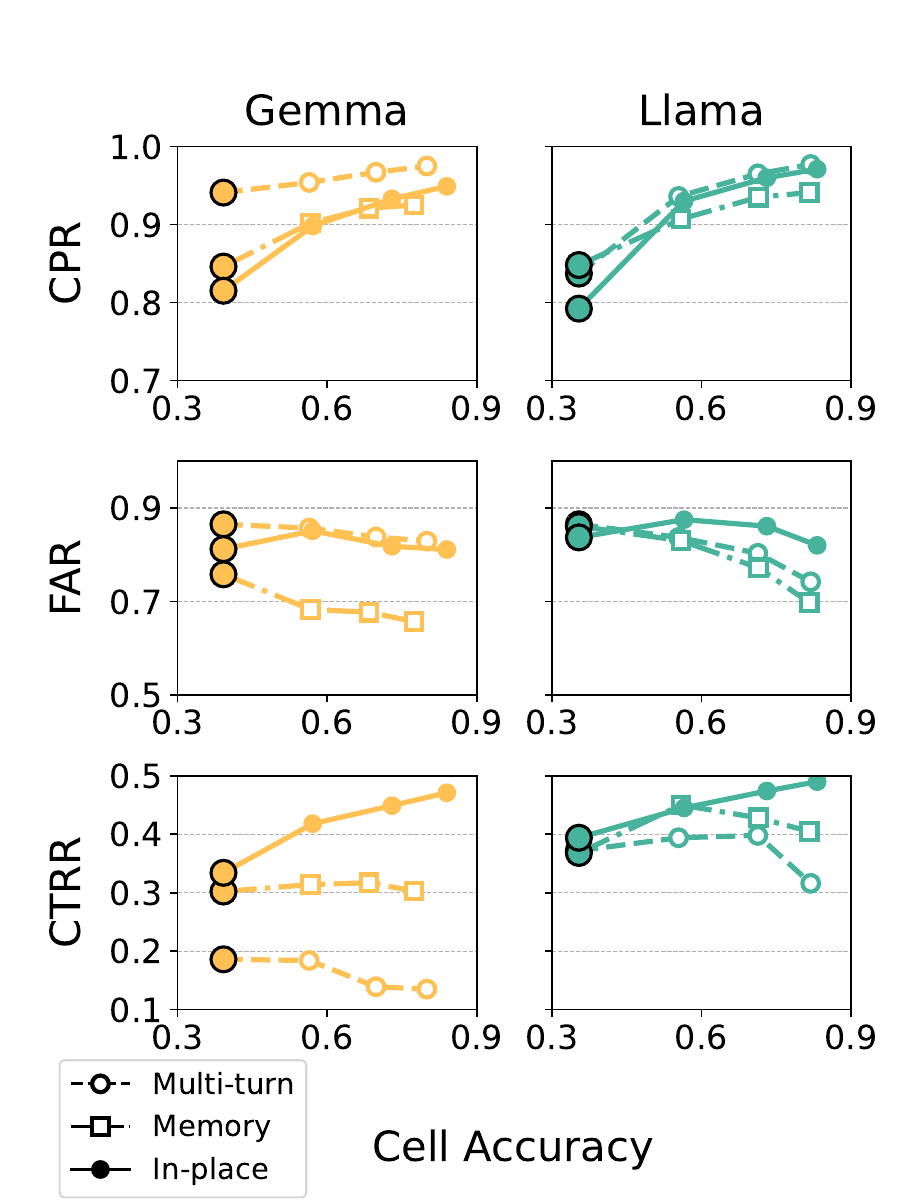}
    \caption{CPR, FAR, and CTRR for five-turn conversations of large-scale LLMs on the ZebraLogic dataset with top-$4$ feedback. The points with a black border represent the second response of the LLMs (i.e., $y_1$), and the subsequent responses across turns are connected by lines.}
    \label{fig:app_zebra_top4_large}
\end{figure}

\begin{figure}[H]
    \centering
    \includegraphics[width=0.8\linewidth]{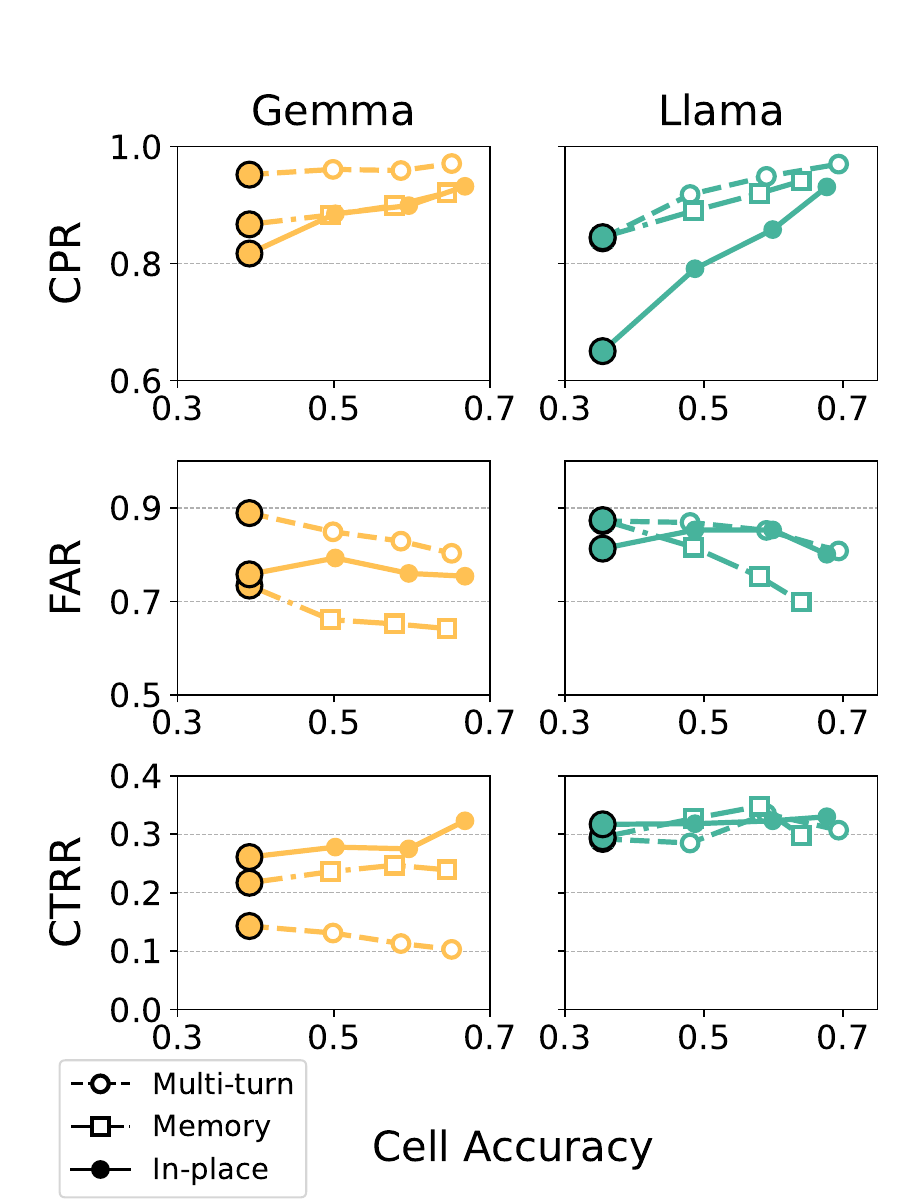}
    \caption{CPR, FAR, and CTRR for five-turn conversations of large-scale LLMs on the ZebraLogic dataset with top-$2$ feedback. The points with a black border represent the second response of the LLMs (i.e., $y_1$), and the subsequent responses across turns are connected by lines.}
    \label{fig:app_zebra_top2_large}
\end{figure}

\cref{fig:app_zebra_top4_large,fig:app_zebra_top2_large} illustrate turn-level dynamics for large-scale LLMs. The overall patterns align with \cref{fig:zebra_reasoning}. In-place feedback maintains the highest CTRR across all turns. In contrast, multi-turn feedback shows a rising CPR but a declining CTRR over turns, indicating a trade-off between preserving correctness and supporting further reasoning. Memory-based and in-place feedback keep CTRR relatively stable across turns, although their CPR is generally lower than that of multi-turn feedback. Unlike the smaller models, the decline in FAR across turns is less pronounced in these large-scale LLMs, likely due to improved instruction-following capability. However, FAR decline is still observable in \texttt{Llama} with top-4 feedback, suggesting that this pattern may emerge with more turns.

\begin{table}[H]
\centering
\resizebox{\linewidth}{!}{
\begin{tabular}{llrrrrr}
\toprule
\multirow{2}{*}{\textbf{Model}}
& \multirow{2}{*}{\textbf{Method}}
& \multicolumn{5}{c}{\textbf{Number of turns}} \\
\cmidrule{3-7}
& & \textbf{0} & \textbf{1} & \textbf{2} & \textbf{3} & \textbf{4} \\
\midrule

% ---------------- Llama ----------------------
\multirow{3}{*}{Llama-3.1-70B}
& Multi-turn & 0.031 & 0.140  & 0.314 & 0.550 & 0.691 \\
& Memory  & 0.031 & 0.134 & 0.326 & 0.489 & 0.674 \\
& In-place   & 0.031 & 0.131  & 0.330  & 0.574  & 0.740  \\
\midrule

% ---------------- Gemma ----------------------
\multirow{3}{*}{Gemma-3-27B}
& Multi-turn & 0.042 & 0.103 & 0.204 & 0.330 & 0.482 \\
& Memory  & 0.042 & 0.122 & 0.241 & 0.384 & 0.518 \\
& In-place   & 0.042 & 0.141  & 0.305  & 0.490  & 0.653  \\
\bottomrule

\end{tabular}
}
\caption{Grid accuracy of large-scale LLMs on ZebraLogic with top-$4$ feedback.}
\label{tab:app_large_scale_game_acc_top4}
\end{table}

\begin{table}[H]
\centering
\resizebox{\linewidth}{!}{
\begin{tabular}{llrrrrr}
\toprule
\multirow{2}{*}{\textbf{Model}}
& \multirow{2}{*}{\textbf{Method}}
& \multicolumn{5}{c}{\textbf{Number of turns}} \\
\cmidrule{3-7}
& & \textbf{0} & \textbf{1} & \textbf{2} & \textbf{3} & \textbf{4} \\
\midrule

% ---------------- Llama ----------------------
\multirow{3}{*}{Llama-3.1-70B}
& Multi-turn & 0.031 & 0.096  & 0.195 & 0.337 & 0.492 \\
& Memory  & 0.031 & 0.108 & 0.234 & 0.360 & 0.455 \\
& In-place   & 0.031 & 0.096  & 0.201  & 0.335  & 0.475  \\
\midrule

% ---------------- Gemma ----------------------
\multirow{3}{*}{Gemma-3-27B}
& Multi-turn & 0.042 & 0.094 & 0.166 & 0.234 & 0.314 \\
& Memory  & 0.042 & 0.099 & 0.180 & 0.262 & 0.344 \\
& In-place   & 0.042 & 0.115  & 0.209  & 0.297  & 0.405  \\
\bottomrule

\end{tabular}
}
\caption{Grid accuracy of large-scale LLMs on ZebraLogic with top-$2$ feedback.}
\label{tab:app_large_scale_game_acc_top2}
\end{table}

\begin{table}[H]
\centering
\resizebox{\linewidth}{!}{
\begin{tabular}{llrrrrr}
\toprule
\multirow{2}{*}{\textbf{Model}}
& \multirow{2}{*}{\textbf{Method}}
& \multicolumn{5}{c}{\textbf{Number of turns}} \\
\cmidrule{3-7}
& & \textbf{0} & \textbf{1} & \textbf{2} & \textbf{3} & \textbf{4} \\
\midrule

% ---------------- Llama ----------------------
\multirow{3}{*}{Llama-3.1-70B}
& Multi-turn & 0.354 & 0.554  & 0.713 & 0.819 & 0.890 \\
& Memory  & 0.354 & 0.559 & 0.714 & 0.816 & 0.871 \\
& In-place   & 0.354 & 0.565  & 0.731 & 0.832 & 0.890  \\
\midrule

% ---------------- Gemma ----------------------
\multirow{3}{*}{Gemma-3-27B}
& Multi-turn & 0.392  & 0.564 & 0.698 & 0.800 & 0.875 \\
& Memory  & 0.392 & 0.567 & 0.684 & 0.774 & 0.834 \\
& In-place   & 0.392 & 0.571  & 0.730  & 0.840 & 0.909  \\
\bottomrule

\end{tabular}
}
\caption{Cell accuracy of large-scale LLMs on ZebraLogic. Top-$4$ feedback is used.}
\label{tab:app_large_scale_cell_acc_top4}
\end{table}

\begin{table}[H]
\centering
\resizebox{\linewidth}{!}{
\begin{tabular}{llrrrrr}
\toprule
\multirow{2}{*}{\textbf{Model}}
& \multirow{2}{*}{\textbf{Method}}
& \multicolumn{5}{c}{\textbf{Number of turns}} \\
\cmidrule{3-7}
& & \textbf{0} & \textbf{1} & \textbf{2} & \textbf{3} & \textbf{4} \\
\midrule

% ---------------- Llama ----------------------
\multirow{3}{*}{Llama-3.1-70B}
& Multi-turn & 0.354 & 0.480  & 0.590 & 0.694 & 0.756 \\
& Memory  & 0.354 & 0.485 & 0.580 & 0.640 & 0.696 \\
& In-place   & 0.354 & 0.487  & 0.599  & 0.677  & 0.730  \\
\midrule

% ---------------- Gemma ----------------------
\multirow{3}{*}{Gemma-3-27B}
& Multi-turn & 0.392 & 0.499 & 0.586 & 0.651 & 0.711 \\
& Memory   & 0.392 & 0.496 & 0.578 & 0.645 & 0.699 \\
& In-place   & 0.392 & 0.502  & 0.596  & 0.668  & 0.747  \\
\bottomrule

\end{tabular}
}
\caption{Cell accuracy of large-scale LLMs on ZebraLogic with top-$2$ feedback.}
\label{tab:app_large_scale_cell_acc_top2}
\end{table}

\section{Tabulated Results}
\label{app.tabulated_results}

This section provides the numerical values corresponding to the main figures presented in the paper.

\subsection{Task Performance}

\begin{table}[H]
\centering
\resizebox{\linewidth}{!}{
\begin{tabular}{llcccccccccc}
\toprule

& \multirow{2}{*}{\textbf{Method}} 
& \multicolumn{10}{c}{\textbf{Number of turns}} \\
\cmidrule{3-12}
& & \textbf{0} & \textbf{1} & \textbf{2} & \textbf{3} 
& \textbf{4} & \textbf{5} & \textbf{6} & \textbf{7} & \textbf{8} & \textbf{9} \\
\midrule

% Qwen
\multirow{3}{*}{\rotatebox{90}{Qwen}}
& Multi-turn & 55.0 & 68.4 & 73.8 & 76.6 & 79.2 & 81.6 & 83.8 & 85.0 & 85.8 & 86.2 \\
& Memory  & 55.0 & 62.6 & 66.2 & 69.2 & 71.2 & 72.8 & 74.2 & 75.0 & 76.4 & 77.4 \\
& In-place   & 55.0 & 71.4 & 76.6 & 81.2 & 85.2 & 88.8 & 90.4 & 91.8 & 92.4 & 92.8 \\
\midrule

% Gemma
\multirow{3}{*}{\rotatebox{90}{Gemma}}
& Multi-turn & 49.2 & 60.6 & 68.4 & 73.0 & 76.0 & 77.6 & 79.2 & 80.4 & 82.0 & 82.8 \\
& Memory  & 49.2 & 60.4 & 65.0 & 68.0 & 70.6 & 72.6 & 74.6 & 76.4 & 77.2 & 78.0 \\
& In-place   & 49.2 & 62.6 & 70.8 & 74.8 & 77.8 & 81.4 & 82.8 & 85.0 & 86.6 & 87.6 \\
\midrule

% Llama
\multirow{3}{*}{\rotatebox{90}{Llama}}
& Multi-turn & 14.8 & 34.0 & 45.8 & 56.4 & 59.8 & 64.4 & 67.4 & 71.2 & 73.4 & 75.6 \\
& Memory  & 14.8 & 29.4 & 40.0 & 47.0 & 52.8 & 57.2 & 62.2 & 65.6 & 67.6 & 69.8 \\
& In-place   & 14.8 & 36.6 & 48.0 & 59.8 & 66.8 & 74.4 & 76.8 & 80.6 & 82.8 & 84.0 \\
\bottomrule
\end{tabular}
}
\caption{Performance comparison on Math-hard.}
\label{tab:app_performance_math_hard}
\end{table}

\begin{table}[H]
\centering
\resizebox{\linewidth}{!}{
\begin{tabular}{llcccccccccc}
\toprule
& \multirow{2}{*}{\textbf{Method}} 
& \multicolumn{10}{c}{\textbf{Number of turns}} \\
\cmidrule{3-12}
& & \textbf{0} & \textbf{1} & \textbf{2} & \textbf{3} 
& \textbf{4} & \textbf{5} & \textbf{6} & \textbf{7} & \textbf{8} & \textbf{9} \\
\midrule

% Qwen
\multirow{3}{*}{\rotatebox{90}{Qwen}}
& Multi-turn & 56.2 & 68.0 & 72.6 & 75.1 & 77.3 & 78.1 & 79.1 & 79.3 & 80.1 & 80.5 \\
& Memory  & 56.2 & 67.1 & 70.8 & 73.4 & 75.7 & 77.7 & 78.5 & 79.7 & 80.7 & 80.9 \\
& In-place   & 56.2 & 69.8 & 78.5 & 80.1 & 81.9 & 83.0 & 83.8 & 84.6 & 85.0 & 85.4 \\
\midrule

% Gemma
\multirow{3}{*}{\rotatebox{90}{Gemma}}
& Multi-turn & 41.4 & 54.0 & 60.0 & 65.9 & 68.4 & 69.8 & 71.2 & 72.0 & 72.8 & 73.4 \\
& Memory  & 41.4 & 52.9 & 57.2 & 63.5 & 66.3 & 68.0 & 69.6 & 70.4 & 71.6 & 72.2 \\
& In-place   & 41.4 & 59.4 & 65.5 & 69.6 & 73.6 & 76.1 & 77.7 & 79.5 & 80.7 & 81.3 \\
\midrule

% Llama
\multirow{3}{*}{\rotatebox{90}{Llama}}
& Multi-turn & 42.2 & 55.8 & 61.5 & 65.7 & 69.0 & 71.2 & 72.0 & 73.4 & 74.8 & 75.1 \\
& Memory  & 42.2 & 57.0 & 63.9 & 68.6 & 71.2 & 74.0 & 75.9 & 77.3 & 78.3 & 78.7 \\
& In-place   & 42.2 & 59.4 & 68.2 & 74.2 & 76.5 & 79.7 & 81.3 & 82.8 & 84.2 & 84.8 \\
\bottomrule

\end{tabular}
}
\caption{Performance comparison on MMLU-pro.}
\label{tab:app_performance_MMLU_Pro}
\end{table}

\begin{table}[H]
\centering
\resizebox{\linewidth}{!}{
\begin{tabular}{llcccccccccc}
\toprule
& \multirow{2}{*}{\textbf{Method}} 
& \multicolumn{10}{c}{\textbf{Number of turns}} \\
\cmidrule{3-12}
& & \textbf{0} & \textbf{1} & \textbf{2} & \textbf{3} 
& \textbf{4} & \textbf{5} & \textbf{6} & \textbf{7} & \textbf{8} & \textbf{9} \\
\midrule

% Qwen
\multirow{3}{*}{\rotatebox{90}{Qwen}}
& Multi-turn & 10.3 & 17.5 & 27.8 & 38.1 & 40.5 & 42.9 & 43.7 & 43.7 & 43.7 & 46.0 \\
& Memory  & 10.3 & 19.0 & 27.0 & 29.4 & 34.9 & 36.5 & 37.3 & 38.1 & 38.1 & 38.9 \\
& In-place   & 10.3 & 25.4 & 36.5 & 48.4 & 51.6 & 57.9 & 63.5 & 64.3 & 66.7 & 69.0 \\
\midrule

% Gemma
\multirow{3}{*}{\rotatebox{90}{Gemma}}
& Multi-turn & 5.6 & 9.5 & 11.9 & 14.3 & 15.1 & 19.0 & 22.2 & 23.0 & 24.6 & 27.0 \\
& Memory  & 5.6 & 8.7 & 10.3 & 13.5 & 15.9 & 17.5 & 18.3 & 22.2 & 22.2 & 23.8 \\
& In-place   & 5.6 & 11.9 & 27.8 & 33.3 & 38.9 & 42.9 & 46.0 & 46.0 & 47.6 & 53.2 \\
\midrule

% Llama
\multirow{3}{*}{\rotatebox{90}{Llama}}
& Multi-turn & 4.0 & 8.7 & 19.0 & 24.6 & 27.0 & 31.0 & 34.1 & 38.1 & 42.1 & 43.7 \\
& Memory  & 4.0 & 8.7 & 15.1 & 20.6 & 23.0 & 27.8 & 31.7 & 34.1 & 37.3 & 40.5 \\
& In-place   & 4.0 & 15.1 & 26.2 & 35.7 & 43.7 & 47.6 & 53.2 & 55.6 & 61.1 & 61.9 \\
\bottomrule

\end{tabular}
}
\caption{Performance comparison on GPQA.}
\label{tab:app_performance_GPQA}
\end{table}

\begin{table}[H]
    \centering
    \resizebox{\columnwidth}{!}{%
    \begin{tabular}{ll cccccccccc}
        \toprule
        \multirow{2}{*}{Model} & \multirow{2}{*}{Method} & \multicolumn{10}{c}{Number of turns} \\
        \cmidrule(lr){3-12}
        & & 0 & 1 & 2 & 3 & 4 & 5 & 6 & 7 & 8 & 9 \\
        \midrule
        \multirow{3}{*}{Qwen}
            & Multi-turn & 0.033 & 0.033 & 0.067 & 0.067 & 0.067 & 0.067 & 0.100 & 0.100 & 0.100 & 0.100 \\
            & Memory     & 0.033 & 0.033 & 0.067 & 0.100 & 0.100 & 0.100 & 0.100 & 0.100 & 0.100 & 0.100 \\
            & In-place   & 0.033 & 0.200 & 0.233 & 0.300 & 0.333 & 0.433 & 0.433 & 0.467 & 0.467 & 0.467 \\
        \midrule
        \multirow{3}{*}{Gemma}
            & Multi-turn & 0.100 & 0.133 & 0.133 & 0.167 & 0.167 & 0.167 & 0.167 & 0.167 & 0.167 & 0.167 \\
            & Memory     & 0.100 & 0.100 & 0.200 & 0.233 & 0.267 & 0.300 & 0.300 & 0.333 & 0.333 & 0.333 \\
            & In-place   & 0.100 & 0.133 & 0.200 & 0.300 & 0.300 & 0.367 & 0.367 & 0.367 & 0.367 & 0.367 \\
        \bottomrule
    \end{tabular}%
    }
    \caption{Cumulative accuracy on AIME~2025.}
    \label{tab:rebuttal-aime-2025}
\end{table}

\begin{table}[H]
    \centering
    \resizebox{\columnwidth}{!}{%
    \begin{tabular}{ll cccccccccc}
        \toprule
        \multirow{2}{*}{Model} & \multirow{2}{*}{Method} & \multicolumn{10}{c}{Number of turns} \\
        \cmidrule(lr){3-12}
        & & 0 & 1 & 2 & 3 & 4 & 5 & 6 & 7 & 8 & 9 \\
        \midrule
        \multirow{3}{*}{Qwen}
            & Multi-turn & 0.067 & 0.100 & 0.100 & 0.100 & 0.100 & 0.133 & 0.133 & 0.133 & 0.133 & 0.133 \\
            & Memory     & 0.067 & 0.067 & 0.067 & 0.067 & 0.067 & 0.067 & 0.067 & 0.067 & 0.067 & 0.067 \\
            & In-place   & 0.067 & 0.133 & 0.233 & 0.267 & 0.333 & 0.467 & 0.533 & 0.533 & 0.533 & 0.533 \\
        \midrule
        \multirow{3}{*}{Gemma}
            & Multi-turn & 0.067 & 0.133 & 0.167 & 0.167 & 0.167 & 0.167 & 0.167 & 0.200 & 0.200 & 0.200 \\
            & Memory     & 0.067 & 0.067 & 0.133 & 0.133 & 0.200 & 0.200 & 0.200 & 0.200 & 0.200 & 0.200 \\
            & In-place   & 0.067 & 0.167 & 0.267 & 0.333 & 0.333 & 0.333 & 0.333 & 0.367 & 0.367 & 0.367 \\
        \bottomrule
    \end{tabular}%
    }
    \caption{Cumulative accuracy on AIME~2026.}
    \label{tab:rebuttal-aime-2026}
\end{table}

\subsection{Number of Input Tokens}

\begin{table}[H]
\centering
\resizebox{\linewidth}{!}{
\begin{tabular}{llrrrrrrrrrr}
\toprule
& \multirow{2}{*}{\textbf{Method}}
& \multicolumn{10}{c}{\textbf{Number of turns}} \\
\cmidrule{3-12}
& & \textbf{0} & \textbf{1} & \textbf{2} & \textbf{3} & \textbf{4} & \textbf{5} & \textbf{6} & \textbf{7} & \textbf{8} & \textbf{9} \\
\midrule

% ---------------- Llama ----------------------
\multirow{3}{*}{\rotatebox{90}{Llama}}
& Multi-turn & 231.0 & 800.9 & 1572.4 & 2453.3 & 3513.4 & 4433.9 & 5513.2 & 6621.6 & 7854.8 & 8949.1 \\
& Memory     & 231.0 & 826.2 & 1073.4 & 1319.3 & 1361.2 & 1414.1 & 1413.2 & 1468.8 & 1515.9 & 1566.9 \\
& In-place   & 231.0 & 373.3 & 456.5 & 533.1 & 608.1 & 666.8 & 733.2 & 790.0 & 825.0 & 872.1 \\
\midrule

% ---------------- Gemma ----------------------
\multirow{3}{*}{\rotatebox{90}{Gemma}}
& Multi-turn & 165.9 & 1822.6 & 3247.8 & 4640.9 & 5850.4 & 7188.1 & 8483.4 & 9833.5 & 11076.5 & 12444.9 \\
& Memory     & 165.9 & 1898.6 & 1537.5 & 1712.3 & 1717.8 & 1737.9 & 1712.2 & 1744.5 & 1739.2 & 1720.0 \\
& In-place   & 165.9 & 770.8 & 1042.0 & 1316.6 & 1334.0 & 1355.4 & 1439.0 & 1635.8 & 1683.7 & 1727.9 \\
\midrule

% ---------------- Qwen ----------------------
\multirow{3}{*}{\rotatebox{90}{Qwen}}
& Multi-turn & 159.6 & 1288.6 & 2538.6 & 3838.7 & 5235.0 & 6806.2 & 8368.0 & 9977.5 & 11671.0 & 13434.7 \\
& Memory     & 159.6 & 1364.8 & 858.4 & 1134.3 & 1100.5 & 1144.9 & 1153.2 & 1160.6 & 1095.7 & 1125.5 \\
& In-place   & 159.6 & 734.7 & 950.6 & 1088.9 & 1135.6 & 1250.6 & 1302.8 & 1392.1 & 1439.3 & 1464.7 \\
\bottomrule

\end{tabular}
}
\caption{Number of input tokens on Math-hard.}
\label{tab:app_input_token_length_math}
\end{table}

\begin{table}[H]
\centering
\resizebox{\linewidth}{!}{
\begin{tabular}{llrrrrrrrrrr}
\toprule
& \multirow{2}{*}{\textbf{Method}}
& \multicolumn{10}{c}{\textbf{Number of turns}} \\
\cmidrule{3-12}
& & \textbf{0} & \textbf{1} & \textbf{2} & \textbf{3} & \textbf{4} & \textbf{5} & \textbf{6} & \textbf{7} & \textbf{8} & \textbf{9} \\
\midrule

% ---------------- Llama ----------------------
\multirow{3}{*}{\rotatebox{90}{Llama}}
& Multi-turn & 165.1 & 800.4 & 1473.2 & 2164.4 & 2871.0 & 3594.4 & 4274.1 & 4962.2 & 5578.6 & 6248.8 \\
& Memory     & 165.1 & 829.6 & 973.4 & 1157.5 & 1209.8 & 1224.9 & 1256.0 & 1237.0 & 1249.4 & 1279.0 \\
& In-place   & 165.1 & 422.1 & 487.9 & 534.0 & 592.6 & 625.2 & 665.0 & 638.6 & 677.9 & 670.8 \\
\midrule

% ---------------- Gemma ----------------------
\multirow{3}{*}{\rotatebox{90}{Gemma}}
& Multi-turn & 149.0 & 975.2 & 1770.5 & 2545.2 & 3392.5 & 4187.6 & 4961.9 & 5791.9 & 6656.0 & 7565.4 \\
& Memory     & 149.0 & 1005.2 & 1078.5 & 1191.2 & 1210.4 & 1217.3 & 1256.2 & 1272.8 & 1267.5 & 1285.5 \\
& In-place   & 149.0 & 410.0 & 579.0 & 658.5 & 750.2 & 830.5 & 841.1 & 802.7 & 863.0 & 863.6 \\
\midrule

% ---------------- Qwen ----------------------
\multirow{3}{*}{\rotatebox{90}{Qwen}}
& Multi-turn & 144.8 & 918.2 & 1809.5 & 2724.6 & 3698.0 & 4786.0 & 5804.9 & 6891.6 & 7984.2 & 8977.5 \\
& Memory     & 144.8 & 947.4 & 1125.8 & 1282.0 & 1318.7 & 1333.7 & 1367.8 & 1345.4 & 1415.4 & 1449.1 \\
& In-place   & 144.8 & 519.4 & 625.5 & 764.1 & 794.8 & 833.9 & 807.9 & 861.3 & 886.9 & 882.1 \\
\bottomrule

\end{tabular}
}
\caption{Number of input tokens on MMLU-pro.}
\label{tab:app_input_token_length_mmlu_pro}
\end{table}

\begin{table}[H]
\centering
\resizebox{\linewidth}{!}{
\begin{tabular}{llrrrrrrrrrr}
\toprule
& \multirow{2}{*}{\textbf{Method}}
& \multicolumn{10}{c}{\textbf{Number of turns}} \\
\cmidrule{3-12}
& & \textbf{0} & \textbf{1} & \textbf{2} & \textbf{3} & \textbf{4} & \textbf{5} & \textbf{6} & \textbf{7} & \textbf{8} & \textbf{9} \\
\midrule

% ---------------- Llama ----------------------
\multirow{3}{*}{\rotatebox{90}{Llama}}
& Multi-turn & 224.1 & 1054.1 & 1859.6 & 2744.6 & 3635.5 & 4503.7 & 5393.5 & 6302.9 & 7250.4 & 8134.1 \\
& Memory     & 224.1 & 1083.4 & 1181.7 & 1394.7 & 1389.8 & 1360.0 & 1377.9 & 1354.2 & 1319.9 & 1340.8 \\
& In-place   & 224.1 & 522.0 & 611.1 & 689.5 & 725.9 & 740.0 & 770.9 & 806.7 & 890.8 & 876.4 \\
\midrule

% ---------------- Gemma ----------------------
\multirow{3}{*}{\rotatebox{90}{Gemma}}
& Multi-turn & 205.0 & 1197.2 & 2138.4 & 3073.1 & 3996.3 & 4913.9 & 5850.2 & 6865.6 & 7829.9 & 8823.9 \\
& Memory     & 205.0 & 1227.2 & 1322.4 & 1482.0 & 1466.4 & 1508.8 & 1522.3 & 1498.5 & 1524.4 & 1503.4 \\
& In-place   & 205.0 & 480.3 & 666.0 & 680.4 & 755.3 & 848.7 & 932.5 & 848.2 & 1004.4 & 980.0 \\
\midrule

% ---------------- Qwen ----------------------
\multirow{3}{*}{\rotatebox{90}{Qwen}}
& Multi-turn & 203.6 & 1079.9 & 2013.6 & 2982.9 & 4127.7 & 5304.3 & 6395.3 & 7570.8 & 8763.9 & 9987.4 \\
& Memory     & 203.6 & 1109.2 & 1258.6 & 1484.6 & 1521.4 & 1541.4 & 1544.0 & 1519.1 & 1511.7 & 1525.8 \\
& In-place   & 203.6 & 553.7 & 678.3 & 771.6 & 759.5 & 766.1 & 847.4 & 848.5 & 876.1 & 966.4 \\
\bottomrule

\end{tabular}
}
\caption{Number of input tokens on GPQA.}
\label{tab:app_input_token_length_gpqa}
\end{table}

\begin{table}[H]
\centering
\resizebox{\linewidth}{!}{
\begin{tabular}{llrrrrrrrrrr}
\toprule
& \multirow{2}{*}{\textbf{Method}}
& \multicolumn{10}{c}{\textbf{Number of turns}} \\
\cmidrule{3-12}
& & \textbf{0} & \textbf{1} & \textbf{2} & \textbf{3} & \textbf{4} & \textbf{5} & \textbf{6} & \textbf{7} & \textbf{8} & \textbf{9} \\
\midrule

% ---------------- Llama ----------------------
\multirow{3}{*}{\rotatebox{90}{Llama}}
& Multi-turn & 932.3 & 2748.8 & 4089.0 & 5760.6 & 7508.6 & 9251.7 & 11123.1 & 12914.4 & 14775.8 & 16653.6 \\
& Memory     & 932.3 & 2700.9 & 2275.6 & 2312.4 & 2370.3 & 2402.9 & 2433.7 & 2439.4 & 2434.0 & 2386.2 \\
& In-place   & 932.3 & 2166.4 & 2709.9 & 3171.8 & 3641.0 & 4069.7 & 4551.9 & 5079.0 & 5519.2 & 5878.4 \\
\midrule
% ---------------- Gemma ----------------------
\multirow{3}{*}{\rotatebox{90}{Gemma}}
& Multi-turn & 1004.4 & 2249.6 & 3464.7 & 4709.9 & 5963.8 & 7295.7 & 8601.6 & 9843.2 & 11092.1 & 12269.7 \\
& Memory     & 1004.4 & 2171.6 & 2341.8 & 2414.9 & 2479.8 & 2470.6 & 2490.3 & 2520.4 & 2534.1 & 2525.8 \\
& In-place   & 1004.4 & 1969.5 & 2286.8 & 2610.3 & 2935.5 & 3262.7 & 3544.8 & 3786.9 & 3983.6 & 4258.2 \\
\midrule
% ---------------- Qwen ----------------------
\multirow{3}{*}{\rotatebox{90}{Qwen}}
& Multi-turn & 897.8 & 2144.7 & 3342.6 & 4579.8 & 5844.5 & 7108.0 & 8416.9 & 9860.3 & 11128.0 & 12343.3 \\
& Memory     & 897.8 & 2128.5 & 2152.0 & 2196.5 & 2212.2 & 2190.5 & 2242.1 & 2234.0 & 2217.8 & 2277.1 \\
& In-place   & 897.8 & 1878.6 & 2100.1 & 2346.6 & 2635.3 & 2941.3 & 3184.2 & 3381.4 & 3511.4 & 3493.4 \\
\bottomrule

\end{tabular}
}
\caption{Number of input tokens on ZebraLogic.}
\label{tab:app_input_token_length_zebralogic}
\end{table}

\subsection{Number of Generated Tokens}

\begin{table}[H]
\centering
\resizebox{\linewidth}{!}{
\begin{tabular}{llrrrrrrrrrr}
\toprule
& \multirow{2}{*}{\textbf{Method}}
& \multicolumn{10}{c}{\textbf{Number of turns}} \\
\cmidrule{3-12}
& & \textbf{0} & \textbf{1} & \textbf{2} & \textbf{3} & \textbf{4} & \textbf{5} & \textbf{6} & \textbf{7} & \textbf{8} & \textbf{9} \\
\midrule

% ---------------- Llama ----------------------
\multirow{3}{*}{\rotatebox{90}{Llama}}
& Multi-turn & 395.2 & 504.4 & 587.3 & 661.0 & 730.8 & 734.3 & 792.9 & 809.8 & 852.6 & 923.6 \\
& Memory     & 395.2 & 479.1 & 588.1 & 647.8 & 696.7 & 705.3 & 743.1 & 790.1 & 844.8 & 861.5 \\
& In-place   & 395.2 & 294.1 & 307.8 & 331.9 & 328.6 & 351.5 & 373.8 & 388.6 & 364.8 & 432.4 \\
\midrule

% ---------------- Gemma ----------------------
\multirow{3}{*}{\rotatebox{90}{Gemma}}
& Multi-turn & 1138.3 & 1152.2 & 1098.2 & 1074.0 & 1070.4 & 1087.0 & 1063.4 & 1091.2 & 1081.9 & 1119.2 \\
& Memory     & 1138.3 & 907.0 & 961.6 & 986.1 & 987.7 & 984.3 & 999.7 & 1041.3 & 1013.6 & 992.2 \\
& In-place   & 1138.3 & 1042.9 & 838.6 & 746.1 & 761.0 & 758.3 & 681.3 & 617.9 & 554.4 & 583.3 \\
\midrule

% ---------------- Qwen ----------------------
\multirow{3}{*}{\rotatebox{90}{Qwen}}
& Multi-turn & 810.9 & 925.4 & 1072.1 & 1186.2 & 1276.7 & 1349.3 & 1360.0 & 1408.0 & 1469.4 & 1525.9 \\
& Memory     & 810.9 & 850.7 & 945.5 & 987.7 & 1011.2 & 1020.3 & 1033.4 & 1031.4 & 1050.0 & 1071.5 \\
& In-place   & 810.9 & 403.1 & 320.2 & 277.7 & 283.0 & 232.6 & 224.2 & 172.0 & 166.8 & 185.7 \\
\bottomrule

\end{tabular}
}
\caption{Number of generated tokens on Math-hard.}
\label{tab:app_output_token_length_math}
\end{table}

\begin{table}[H]
\centering
\resizebox{\linewidth}{!}{
\begin{tabular}{llrrrrrrrrrr}
\toprule
& \multirow{2}{*}{\textbf{Method}}
& \multicolumn{10}{c}{\textbf{Number of turns}} \\
\cmidrule{3-12}
& & \textbf{0} & \textbf{1} & \textbf{2} & \textbf{3} & \textbf{4} & \textbf{5} & \textbf{6} & \textbf{7} & \textbf{8} & \textbf{9} \\
\midrule

% ---------------- Llama ----------------------
\multirow{3}{*}{\rotatebox{90}{Llama}}
& Multi-turn & 381.4 & 432.4 & 477.4 & 480.0 & 499.4 & 537.3 & 537.0 & 514.7 & 518.2 & 546.3 \\
& Memory     & 381.4 & 429.7 & 515.0 & 574.6 & 603.1 & 627.7 & 617.2 & 650.2 & 662.3 & 642.3 \\
& In-place   & 381.4 & 236.2 & 236.7 & 206.9 & 214.0 & 212.8 & 188.0 & 249.9 & 223.5 & 253.1 \\
\midrule

% ---------------- Gemma ----------------------
\multirow{3}{*}{\rotatebox{90}{Gemma}}
& Multi-turn & 551.7 & 567.9 & 562.3 & 585.7 & 603.2 & 635.0 & 641.1 & 662.3 & 683.4 & 683.2 \\
& Memory     & 551.7 & 564.8 & 600.9 & 593.8 & 614.2 & 655.1 & 664.4 & 667.3 & 688.2 & 696.4 \\
& In-place   & 551.7 & 505.6 & 475.5 & 468.4 & 398.9 & 414.5 & 395.9 & 455.7 & 399.0 & 384.0 \\
\midrule

% ---------------- Qwen ----------------------
\multirow{3}{*}{\rotatebox{90}{Qwen}}
& Multi-turn & 456.1 & 614.2 & 689.3 & 757.0 & 813.6 & 851.4 & 860.8 & 907.1 & 919.7 & 943.1 \\
& Memory     & 456.1 & 597.1 & 678.3 & 721.4 & 736.5 & 780.6 & 766.5 & 794.5 & 847.6 & 833.7 \\
& In-place   & 456.1 & 280.4 & 305.4 & 265.8 & 269.9 & 289.9 & 336.0 & 298.2 & 320.2 & 308.2 \\
\bottomrule

\end{tabular}
}
\caption{Number of generated tokens on MMLU-pro.}
\label{tab:app_output_token_length_mmlu_pro}
\end{table}

\begin{table}[H]
\centering
\resizebox{\linewidth}{!}{
\begin{tabular}{llrrrrrrrrrr}
\toprule
& \multirow{2}{*}{\textbf{Method}}
& \multicolumn{10}{c}{\textbf{Number of turns}} \\
\cmidrule{3-12}
& & \textbf{0} & \textbf{1} & \textbf{2} & \textbf{3} & \textbf{4} & \textbf{5} & \textbf{6} & \textbf{7} & \textbf{8} & \textbf{9} \\
\midrule

% ---------------- Llama ----------------------
\multirow{3}{*}{\rotatebox{90}{Llama}}
& Multi-turn & 665.7 & 644.4 & 659.2 & 683.2 & 721.5 & 699.2 & 697.8 & 689.6 & 697.8 & 690.8 \\
& Memory     & 665.7 & 614.8 & 671.1 & 678.5 & 667.1 & 673.4 & 660.3 & 646.5 & 651.6 & 627.4 \\
& In-place   & 665.7 & 421.3 & 351.0 & 283.8 & 312.3 & 297.0 & 292.7 & 306.9 & 241.3 & 266.8 \\
\midrule

% ---------------- Gemma ----------------------
\multirow{3}{*}{\rotatebox{90}{Gemma}}
& Multi-turn & 822.6 & 811.8 & 756.1 & 731.8 & 773.5 & 736.9 & 796.8 & 809.3 & 806.9 & 820.0 \\
& Memory     & 822.6 & 796.5 & 797.6 & 774.2 & 815.5 & 837.2 & 816.6 & 835.6 & 823.3 & 815.7 \\
& In-place   & 822.6 & 625.9 & 514.9 & 533.5 & 516.3 & 472.7 & 376.3 & 462.9 & 315.5 & 386.8 \\
\midrule

% ---------------- Qwen ----------------------
\multirow{3}{*}{\rotatebox{90}{Qwen}}
& Multi-turn & 715.5 & 751.6 & 832.8 & 860.5 & 946.8 & 964.7 & 1040.5 & 1035.5 & 1067.5 & 1068.6 \\
& Memory     & 715.5 & 709.6 & 787.0 & 824.5 & 847.2 & 853.0 & 835.1 & 815.3 & 837.1 & 850.9 \\
& In-place   & 715.5 & 399.5 & 332.2 & 270.9 & 272.5 & 275.7 & 245.1 & 253.2 & 247.8 & 192.1 \\
\bottomrule

\end{tabular}
}
\caption{Number of generated tokens on GPQA.}
\label{tab:app_output_token_length_gpqa}
\end{table}

\begin{table}[H]
\centering
\resizebox{\linewidth}{!}{
\begin{tabular}{llrrrrrrrrrr}
\toprule
& \multirow{2}{*}{\textbf{Method}}
& \multicolumn{10}{c}{\textbf{Number of turns}} \\
\cmidrule{3-12}
& & \textbf{0} & \textbf{1} & \textbf{2} & \textbf{3} & \textbf{4} & \textbf{5} & \textbf{6} & \textbf{7} & \textbf{8} & \textbf{9} \\
\midrule

% ---------------- Llama ----------------------
\multirow{3}{*}{\rotatebox{90}{Llama}}
& Multi-turn & 1685.8 & 1237.7 & 1502.3 & 1597.5 & 1535.5 & 1558.8 & 1558.4 & 1623.8 & 1572.2 & 1600.6 \\
& Memory     & 1685.8 & 1181.6 & 1118.8 & 1164.6 & 1182.5 & 1176.5 & 1175.7 & 1169.2 & 1133.1 & 1144.5 \\
& In-place   & 1685.8 & 746.2 & 566.9 & 523.3 & 501.3 & 528.7 & 529.6 & 563.8 & 582.7 & 662.8 \\
\midrule
% ---------------- Gemma ----------------------
\multirow{3}{*}{\rotatebox{90}{Gemma}}
& Multi-turn & 1123.1 & 1073.6 & 1090.5 & 1096.2 & 1104.9 & 1118.8 & 1129.5 & 1140.4 & 1147.1 & 1143.0 \\
& Memory     & 1123.1 & 1207.0 & 1194.1 & 1263.0 & 1251.1 & 1271.5 & 1299.5 & 1313.1 & 1304.8 & 1318.6 \\
& In-place   & 1123.1 & 454.8 & 427.6 & 425.8 & 427.7 & 436.6 & 454.6 & 479.6 & 477.8 & 510.7 \\
\midrule
% ---------------- Qwen ----------------------
\multirow{3}{*}{\rotatebox{90}{Qwen}}
& Multi-turn & 1135.0 & 1081.0 & 1085.3 & 1089.1 & 1104.9 & 1117.3 & 1130.4 & 1152.4 & 1158.7 & 1169.2 \\
& Memory     & 1135.0 & 1074.9 & 1021.9 & 1019.1 & 1006.9 & 1024.0 & 1029.1 & 996.6 & 1039.3 & 1071.6 \\
& In-place   & 1135.0 & 333.1 & 322.4 & 333.0 & 358.4 & 381.3 & 403.5 & 415.0 & 416.9 & 405.1 \\
\bottomrule

\end{tabular}
}
\caption{Number of generated tokens on ZebraLogic.}
\label{tab:app_output_token_length_zebralogic}
\end{table}

\subsection{Results on ZebraLogic}

\begin{table}[H]
\centering
\resizebox{\linewidth}{!}{
\begin{tabular}{llrrrrrrrrrr}
\toprule
& \multirow{2}{*}{\textbf{Method}}
& \multicolumn{10}{c}{\textbf{Number of turns}} \\
\cmidrule{3-12}
& & \textbf{0} & \textbf{1} & \textbf{2} & \textbf{3} & \textbf{4} & \textbf{5} & \textbf{6} & \textbf{7} & \textbf{8} & \textbf{9} \\
\midrule

% ---------------- Gemma ----------------------
\multirow{3}{*}{\rotatebox{90}{Gemma}}
& Multi-turn & 0.000 & 0.028 & 0.078 & 0.148 & 0.271 & 0.421 & 0.556 & 0.650 & 0.728 & 0.771 \\
& Memory     & 0.000 & 0.007 & 0.028 & 0.052 & 0.078 & 0.105 & 0.135 & 0.167 & 0.199 & 0.234 \\
& In-place    & 0.000 & 0.031 & 0.115 & 0.265 & 0.441 & 0.582 & 0.712 & 0.816 & 0.868 & 0.906 \\
\midrule

% ---------------- Qwen ----------------------
\multirow{3}{*}{\rotatebox{90}{Qwen}}
& Multi-turn & 0.003 & 0.022 & 0.071 & 0.150 & 0.249 & 0.356 & 0.468 & 0.566 & 0.645 & 0.690 \\
& Memory     & 0.004 & 0.025 & 0.073 & 0.166 & 0.251 & 0.348 & 0.434 & 0.520 & 0.600 & 0.665 \\
& In-place    & 0.004 & 0.036 & 0.144 & 0.341 & 0.534 & 0.672 & 0.779 & 0.850 & 0.895 & 0.922 \\
\midrule

% ---------------- Llama ----------------------
\multirow{3}{*}{\rotatebox{90}{Llama}}
& Multi-turn & 0.006 & 0.017 & 0.056 & 0.126 & 0.225 & 0.358 & 0.470 & 0.551 & 0.612 & 0.652 \\
& Memory     & 0.006 & 0.023 & 0.076 & 0.160 & 0.264 & 0.369 & 0.457 & 0.543 & 0.625 & 0.682 \\
& In-place    & 0.006 & 0.034 & 0.105 & 0.223 & 0.359 & 0.478 & 0.590 & 0.673 & 0.738 & 0.790 \\
\bottomrule

\end{tabular}
}
\caption{Grid accuracy of LLMs on ZebraLogic with top-$4$ feedback.}
\label{tab:app_game_acc_top4}
\end{table}

\begin{table}[H]
\centering
\resizebox{\linewidth}{!}{
\begin{tabular}{llrrrrrrrrrr}
\toprule
& \multirow{2}{*}{\textbf{Method}}
& \multicolumn{10}{c}{\textbf{Number of turns}} \\
\cmidrule{3-12}
& & \textbf{0} & \textbf{1} & \textbf{2} & \textbf{3} & \textbf{4} & \textbf{5} & \textbf{6} & \textbf{7} & \textbf{8} & \textbf{9} \\
\midrule

% ---------------- Gemma ----------------------
\multirow{3}{*}{\rotatebox{90}{Gemma}}
& Multi-turn & 0.000 & 0.013 & 0.035 & 0.070 & 0.109 & 0.156 & 0.208 & 0.266 & 0.325 & 0.390 \\
& Memory     & 0.000 & 0.005 & 0.018 & 0.033 & 0.044 & 0.058 & 0.078 & 0.089 & 0.101 & 0.120 \\
& In-place    & 0.000 & 0.013 & 0.057 & 0.103 & 0.167 & 0.252 & 0.334 & 0.421 & 0.494 & 0.582 \\
\midrule

% ---------------- Qwen ----------------------
\multirow{3}{*}{\rotatebox{90}{Qwen}}
& Multi-turn & 0.004 & 0.013 & 0.029 & 0.059 & 0.093 & 0.136 & 0.188 & 0.229 & 0.276 & 0.322 \\
& Memory     & 0.004 & 0.015 & 0.032 & 0.071 & 0.119 & 0.157 & 0.217 & 0.278 & 0.342 & 0.395 \\
& In-place    & 0.004 & 0.020 & 0.048 & 0.114 & 0.194 & 0.286 & 0.382 & 0.471 & 0.552 & 0.624 \\
\midrule

% ---------------- Llama ----------------------
\multirow{3}{*}{\rotatebox{90}{Llama}}
& Multi-turn & 0.006 & 0.018 & 0.040 & 0.073 & 0.125 & 0.183 & 0.250 & 0.324 & 0.385 & 0.433 \\
& Memory     & 0.005 & 0.017 & 0.044 & 0.089 & 0.143 & 0.193 & 0.248 & 0.303 & 0.339 & 0.385 \\
& In-place    & 0.006 & 0.021 & 0.060 & 0.110 & 0.174 & 0.258 & 0.328 & 0.391 & 0.448 & 0.502 \\
\bottomrule

\end{tabular}
}
\caption{Grid accuracy of LLMs on ZebraLogic with top-$2$ feedback.}
\label{tab:app_game_acc_top2}
\end{table}

\begin{table}[H]
\centering
\resizebox{\linewidth}{!}{
\begin{tabular}{llrrrrrrrrrr}
\toprule
& \multirow{2}{*}{\textbf{Method}}
& \multicolumn{10}{c}{\textbf{Number of turns}} \\
\cmidrule{3-12}
& & \textbf{0} & \textbf{1} & \textbf{2} & \textbf{3} & \textbf{4} & \textbf{5} & \textbf{6} & \textbf{7} & \textbf{8} & \textbf{9} \\
\midrule

% ---------------- Gemma ----------------------
\multirow{3}{*}{\rotatebox{90}{Gemma}}
& Multi-turn & 0.272 & 0.435 & 0.567 & 0.681 & 0.779 & 0.848 & 0.896 & 0.927 & 0.947 & 0.959 \\
& Memory     & 0.272 & 0.350 & 0.411 & 0.464 & 0.505 & 0.528 & 0.548 & 0.571 & 0.586 & 0.603 \\
& In-place    & 0.272 & 0.432 & 0.573 & 0.698 & 0.794 & 0.859 & 0.914 & 0.944 & 0.963 & 0.972 \\
\midrule

% ---------------- Qwen ----------------------
\multirow{3}{*}{\rotatebox{90}{Qwen}}
& Multi-turn & 0.263 & 0.417 & 0.543 & 0.652 & 0.736 & 0.806 & 0.859 & 0.891 & 0.921 & 0.935 \\
& Memory     & 0.264 & 0.406 & 0.518 & 0.618 & 0.690 & 0.750 & 0.794 & 0.835 & 0.866 & 0.891 \\
& In-place    & 0.264 & 0.443 & 0.607 & 0.738 & 0.830 & 0.891 & 0.931 & 0.956 & 0.968 & 0.977 \\
\midrule

% ---------------- Llama ----------------------
\multirow{3}{*}{\rotatebox{90}{Llama}}
& Multi-turn & 0.191 & 0.185 & 0.366 & 0.460 & 0.546 & 0.612 & 0.671 & 0.663 & 0.701 & 0.718 \\
& Memory     & 0.191 & 0.339 & 0.487 & 0.599 & 0.693 & 0.758 & 0.807 & 0.841 & 0.871 & 0.889 \\
& In-place    & 0.191 & 0.367 & 0.480 & 0.591 & 0.679 & 0.751 & 0.798 & 0.833 & 0.847 & 0.883 \\
\bottomrule

\end{tabular}
}
\caption{Cell accuracy of LLMs on ZebraLogic with top-$4$ feedback.}
\label{tab:app_cell_acc_top4}
\end{table}

\begin{table}[H]
\centering
\resizebox{\linewidth}{!}{
\begin{tabular}{llrrrrrrrrrr}
\toprule
& \multirow{2}{*}{\textbf{Method}}
& \multicolumn{10}{c}{\textbf{Number of turns}} \\
\cmidrule{3-12}
& & \textbf{0} & \textbf{1} & \textbf{2} & \textbf{3} & \textbf{4} & \textbf{5} & \textbf{6} & \textbf{7} & \textbf{8} & \textbf{9} \\
\midrule

% ---------------- Gemma ----------------------
\multirow{3}{*}{\rotatebox{90}{Gemma}}
& Multi-turn & 0.272 & 0.367 & 0.441 & 0.511 & 0.575 & 0.629 & 0.679 & 0.723 & 0.761 & 0.793 \\
& Memory     & 0.272 & 0.311 & 0.347 & 0.376 & 0.401 & 0.428 & 0.445 & 0.461 & 0.482 & 0.501 \\
& In-place    & 0.272 & 0.359 & 0.434 & 0.510 & 0.580 & 0.642 & 0.702 & 0.754 & 0.793 & 0.832 \\
\midrule

% ---------------- Qwen ----------------------
\multirow{3}{*}{\rotatebox{90}{Qwen}}
& Multi-turn & 0.264 & 0.350 & 0.421 & 0.482 & 0.538 & 0.585 & 0.629 & 0.666 & 0.698 & 0.730 \\
& Memory     & 0.264 & 0.339 & 0.410 & 0.471 & 0.529 & 0.581 & 0.628 & 0.669 & 0.703 & 0.735 \\
& In-place    & 0.264 & 0.361 & 0.452 & 0.538 & 0.614 & 0.682 & 0.742 & 0.789 & 0.828 & 0.858 \\
\midrule

% ---------------- Llama ----------------------
\multirow{3}{*}{\rotatebox{90}{Llama}}
& Multi-turn & 0.191 & 0.289 & 0.331 & 0.404 & 0.452 & 0.514 & 0.543 & 0.595 & 0.608 & 0.649 \\
& Memory     & 0.195 & 0.283 & 0.373 & 0.458 & 0.517 & 0.573 & 0.609 & 0.637 & 0.668 & 0.705 \\
& In-place    & 0.191 & 0.311 & 0.376 & 0.451 & 0.506 & 0.566 & 0.605 & 0.645 & 0.674 & 0.705 \\
\bottomrule

\end{tabular}
}
\caption{Cell accuracy of LLMs on ZebraLogic with top-$2$ feedback.}
\label{tab:app_cell_acc_top2}
\end{table}

\section{Qualitative Examples}

\subsection{In-Place Feedback Example}
\label{app.in_place}

\cref{fig:inplace_math1,fig:inplace_math2,fig:inplace_gpqa,fig:inplace_mmlu} are the qualitative examples of in-place feedback on the three benchmarks.

\subsection{Multi-Turn Feedback Failure Example}
\label{app.multi_turn_failure_example}

We observe failure cases of multi-turn feedback, and present the instances in \cref{fig:fail_case1,fig:fail_case2,fig:fail_case3}.

\begin{figure*}[h]
    \centering
    \begin{prompt}[Prompt format used for memory-based feedback]
    \small
        \textbf{USER PROMPT}\\
        Question: \textcolor{chart_blue}{\{QUESTION\}} \\
        Previous reasoning response: \textcolor{chart_blue}{\{PREVIOUS RESPONSE\}} \\
        Accumulated Feedback: \textcolor{chart_blue}{\{ACCUMULATED FEEDBACK\}}\\
        Please revise your previous reasoning response based on the accumulated feedback.
    \end{prompt}
    \caption{Prompt format used for memory-based feedback in MATH-hard, MMLU-pro free-form, GPQA free-form, and LiveCodeBench. The system prompt is identical to that in multi-turn feedback.}
    \label{fig.memory}
\end{figure*}

\begin{figure*}[h]
    \centering
    \begin{subfigure}[]{\textwidth}
        \centering
        \begin{prompt}[LLM-as-a-judge prompt for MATH-hard]
        \small
        \textbf{SYSTEM PROMPT}\\
        You are a mathematical equivalence judge. \\
        Given a question, a correct answer, and a model's prediction, determine if they are mathematically equivalent even if they have different formatting or representation. \\
        Respond with ONLY `YES' if they are equivalent, or `NO' if they are not. \\
        If the model's prediction is correct, but the formatting is wrong, please respond with `YES'.\\
        DO NOT RESPOND WITH ANYTHING ELSE.
        \tcblower
        \small
        \textbf{USER PROMPT}\\
        Question: \textcolor{chart_blue}{\{QUESTION\}}\\
        Correct answer: \textcolor{chart_blue}{\{CORRECT\_ANSWER\}}\\
        Model prediction: \textcolor{chart_blue}{\{PREDICTION\}}\\
        Are these equivalent? Answer YES or NO.
        
        \end{prompt}
        \caption{Prompt format used in MATH-hard.}
    \end{subfigure}

    \begin{subfigure}[]{\textwidth}
        \centering
        \begin{prompt}[LLM-as-a-judge prompt for MMLU-pro free-form and GPQA free-form]
        \small
        \textbf{SYSTEM PROMPT}
        
        You are a strict and impartial judge for evaluating model predictions. \\
        Given a question, a correct answer, and a model's prediction, decide whether the prediction is equivalent to the correct answer. \\
        Respond with ONLY `YES' or `NO'. Never output anything else. \\
        
        \textbf{Judgment Criteria}\\
        1. The prediction must fully answer the question and cannot be vague. \\
        2. Ignore differences in formatting, punctuation, capitalization, or spacing. \\
        3. For numeric answers:\\
            - Compute the relative error = $|\text{prediction} - \text{ground truth}|$ / mean(prediction, ground truth). \\
            - The prediction is correct if the relative error $<$ 1\%. \\
            - If the ground truth is a single numeric value but the prediction is a range, treat it as incorrect (even if the range contains the ground truth).\\
            - If the ground truth is a single value but the prediction is a range, output `NO'. \\
        4. If the prediction is correct in substance but expressed in a different format, respond `YES'. \\
        
        Final Rule: Output must be exactly `YES' or `NO'. No explanations, no extra words.
        \tcblower
        \small
        \textbf{USER PROMPT}\\
        Question: \textcolor{chart_blue}{\{QUESTION\}}\\
        Correct answer: \textcolor{chart_blue}{\{CORRECT\_ANSWER\}}\\
        Model prediction: \textcolor{chart_blue}{\{PREDICTION\}}\\
        Are these equivalent? Answer YES or NO.
        
        \end{prompt}
        \caption{Prompt format used in MMLU-pro free-form and GPQA free-form.}
    \end{subfigure}
    
    \caption{Prompt format used for LLM-as-a-judge in MATH-hard, MMLU-pro free-form, and GPQA free-form.}
    \label{fig:judge_prompt}
\end{figure*}

\begin{figure*}[h]
    \centering
    \begin{subfigure}[]{\textwidth}
    \begin{prompt}[Prompt format used for feedback agent in MATH-hard, MMLU-pro free-form, and GPQA free-form]
    \small
        \textbf{SYSTEM PROMPT} \\
        You are a rigorous verifier of LLM answers. Your task is to identify the first and most critical reasoning flaw that causes the error, given a question, the model’s full answer, and the correct answer. \\
        You MUST NOT reveal or speculate about the correct final answer.\\
         - Note that the options in previous questions might have been switched in each different attempt.\\
        \textbf{Your output must follow the structure below:}\\
        Explain in (maximum 2–3) precise sentences why that step is incorrect. Be specific in your critique.\\
        Give short and specific actionable advice to fix it.\\
        \textbf{Constraints:}\\
        You may only give feedback on one reasoning error at a time — the earliest critical one.\\
        If the model's answer is correct in logic but wrong in formatting, your feedback should address the formatting issue.\\
        You DO NOT leak and mention the correct final answer.\\
        \textbf{DO NOT SOLVE THE PROBLEM, JUST PROVIDE FEEDBACK WITHOUT MENTIONING THE CORRECT ANSWER}\\
        Your output strictly starting with ``Your output is wrong because'' and then followed by the explanation.\\
        Do not output anything else.
        \tcblower
        \small       
        \textbf{USER PROMPT}\\
       Your goal is to identify the \textbf{earliest and most critical logical flaw} in the model’s reasoning\\
        DO NOT provide or hint at the correct final answer.\\
        - Note that the options in previous questions might have been switched in each different attempt.\\
        \textbf{Problem}\\
        \textcolor{chart_blue}{\{PROBLEM\}}\\
        \textbf{Most Recent Model Answer}\\
        \textcolor{chart_blue}{\{MODEL ANSWER\}}\\
        \textbf{Correct Final Answer} \\
        \textcolor{chart_blue}{\{GROUND TRUTH\}}\\
        Based on this, provide feedback on the single most important error in the model’s answer.\\
        Do not leak and mention the correct final answer and do not add any extra commentary.
        \end{prompt}
        
        \vspace{-0.5 em}
        \caption{Prompt format used for feedback agent in MATH-hard, MMLU-pro free-form, and GPQA free-form.}
        \label{fig.feedback_math}
        \end{subfigure}

    \begin{subfigure}[]{\textwidth}
    \begin{prompt}[Prompt format used for feedback agent in LiveCodeBench]
    \small
        \textbf{SYSTEM PROMPT} \\
        You are a rigorous verifier of coding solutions. Your task is to identify the first and most critical logical flaw in the code that causes test failures, given a coding problem, the model's submitted code.\\
        You MUST NOT provide the complete corrected code.\\
        \textbf{Your output must follow this structure:}\\
        1. Start with ``Your output is wrong because''\\
        2. Explain what is wrong\\
        3. End with how to fix it.\\
        \textbf{Example:}\\
        ``Your output is wrong because the loop breaks after finding the first valid partition, but the problem requires finding the minimum among all possible partitions.\\
        Iterate through all possible split points and track the minimum maximum sum using min().''\\
        \textbf{Constraints:}\\
        You may only give feedback on one coding error at a time — the earliest critical one.\\
        Be specific about which part of the code is problematic (mention line numbers or code patterns if helpful).\\
        Do not mention the complete implementation.\\
        DO NOT provide the full corrected code, and do not hard-code the example input/output in the code.\\
        Do not output anything else.
        \tcblower
        \small       
        \textbf{USER PROMPT}\\
        Your goal is to identify the \textbf{earliest and most critical logical flaw} in the model’s reasoning\\
        DO NOT provide the complete solution.\\
        - Note that the options in previous questions might have been switched in each different attempt.\\
        \textbf{Problem}\\
        \textcolor{chart_blue}{\{PROBLEM\}}\\
        \textbf{Most Recent Model Answer}\\
        \textcolor{chart_blue}{\{MODEL ANSWER\}}\\

        \end{prompt}
        \vspace{-0.5 em}
        \caption{Prompt format used in LiveCodeBench.}
    
    \label{fig.feedback_code}
    \end{subfigure}
    \vspace{-1 em}
    \caption{Prompt format used for the feedback agent in empirical experiments.}
    \label{fig.feedback}
\end{figure*}

\begin{figure*}[h]
    \centering
    \begin{prompt}[Prompt format used for in-place feedback agent in MATH-hard, MMLU-pro free-form, and GPQA free-form]
    \small
        \textbf{SYSTEM PROMPT} \\
        You are a helpful assistant who INTERVENES in a math solution based on user feedback.\\
        Your job is to produce a JSON object for a single \textbf{in-place replace} operation:\\
        - Identify the \textbf{shortest unique substring (SUS)} from the original solution that must be edited to apply the feedback.\\
        - Produce the revised text for that exact span.\\
        - \textbf{Do NOT} change anything before the flaw, and \textbf{do NOT} continue solving the problem beyond where the feedback applies.\\
        - \textbf{Preserve} all whitespace, punctuation, LaTeX, and casing exactly as in the original solution for the target substring.\\
        - The target must be a \textbf{contiguous} substring that occurs \textbf{exactly once} in the original solution. If not unique, minimally extend the span (e.g., include adjacent tokens or punctuation) until it becomes unique. \\
        - Return \textbf{ONLY} valid JSON with UTF-8 and proper escaping (no trailing commas, no extra commentary). \\
        
        Return JSON with this schema (single edit only): \\ 

          \{\\
          \phantom{~~~~~~}``target\_sentence": ``$<$ exact shortest unique substring copied from the original$>$", \\
          \phantom{~~~~~~}``edit\_sentence": ``$<$the revised substring after applying the feedback$>$"\\
          \}\\

        Constraints:\\
        - Output must be a single-line or multi-line JSON object; do not include any extra text. \\
        - Do not normalize quotes/hyphens/spaces; copy exactly from the original for target\_sentence. \\
        - Do not introduce additional edits beyond the specified span. \\
        - Do not provide a reasoning process beyond the feedback.
        \tcblower
        \small
        \textbf{USER PROMPT}\\
        $<$The Start of Answer$>$ \\
        \textcolor{chart_blue}{\{ANSWER\}} \\
        $<$The End of Answer$>$ \\
        $<$The Start of Original Solution$>$\\
        \textcolor{chart_blue}{\{ORIGINAL\_SOLUTION\}}\\ 
        $<$The End of Original Solution$>$ \\
        $<$The Start of User Feedback$>$\\ 
        \textcolor{chart_blue}{\{USER\_FEEDBACK\}}\\ 
        $<$The End of User Feedback$>$\\
        $<$The Start of Instructions$>$\\
        Write the JSON according to the following:\\
        - Apply ONLY the given feedback to the original solution.\\
        - Identify the \textbf{shortest unique substring}  in the original that must change to satisfy the feedback; this must appear \textbf{exactly once}. \\
        - If the obvious sentence occurs multiple times, \textbf{minimally extend} the span (e.g., prepend/append one or two nearby tokens or punctuation) until uniqueness holds.\\
        - Put the original substring in ``target\_sentence" (copied \textbf{verbatim} from the original, including whitespace/newlines). \\
        - Put the corrected version in ``edit\_sentence". \\
        - If the feedback is sentence-like, keep it within \textbf{three sentences} in the edited span. \\
        - STRICTLY FOLLOW: \\
          - \textbf{Do not solve the problem} beyond where the feedback applies. \\
          - Stop right after applying the feedback. \\
          - Return \textbf{ONLY} valid JSON with keys ``target\_sentence" and ``edit\_sentence".
        
    \end{prompt}

    \caption{Prompt format used for in-place feedback agent in MATH-hard, MMLU-pro free-form, and GPQA free-form.}
    \label{fig.intervention_math}
\end{figure*}

\begin{figure*}[h]
    \centering
    \begin{prompt}[Prompt format used for in-place feedback agent in LiveCodeBench]
    \small
        \textbf{SYSTEM PROMPT} \\
        You are a rigorous verifier of code solutions. Your job is to produce a JSON object for a single \textbf{in-place replace} operation:\\
        - Identify the \textbf{shortest unique substring (SUS)} from the original solution that must be edited to apply the feedback.\\
        - Produce the revised text for that exact span.\\
        - The edit\_sentence must contain the whole feedback information, do not miss any information.\\
        - Ensure that the edit\_sentence matches the style and tone of the original solution, making it plausible to fit as a replacement.\\
        - \textbf{Do NOT}change anything before the flaw, and \textbf{do NOT} continue solving the problem beyond where the feedback applies.\\
        - \textbf{Preserve} all whitespace, indentation, and casing exactly as in the original solution for the target substring.\\
        - The target must be a \textbf{contiguous} substring that occurs \textbf{exactly once} in the original solution.\\
        
        \textbf{Important for code continuation:}\\
        - The edit\_sentence should end in a way that naturally invites continuation. Avoid ending with a complete function or return statement.\\
        - If the original flaw occurs mid-code, the edit\_sentence should correct the error and set up the next logical step without completing it.\\
        - Prefer ending with incomplete structures (e.g., opening a loop, starting a conditional, beginning a function body, or a partial expression).\\
        - Do NOT include any content after the corrected span; the model will generate the continuation.\\
        
        Return JSON with this schema (single edit only):\\

      \{\\
      \phantom{~~~~~~}``target\_sentence": ``$<$ exact shortest unique substring copied from the original$>$", \\
      \phantom{~~~~~~}``edit\_sentence": ``$<$the revised substring after applying the feedback$>$"\\
      \}\\
        
        Constraints:\\
        - Do not normalize quotes/hyphens/spaces; copy exactly from the original for target\_sentence.\\
        - The edit\_sentence must contain the whole feedback information, do not miss any information.\\
        - Ensure that the edit\_sentence matches the style and tone of the original solution, making it plausible to fit as a replacement.\\
        - Do not provide reasoning process beyond the feedback.\\
        - The edit\_sentence should be written as much as possible in code, not in comments. Only include comments when absolutely necessary for guidance.
        \tcblower
        \small
        \textbf{USER PROMPT}\\
        $<$The Start of Answer$>$ \\
        \textcolor{chart_blue}{\{ANSWER\}} \\
        $<$The End of Answer$>$ \\
        $<$The Start of Original Solution$>$\\
        \textcolor{chart_blue}{\{ORIGINAL\_SOLUTION\}}\\ 
        $<$The End of Original Solution$>$ \\
        $<$The Start of User Feedback$>$\\ 
        \textcolor{chart_blue}{\{USER\_FEEDBACK\}}\\ 
        $<$The End of User Feedback$>$\\
        $<$The Start of Instructions$>$\\
        Write the JSON according to the following:\\
        - Apply ONLY the given feedback to the original solution.\\
        - Identify the \textbf{shortest unique substring}  in the original that must change to satisfy the feedback; this must appear \textbf{exactly once}. \\
        - If the obvious sentence occurs multiple times, \textbf{minimally extend} the span (e.g., prepend/append one or two nearby tokens or punctuation) until uniqueness holds.\\
        - Put the original substring in ``target\_sentence'' (copied \textbf{verbatim} from the original, including whitespace/newlines). \\
        - Put the corrected version in ``edit\_sentence''. \\
        - If the feedback is sentence-like, keep it within \textbf{three sentences} in the edited span. \\
        - STRICTLY FOLLOW: \\
          - \textbf{Do not solve the problem} beyond where the feedback applies. \\
          - Stop right after applying the feedback. \\
          - Return \textbf{ONLY} valid JSON with keys ``target\_sentence'' and ``edit\_sentence''.
        
    \end{prompt}

    \caption{Prompt format used for in-place feedback agent in LiveCodeBench.}
    \label{fig.intervention_code}
\end{figure*}

\begin{figure*}[h]
    \centering
    \begin{prompt}[Prompt format used for ZebraLogic]
    \small
        \textbf{SYSTEM PROMPT} \\
         You are a helpful assistant that solves zebra puzzles. \\
        Given a puzzle and a json template, you need to solve the puzzle and fill in the json template. \\
        You need to fill in the json template with the correct attributes.
        \tcblower
        \small
        \textbf{USER PROMPT}\\
        \# Example Puzzle \\

        There are 3 houses, numbered 1 to 3 from left to right, as seen from across the street. Each house is occupied by a different person.\\ Each house has a unique attribute for each of the following characteristics:\\
         - Each person has a unique name: `Peter', `Eric', `Arnold'.\\
         - Each person has a unique favorite drink: `tea', `water', `milk'\\
         
        \#\# Clues for the Example Puzzle\\
        
        1. Peter is in the second house.\\
        2. Arnold is directly left of the one who only drinks water.\\
        3. The one who only drinks water is directly left of the person who likes milk.\\
        
        \#\# Answer to the Example Puzzle\\
        
        \{\{\\
   \phantom{~~~~~~}``solution": \{\{\\
   \phantom{~~~~~~}\phantom{~~~~~~}``House 1": \{\{\\
 \phantom{~~~~~~}\phantom{~~~~~~}\phantom{~~~~~~}``Name": ``Arnold",\\
 \phantom{~~~~~~}\phantom{~~~~~~}\phantom{~~~~~~}``Drink": ``tea"\\
  \phantom{~~~~~~}\phantom{~~~~~~}\}\},\\
  \phantom{~~~~~~}\phantom{~~~~~~}``House 2": \{\{\\
  \phantom{~~~~~~}\phantom{~~~~~~}\phantom{~~~~~~}``Name": ``Peter",\\
   \phantom{~~~~~~}\phantom{~~~~~~}\phantom{~~~~~~}``Drink": ``water"\\
  \phantom{~~~~~~}\phantom{~~~~~~}  \}\},\\
  \phantom{~~~~~~}\phantom{~~~~~~}``House 3": \{\{\\
   \phantom{~~~~~~}\phantom{~~~~~~}\phantom{~~~~~~} ``Name": ``Eric",\\
   \phantom{~~~~~~}\phantom{~~~~~~}\phantom{~~~~~~}``Drink": ``milk"\\
      \phantom{~~~~~~}\phantom{~~~~~~} \}\}\\
     \phantom{~~~~~~}\}\}\\
        \}\}\\
        
        \# Puzzle to Solve\\ 
        
        \textcolor{chart_blue}{\{PUZZLE\}}\\

        \# Instruction\\
        
        Now please solve the above puzzle. Present your reasoning and solution in the following json format:\\
        \textcolor{chart_blue}{\{JSON TEMPLATE\}}\\
    \end{prompt}
    
    \caption{Input template used for ZebraLogic.}
    \label{fig.input_zebra}
\end{figure*}

\begin{figure*}[h]
    \centering
    \begin{subfigure}[]{\textwidth}
        \centering
        \begin{prompt}[Feedback template for \texttt{Gemma} and \texttt{Qwen} in ZebraLogic]
        \small
        Your answer is incorrect. Please revise your solution based on the following feedback.\\
         - `\textcolor{chart_blue}{\{CATEGORY\}}' of the `\textcolor{chart_blue}{\{HOUSE\}}' is `\textcolor{chart_blue}{\{GROUND TRUTH\}}', not `\textcolor{chart_blue}{\{PREDICTION\}}'.\\
         - $\cdots$
        
        \end{prompt}
        \caption{Feedback template for \texttt{Gemma} and \texttt{Qwen}.}
    \end{subfigure}

    \vspace{1.5em}

    \begin{subfigure}[]{\textwidth}
        \centering
        \begin{prompt}[Feedback template for \texttt{Llama} in ZebraLogic]
        \small
        Please revise your step-by-step reasoning based on the following feedback, and then provide a solution in the following json format.\\ 
        Do not just provide the final solution, but also provide the reasoning process.\\
         - `\textcolor{chart_blue}{\{CATEGORY\}}' of the `\textcolor{chart_blue}{\{HOUSE\}}' is `\textcolor{chart_blue}{\{GROUND TRUTH\}}', not `\textcolor{chart_blue}{\{PREDICTION\}}'.\\
         - $\cdots$
        
        \end{prompt}
        \caption{Feedback template for \texttt{Llama}.}
    \end{subfigure}
    
    \caption{Rule-based feedback template in ZebraLogic.}
    \label{fig:feedback_zebra}
\end{figure*}

\begin{figure*}[h]
    \centering
    \begin{prompt}[Prompt format used for in-place feedback agent in ZebraLogic]
    \small
        \textbf{SYSTEM PROMPT} \\
        You are an in-place patcher. Given a reasoning step and feedback items, decide if the step conflicts with the feedback. If so, produce the minimally edited step that preserves all non-conflicting content and enforces the feedback exactly. \\
        
        LOCKED CLUE SPANS\\ 
        - Any substring beginning with `Clue' and ending at the first period is read-only. Do not alter or delete it.\\
        - If a step contains locked span(s), edit only after the last locked span. \\
        - If formatted as ``Clue N: ... . - editable text", edit only after the first ` - '. If editable text is deleted, remove the dangling ` - '.\\
        
        REFERENCE ATTRIBUTES \\
        \textcolor{chart_blue}{\{CATEGORY\}} \\
        - Use this only as a reference to detect conflicts and apply equivalence normalization. Do not generate new information beyond feedback. \\
        
        EDIT RULES\\
        - Feedback is always given in positive-only form, e.g., `attr' of the `house i' is `A'.\\
        - Always overwrite the attribute with the provided value A.\\
        - Modify only attributes mentioned in the feedback.\\
        - Do not add new reasoning, deductions, constraints, or explanations.\\ 
        - Do not insert additional sentences beyond the minimal correction.\\
        - Do not add new information, negations, or style fixes.\\
        - Equivalence normalization:\\
          * `name' of house i is X $\leftrightarrow$ X is in House \\
          * `musicgenre' of house i is Y,  The Y house is House i\\
          * Apply the same for other attributes.\\
        - Rewrite conflicts to the positive corrected assignment. Never use negations.\\
        
        CONFLICT CRITERIA\\
        A statement conflicts only if it:\\
        - Mentions the same (house, attribute) with a contradictory value, or\\
        - Assigns a feedback value to the wrong house by equivalence.\\
        
        DEPENDENCY PRUNING\\
        - If a sentence depends on a premise that is corrected or negated by the feedback and cannot be minimally rewritten, delete that sentence only.\\
        - Do not delete the entire reasoning step.\\
        - Never alter or delete locked clue text.\\
        
        OUTPUT\\
        - If no conflict: {{``conflict":``no"}}\\
        - If conflict: {{``conflict":``yes",``replacement":``..."}}\\
          where replacement is the fully patched step, including unchanged locked clues.\\
        
        RULES\\
        - Make the smallest edits necessary.\\
        - Preserve capitalization, spacing, and order.\\
        - Never output empty replacement if a locked clue exists.\\
        - Absolutely never add new reasoning, deductions, or commentary not present in the original.
        \tcblower
        \small
        \textbf{USER PROMPT}\\
        Reasoning step\\
        \textcolor{chart_blue}{\{REASONING\_STEP\}}\\
         Feedback\\
        \textcolor{chart_blue}{\{FEEDBACK\}}
        
    \end{prompt}
    \caption{Prompt format used for in-place feedback agent in ZebraLogic.}
    \label{appendix.intervention_zebra}
\end{figure*}

%%%%%%%%%%%%In-place feedback example - MATH-hard%%%%%%%%%%%%%%%%
\begin{figure*}[h]
    \centering
    \begin{inplace}[In-place feedback example - MATH-hard]
    \small
        \textbf{Problem} \\
        Express the infinite series $\frac{3}{206}+\frac{9}{2\cdot103^2}+\frac{27}{2\cdot103^3}+\cdots$as a terminating decimal. 
        \Sepline
        \small
        \textbf{Model response}\\
Let $S = \frac{3}{206} + \frac{9}{2 \cdot 103^2} + \frac{27}{2 \cdot 103^3} + \cdots$. We can rewrite this as $$S = \color{red}{\sum_{n=1}^{\infty} \frac{3^n}{2 \cdot 103^{n+1}}} = \frac{1}{2} \sum_{n=1}^{\infty} \frac{3^n}{103^{n+1}} \color{black}= \frac{1}{2} \sum_{n=1}^{\infty} \frac{3^n}{103 \cdot 103^n} = \frac{1}{2 \cdot 103} \sum_{n=1}^{\infty} \left(\frac{3}{103}\right)^n$$ The sum $\sum_{n=1}^{\infty} \left(\frac{3}{103}\right)^n$ is a geometric series with first term $\frac{3}{103}$ and common ratio $\frac{3}{103}$. Since $\left|\frac{3}{103}\right| < 1$, the series converges to $$\sum_{n=1}^{\infty} \left(\frac{3}{103}\right)^n = \frac{\frac{3}{103}}{1 - \frac{3}{103}} = \frac{\frac{3}{103}}{\frac{103-3}{103}} = \frac{\frac{3}{103}}{\frac{100}{103}} = \frac{3}{100}$$ Therefore, $S = \frac{1}{2 \cdot 103} \cdot \frac{3}{100} = \frac{3}{20600}$ Now we need to express $\frac{3}{20600}$ as a terminating decimal. We can divide 3 by 20600: $\frac{3}{20600} = 0.00014645 \dots$ We can write $\frac{3}{20600} = \frac{3}{206 \cdot 100} = \frac{3}{206} \cdot \frac{1}{100}$. We have $\frac{3}{206} \approx 0.01464539$. Then $\frac{3}{20600} = \frac{3}{206} \cdot \frac{1}{100} \approx 0.01464539 \cdot \frac{1}{100} = 0.0001464539$. We can also write $\frac{3}{20600} = \frac{3}{206 \cdot 100} = \frac{3}{206} \cdot \frac{1}{100}$. We can divide 3 by 206 to get $0.01464539...$. Then we multiply by $\frac{1}{100}$ to get $0.0001464539...$. We can write $\frac{3}{20600} = 0.0001464539...$ The decimal representation is $0.0001464539...$. We want to express this as a terminating decimal. We have $S = \frac{3}{20600}$. Dividing 3 by 20600, we get $3 \div 20600 = 0.0001464539...$ Since we want a terminating decimal, we need to find a common denominator. We have $20600 = 2^3 \cdot 5^2 \cdot 103$. So $\frac{3}{20600} = \frac{3}{2^3 \cdot 5^2 \cdot 103}$. We can write $\frac{3}{20600} = 0.0001464539...$ The decimal representation is $0.0001464539...$. Final Answer: The final answer is \color{red}{$\boxed{0.000146}$}\color{black}
        \Sepline
        \small
        \textbf{1) In-place edit}\\
Let $S = \frac{3}{206} + \frac{9}{2 \cdot 103^2} + \frac{27}{2 \cdot 103^3} + \cdots$. We can rewrite this as $S = \color{chart_blue}{ \sum_{n=1}^{\infty} \frac{3^n}{2 \cdot 103^{n}} = \frac{1}{2} \sum_{n=1}^{\infty}\left(\frac{3}{103}\right)^n}$
        \Sepline
        \small
        \textbf{2) Continuation generation}\\    
This is a geometric series with first term $a = \frac{3}{103}$ and common ratio $r = \frac{3}{103}$. Since $|r| = \frac{3}{103} < 1$, the series converges and its sum is given by $$ \sum_{n=1}^{\infty} \left(\frac{3}{103}\right)^n = \frac{\frac{3}{103}}{1 - \frac{3}{103}} = \frac{\frac{3}{103}}{\frac{103-3}{103}} = \frac{\frac{3}{103}}{\frac{100}{103}} = \frac{3}{100} $$ Therefore, $ S = \frac{1}{2} \cdot \frac{3}{100} = \frac{3}{200} $ To express this as a terminating decimal, we divide 3 by 200: $ \frac{3}{200} = \frac{3}{2 \cdot 100} = \frac{3}{2 \cdot 10^2} = \frac{3}{2} \cdot \frac{1}{100} = 1.5 \cdot 0.01 = 0.015 $ Alternatively, we can directly divide 3 by 200: $ \frac{3}{200} = 0.015 $ Thus, the infinite series is equal to \color{chart_blue}{\boxed{0.015}}
    \end{inplace}
    \caption{In-place feedback example for the MATH-hard dataset.}
    \label{fig:inplace_math1}
\end{figure*}
%%%%%%%%%%%%%%%%%%%%%%%%%%%%%%%%%%%%%%%%%%%%%%%%%%%%%%%%%%%%%%%%

%%%%%%%%%%%%In-place feedback example - MATH-hard%%%%%%%%%%%%%%%
\begin{figure*}[h]
    \centering
    \begin{inplace}[In-place feedback example - MATH-hard]
    \small
        \textbf{Problem} \\
Dr. Zaius invests $\$10,000$ in a CD with an annual interest rate of $4\%$ that compounds semi-annually (twice a year). After six months, he rolls over the CD into another CD with an annual interest rate of $5\%$ that also compounds semi-annually. After six months in the second CD, how much does Dr. Zaius have, in dollars?
        \Sepline
        \textbf{Model response}\\
First, we need to calculate the amount after the first six months with the first CD. The principal is $10000$, the annual interest rate is $4\%$, and the interest compounds semi-annually. So the interest rate per period is $4\% / 2 = 2\% = 0.02$. The number of periods is \color{red}{$6/2 = 3$}. \color{black}The formula for compound interest is $A = P(1 + r)^n$, where $A$ is the amount after $n$ periods, $P$ is the principal, and $r$ is the interest rate per period. In this case, $P = 10000$, $r = 0.02$, and $n = 3$. So, $A = 10000(1 + 0.02)^3 = 10000(1.02)^3 = 10000(1.061208) = 10612.08$. After six months, Dr. Zaius rolls over the CD into another CD with an annual interest rate of $5\%$ that also compounds semi-annually. The principal is $10612.08$. The annual interest rate is $5\%$, so the interest rate per period is $5\% / 2 = 2.5\% = 0.025$. The number of periods is $6/2 = 3$. Using the same formula, $A = P(1 + r)^n$, we have $P = 10612.08$, $r = 0.025$, and $n = 3$. So, $A = 10612.08(1 + 0.025)^3 = 10612.08(1.025)^3 = 10612.08(1.076890625) \approx 11418.53$. Therefore, after six months in the second CD, Dr. Zaius has approximately \color{red}\boxed{11418.53}\color{black}
        \Sepline
        \small
        \textbf{1) In-place edit}\\
First, we need to calculate the amount after the first six months with the first CD. The principal is $10000$, the annual interest rate is $4\%$, and the interest compounds semi-annually. So the interest rate per period is $4\% / 2 = 2\% = 0.02$. The number of periods is \color{chart_blue}$1$.\color{black}
        \Sepline
        \small
        \textbf{2) Continuation generation}\\    
The amount after six months is $P(1 + r)^n = 10000(1 + 0.02)^1 = 10000(1.02) = 10200$. Then, Dr. Zaius rolls over the CD into another CD with an annual interest rate of $5\%$ that also compounds semi-annually. The principal is $10200$, the annual interest rate is $5\%$, and the interest rate per period is $5\% / 2 = 2.5\% = 0.025$. The number of periods is $1$. The amount after six months in the second CD is $10200(1 + 0.025)^1 = 10200(1.025) = 10200 + 10200 \times 0.025 = 10200 + 255 = 10455$. So, after six months in the second CD, Dr. Zaius has \color{chart_blue}\boxed{10455}\color{black}
    \end{inplace}
    \caption{In-place feedback example for the MATH-hard dataset.}
    \label{fig:inplace_math2}
\end{figure*}
%%%%%%%%%%%%%%%%%%%%%%%%%%%%%%%%%%%%%%%%%%%%%%%%%%%%%%%%%

%%%%%%%%%%%%In-place feedback example - GPQA %%%%%%%%%%%%%%%
\begin{figure*}[h]
    \centering
    \begin{inplace}[In-place feedback example - GPQA free-form]
    \small
        \textbf{Problem} \\
Consider an isolated system of 13 identical particles each with charge 2e and negligible mass. 12 of these charges are constrained to stay at 2 m from a point P. The 13th charge is fixed at P. What is the minimum energy (in Joules) of this system correct to three decimals?
        \Sepline
        \textbf{Model response}\\
Let $q$ be the charge of each particle, and $r$ be the distance of the 12 particles from the point P. The 13th charge is at the point P.
The Coulomb force between two charges $q_1$ and $q_2$ separated by a distance $r$ is given by $F = \frac{kq_1q_2}{r^2}$, where $k = \frac{1}{4\pi\epsilon_0}$ is the Coulomb constant. \textit{[omitted for brevity...]}\\

The potential energy of the system is given by
$$\color{red} U = \sum_{i=1}^{12} \frac{k q^2}{r_i} $$
where $r_i$ is the distance from the $i$-th charge to the point P. Since all 12 charges are at a distance of 2 m from P, $r_i = 2$ for all $i$.
$ U = 12 \cdot \frac{k q^2}{2} = 6kq^2 $ \textit{[omitted for brevity...]}\\

The total energy is $U = 6kq^2$.
$U = 6 \cdot \frac{1}{4\pi\epsilon_0} (2e)^2 = \frac{6 \cdot 4e^2}{4\pi\epsilon_0} = \frac{6e^2}{\pi\epsilon_0} \approx \frac{6(1.602 \times 10^{-19})^2}{\pi(8.854 \times 10^{-12})} \approx \frac{6(2.5664 \times 10^{-38})}{\pi(8.854 \times 10^{-12})} \approx \frac{15.40 \times 10^{-38}}{2.783 \times 10^{-11}} \approx 5.57 \times 10^{-27} J$.
The minimum energy is approximately \color{red}\boxed{5.570e-27 $J$}\color{black}

        \Sepline
        \small
        \textbf{1) In-place edit}\\
Let $q$ be the charge of each particle, and $r$ be the distance of the 12 particles from the point P. The 13th charge is at the point P.
The Coulomb force between two charges $q_1$ and $q_2$ separated by a distance $r$ is given by $F = \frac{kq_1q_2}{r^2}$, where $k = \frac{1}{4\pi\epsilon_0}$ is the Coulomb constant. \textit{[omitted for brevity...]}\\
The potential energy of the system is given by $$\color{chart_blue} U = \sum_{i<j} \frac{k q_i q_j}{r_{ij}} $$ 

        \Sepline
        \small
        \textbf{2) Continuation generation}\\    
which includes the 12 central--outer terms and the 66 outer--outer pairwise terms; the outer--outer distances $r_{ij}$ depend on the arrangement of the 12 charges on the sphere of radius 2, so one must evaluate this pairwise sum for the energy-minimizing configuration before computing any numeric value. \textit{[omitted for brevity...]}\\ The minimum energy is $U = \frac{13}{2} \frac{k q^2}{r} = \frac{13}{2} \frac{\frac{1}{4\pi\epsilon_0} (2e)^2}{2} = \frac{13 e^2}{4\pi\epsilon_0} \approx \frac{13(1.602 \times 10^{-19})^2}{4\pi(8.854 \times 10^{-12})} \approx 2.82 \times 10^{-26} J$.
\color{chart_blue}\boxed{2.820e-26 $J$}\color{black}
    \end{inplace}
    \caption{In-place feedback example for the GPQA free-form dataset.}
    \label{fig:inplace_gpqa}
\end{figure*}

%%%%%%%%%%%%%%%%%%%%%%%%%%%%%%%%%%%%%%%%%%%%%%%%%%%%%%%%%

%%%%%%%%%%%%In-place feedback example - MMLU-pro freeform %%%%%%%%%%%%
\begin{figure*}[h]
    \centering
    \begin{inplace}[In-place feedback example - MMLU-pro free-form]
    \small
        \textbf{Problem} \\
A particle of mass $m$ and velocity $u_1$ makes a head-on collision with another particle of mass $2 m$ at rest. If the coefficient of restitution is such to make the loss of total kinetic energy a maximum, what are the velocities $v_1$ after the collision?
        \Sepline
        \textbf{Model response}\\
Let $m_1 = m$ and $v_1 = u_1$ be the mass and initial velocity of the first particle. Let $m_2 = 2m$ and $v_2 = 0$ be the mass and initial velocity of the second particle. Let $v_1'$ and $v_2'$ be the velocities of the first and second particles after the collision, respectively.
The initial total kinetic energy is $K_i = \frac{1}{2} m_1 u_1^2 + \frac{1}{2} m_2 (0)^2 = \frac{1}{2} m u_1^2$.
The final total kinetic energy is \color{red}$K_f = \frac{1}{2} m_1 v_1' + \frac{1}{2} m_2 v_2'$.\color{black}
The loss of kinetic energy is $\Delta K = K_i - K_f = \frac{1}{2} m u_1^2 - \frac{1}{2} m v_1' - \frac{1}{2} (2m) v_2'$. \textit{[omitted for brevity...]}\\

To maximize $\Delta K$, we need to minimize $v_2'$.
Since $v_2' = \frac{1}{2} v_1'$, we have $v_2' = \frac{1}{2} v_1'$.
Then $v_1' = 2v_2'$.
$\Delta K = \frac{1}{2} m u_1^2 - 2m \frac{1}{2} v_1' = \frac{1}{2} m u_1^2 - m v_1'$.
Final Answer: The final answer is \color{red}$\boxed{0}$\color{black}
        \Sepline
        \small
        \textbf{1) In-place edit}\\
Let $m_1 = m$ and $v_1 = u_1$ be the mass and initial velocity of the first particle. Let $m_2 = 2m$ and $v_2 = 0$ be the mass and initial velocity of the second particle. Let $v_1'$ and $v_2'$ be the velocities of the first and second particles after the collision, respectively.
The initial total kinetic energy is $K_i = \frac{1}{2} m_1 u_1^2 + \frac{1}{2} m_2 (0)^2 = \frac{1}{2} m u_1^2$.
The final total kinetic energy is \color{chart_blue}$K_f = \frac{1}{2} m_1 v_1'^2 + \frac{1}{2} m_2 v_2'^2$.\color{black}

        \Sepline
        \small
        \textbf{2) Continuation generation}\\    
The loss of kinetic energy is $\Delta K = K_i - K_f = \frac{1}{2} m u_1^2 - \frac{1}{2} m_1 v_1'^2 - \frac{1}{2} m_2 v_2'^2$.
We want to maximize $\Delta K$.
The coefficient of restitution is $e = \frac{v_2' - v_1'}{u_1 - v_2'}$.
Since it's a head-on collision, we can use the conservation of momentum:
$m_1 u_1 + m_2 (0) = m_1 v_1' + m_2 v_2'$
$m u_1 = m v_1' + 2m v_2'$
$u_1 = v_1' + 2 v_2'$
$v_2' = \frac{u_1 - v_1'}{2}$ \textit{[omitted for brevity...]}\\

To maximize $\Delta K$, we take the derivative with respect to $v_1'$ and set it to zero. However, we are given that the loss of kinetic energy is maximized. This occurs when $\Delta K$ is maximized.
For maximum loss of kinetic energy, we want $e$ to be as small as possible, i.e., $e = 0$.
If $e = 0$, then $v_1' = \frac{u_1(1-0)}{0+3} = \frac{u_1}{3}$.
Then $v_2' = \frac{u_1 - v_1'}{2} = \frac{u_1 - \frac{u_1}{3}}{2} = \frac{\frac{2u_1}{3}}{2} = \frac{u_1}{3}$.
\color{chart_blue}\boxed{v_1' = \frac{u_1}{3}}\color{black}
    \end{inplace}
    \caption{In-place feedback example for the MMLU-pro free-form dataset.}
    \label{fig:inplace_mmlu}
\end{figure*}

%%%%%%%%%%%%%%%%%%%%%%%%%%%%%%%%%%%%%%%%%%%%%%%%%%%%%%%%%

%%%%%%%%%%%% Multi-turn failure - corrupt correct content %%%%%%%%%%%%%%%
\begin{figure*}[h]
    \centering
    \begin{failure}[Failure example of multi-turn refinement - 1) corrupting correct content]
    \small
        \textbf{Problem} \\
In convex hexagon $ABCDEF$, all six sides are congruent, \(\angle A\) and \(\angle D\) are right angles, and \(\angle B\), \(\angle C\), \(\angle E\), and \(\angle F\) are congruent. The area of the hexagonal region is $2116(\sqrt2+1)$. Find $AB$.
        \Sepline
        \textbf{Model response}\\
To solve for \( AB \) in the given convex hexagon \( ABCDEF \), we will follow these steps:\\

1. \textbf{Identify the properties of the hexagon:}\\
   - All six sides are congruent.\\
   - \(\angle A\) and \(\angle D\) are right angles.\\
   - \(\angle B\), \(\angle C\), \(\angle E\), and \(\angle F\) are congruent.\\

2. \textbf{Determine the measure of the congruent angles:}\\
   - Since the sum of the interior angles of a hexagon is \( (6-2) \times 180^\circ = 720^\circ \).\\
   - Let \(\angle B = \angle C = \angle E = \angle F = x\).\\
   - We have \(\angle A = 90^\circ\) and \(\angle D = 90^\circ\).\\
   - Therefore, \(4x + 180^\circ = 720^\circ\).\\
   - Solving for \(x\): $4x = 540^\circ \implies x = 135^\circ$\\

3. \textbf{Visualize the hexagon:}
   - Place the hexagon on a coordinate plane with \(A\) at \((0, 0)\) and \(D\) at \((a, 0)\).\\
   - Since \(\angle A = 90^\circ\) and \(\angle D = 90^\circ\), the hexagon can be divided into a rectangle and two right triangles.\\

4. \textbf{Divide the hexagon into simpler shapes:}\\
   - The hexagon can be divided into a rectangle \(ABCD\) and two right triangles \(ABE\) and \(CDF\).\\

Since the problem asks for \(AB\), and we have determined that \(AB = s\), the final answer is: \color{red}\boxed{23}\color{black}

        \Sepline
        \small
        \textbf{Feedback}\\
 \color{chart_blue}  Your output is wrong because you assumed without justification that the hexagon \textbf{``can be divided into a rectangle and two right triangles" and then treated \(ABCD\) as a rectangle (even as a square) with both width and height equal to $s$; right angles at $A$ and $D$ alone do not imply $AB$ $ \parallel $ $CD$ or $BC$ $ \parallel $ $AD$, so $ABCD$ need not be a rectangle and the triangle legs need not equal $s$.} \\

Fix it by producing a correct partition justified by the given angles (for example, draw and justify specific diagonals such as $BF$ and $CE$), use the 135° measure of the other four angles to deduce the shapes and side relationships of the pieces, and compute the area from that valid decomposition rather than assuming a rectangle/square. 
 \color{black}
        \Sepline
        \small
        \textbf{Model response}\\    
 Let's re-evaluate the problem step-by-step, ensuring we correctly decompose the hexagon based on the given angles and properties.\\

1. \textbf{Identify the properties of the hexagon:}\\
   - All six sides are congruent.\\
   - \(\angle A = 90^\circ\) and \(\angle D = 90^\circ\).\\
   - \(\angle B = \angle C = \angle E = \angle F = 135^\circ\).\\

2. \textbf{Visualize the hexagon:}\\
   - Place the hexagon on a coordinate plane with \(A\) at \((0, 0)\) and \color{red}\(D\) at \((s, 0)\).\color{black}\\

3. \textbf{Draw and justify specific diagonals:}\\
   - Draw diagonal \(BF\) and \(CE\). These diagonals will help us decompose the hexagon into simpler shapes.\\

4. \textbf{Analyze the resulting shapes:}\\
   - Diagonal \(BF\) splits \(\angle B\) and \(\angle F\) into two angles each: \(\angle ABF = \angle FBD = 45^\circ\).\\
   - Similarly, diagonal \(CE\) splits \(\angle C\) and \(\angle E\) into two angles each: \(\angle BCE = \angle ECD = 45^\circ\).\\

Since the problem asks for \(AB\), and we have determined that \(AB = s\), the final answer is: \color{red}\boxed{23}\color{black}

    \end{failure}
    \caption{Failure example of multi-turn refinement on the MATH-hard dataset: corrupting correct content.}
    \label{fig:fail_case1}
\end{figure*}

%%%%%%%%%%%%%%%%%%%%%%%%%%%%%%%%%%%%%%%%%%%%%%%%%%%%%%%%%

%%%%%%%%%%%% Multi-turn failure - ignore feedback %%%%%%%%%%%%%%%
\begin{figure*}[h]
    \centering
    \begin{failure}[Failure example of multi-turn refinement - 2) ignoring feedback]
    \small
        \textbf{Problem} \\
A container weighs 3.22 lb force when empty. Filled with waterat 60°F the mass of the container and its contents is 1.95 slugs. Find the weight of the water in the container and its volume in cubic feet. Assume density of water= 62.4 lb force/$\text{ft}^3$.
        \Sepline
        \textbf{Model response}\\
Let $W_c$ be the weight of the empty container in lb force, which is given as $W_c = 3.22$ lb.
Let $m$ be the mass of the container and its contents in slugs, which is given as $m = 1.95$ slugs.
Let $\rho$ be the density of water in lb force/ft$^3$, which is given as $\rho = 62.4$ lb/ft$^3$.
We want to find the weight of the water in the container, $W_w$, and the volume of the water in the container, $V$ in cubic feet.\\

First, we find the weight of the container and its contents due to gravity. We know that weight $W = mg$, where $m$ is mass and $g$ is the acceleration due to gravity. We can use the given mass in slugs and the standard value of $g = 32.2$ ft/s$^2$.
\color{red}$W = m \cdot g = 1.95 \text{ slugs} \cdot 32.2 \text{ ft/s}^2 = 62.19 \text{ lb}$.\color{black}\\

The weight of the container and its contents is also the sum of the weight of the empty container and the weight of the water.
$W_c + W_w = 62.19$ lb
$3.22 + W_w = 62.19$
$W_w = 62.19 - 3.22 = 58.97$ lb.\\

Now we can find the volume of the water in the container using the density formula: $\rho = \frac{W}{V}$, where $W$ is the weight of the water and $V$ is the volume.
$V = \frac{W}{\rho} = \frac{58.97 \text{ lb}}{62.4 \text{ lb/ft}^3} \approx 0.9455 \text{ ft}^3$.\\

Therefore, the weight of the water is 58.97 lb and the volume of the water is approximately 0.9455 ft$^3$.\\

\color{red}\boxed{\text{Weight of water} = 58.97 \text{ lb}, \text{Volume} = 0.9455 \text{ ft}^3}\color{black}

        \Sepline
        \small
        \textbf{Feedback}\\
 \color{chart_blue} \textbf{Your output is wrong because you miscomputed the total weight: you evaluated $W = m \cdot g$ incorrectly (you used 1.95·32.2 = 62.19)}, which led to an incorrect water weight. Recompute $W = m \cdot g$ with the correct product, then subtract the empty container weight to get the water weight and divide that by the given density to obtain the volume. \color{black}
        \Sepline
        \small
        \textbf{Model response}\\    
 Let $W_c$ be the weight of the empty container in lb force, which is given as $W_c = 3.22$ lb.
Let $m$ be the mass of the container and its contents in slugs, which is given as $m = 1.95$ slugs.
Let $\rho$ be the density of water in lb force/ft$^3$, which is given as $\rho = 62.4$ lb/ft$^3$.
We want to find the weight of the water in the container, $W_w$, and the volume of the water in the container, $V$ in cubic feet.\\

First, we find the weight of the container and its contents due to gravity. We know that weight $W = mg$, where $m$ is mass and $g$ is the acceleration due to gravity. We can use the given mass in slugs and the standard value of $g = 32.2$ ft/s$^2$.
\color{red}$W = m \cdot g = 1.95 \text{ slugs} \cdot 32.2 \text{ ft/s}^2 = 62.19 \text{ lb}$.\color{black}\\

The weight of the container and its contents is $62.19$ lb.
The weight of the empty container is $3.22$ lb.
The weight of the water is the difference between the total weight and the weight of the empty container:
$W_w = 62.19 - 3.22 = 58.97$ lb.\\

Now we can find the volume of the water in the container using the density formula: $\rho = \frac{W}{V}$, where $W$ is the weight of the water and $V$ is the volume.
$V = \frac{W}{\rho} = \frac{58.97 \text{ lb}}{62.4 \text{ lb/ft}^3} \approx 0.9455 \text{ ft}^3$.\\

Therefore, the weight of the water is 58.97 lb and the volume of the water is approximately 0.9455 ft$^3$.\\

\color{red}\boxed{\text{Weight of water} = 58.97 \text{ lb}, \text{Volume} = 0.9455 \text{ ft}^3}\color{black}
    \end{failure}
    \caption{Failure example of multi-turn refinement on the MMLU-pro free-form dataset: ignoring feedback.}
    \label{fig:fail_case2}
\end{figure*}

%%%%%%%%%%%%%%%%%%%%%%%%%%%%%%%%%%%%%%%%%%%%%%%%%%%%%%%%%

%%%%%%%%%%%% Multi-turn failure - ignore feedback %%%%%%%%%%%%%%%
\begin{figure*}[h]
    \centering
    \begin{failure}[Failure example of multi-turn refinement - 3) introducing a new error]
    \small
        \textbf{Problem} \\
A 125/250-volt three-wire system has a load on the positive side of 500 amp, on the negative side of 450 amp, the neutral current therefore being 50 amp. If each machine of the balancer set has an efficiency of 86 percent, calculate the current for each of these machines and for the main generator.

        \Sepline
        \textbf{Model response}\\
Here's a step-by-step solution to the problem:\\

1. \textbf{Calculate the total power in the positive phase:}\\
   $P_{pos} = V_{pos} \times I_{pos} = 125 V \times 500 A = 62500 W$

2. \textbf{Calculate the total power in the negative phase:}\\
   $P_{neg} = V_{neg} \times I_{neg} = 125 V \times 450 A = 56250 W$

3. \textbf{Calculate the total power in the neutral phase:}\\
   $P_{neutral} = V_{neutral} \times I_{neutral} = 125 V \times 50 A = 6250 W$

4. \textbf{Calculate the total system power:}\\
   $P_{total} = P_{pos} + P_{neg} + P_{neutral} = 62500 W + 56250 W + 6250 W = 125000 W$

5. \textbf{Determine the number of balancer sets:}\\
   The problem states each balancer set has an efficiency of 86\%, which implies that there are two balancer sets. \textit{[omitted for brevity...]}\\

The balancer sets are designed to correct the imbalance. The generator provides 1000A. The imbalance is 500A - 450A = 50A. The balancer sets must provide 50A.
Since each set has 86\% efficiency, the current each set provides is $I = \frac{50}{0.86} \approx 581.39 A$.
So, $I_1 = I_2 = 581.39 A$.

\color{red}\boxed{I_{generator} = 1000, I_{set1} = 581.39, I_{set2} = 581.39} \color{black}

        \Sepline
        \small
        \textbf{Feedback}\\
 \color{chart_blue}  Your output is wrong because \textbf{you infer there are two balancer sets from the 86\% efficiency — efficiency does not determine the number of machines, so splitting total power by two is an unjustified non sequitur that corrupts all subsequent calculations.} To fix it, do not assume a quantity: either use the actual number of machines stated in the problem or introduce a variable n, first compute the required balancing current from the load imbalance (or neutral current), then divide that required output among n machines and account for each machine’s 86\% efficiency by dividing the required output current/power by 0.86 to obtain the input/current for each machine. 
\color{black}
        \Sepline
        \small
        \textbf{Model response}\\    
  Okay, you are absolutely correct. My apologies for the flawed reasoning. Let’s redo the calculation with the correct approach.\\

1. \textbf{Calculate the total power}:\\
   $P_{total} = 62500 W + 56250 W + 6250 W = 125000 W$

2. \textbf{Calculate the imbalance:}\\
   The imbalance is the difference between the positive and negative currents: $500 A - 450 A = 50 A$.  This is the current the balancer sets need to correct.

\color{chart_blue}3. \textbf{Determine the number of balancer sets:}\\
The problem doesn't explicitly state the number of balancer sets. Let's denote the number of balancer sets as $n$. \color{black}

4. \textbf{Calculate the required current per balancer set:}\\
   Each balancer set needs to provide $50 A$.

5. \textbf{Calculate the current required from each balancer set, accounting for efficiency:}\\
   \color{red}Current per set = (Required current) / (Efficiency) = $50 A / 0.86 = 581.395 A$ (approximately)\color{black}

\color{red}\boxed{I_{generator} = 1000, I_{set1} = 58.14, I_{set2} = 58.14} \color{black}

    \end{failure}
    \caption{Failure example of multi-turn refinement on the MMLU-pro free-form dataset: introducing a new error.}
    \label{fig:fail_case3}
\end{figure*}

%%%%%%%%%%%%%%%%%%%%%%%%%%%%%%%%%%%%%%%%%%%%%%%%%%%%%%%%%

\section{User Study Interface}
\label{app.user_study_interface}

\begin{figure*}[h]
    \centering
    \includegraphics[width=.8\linewidth]{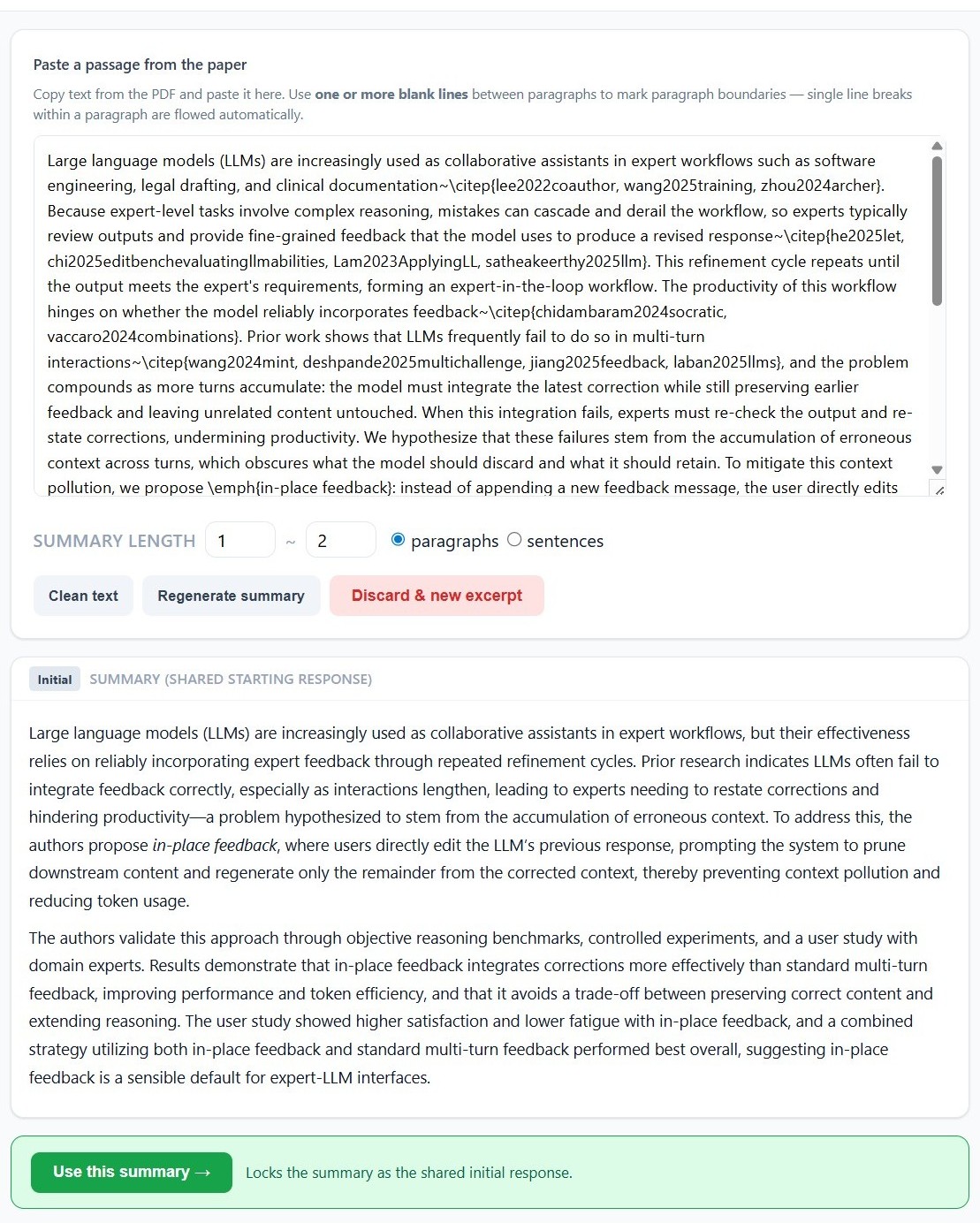}
    \caption{Screenshot of the summarization page. Participants paste a passage from their chosen paper and click \texttt{Summarize} to obtain an initial summary from the model.}
    \label{fig:user_study_summarization_page}
\end{figure*}

\begin{figure*}[h]
    \centering
    \includegraphics[width=.8\linewidth]{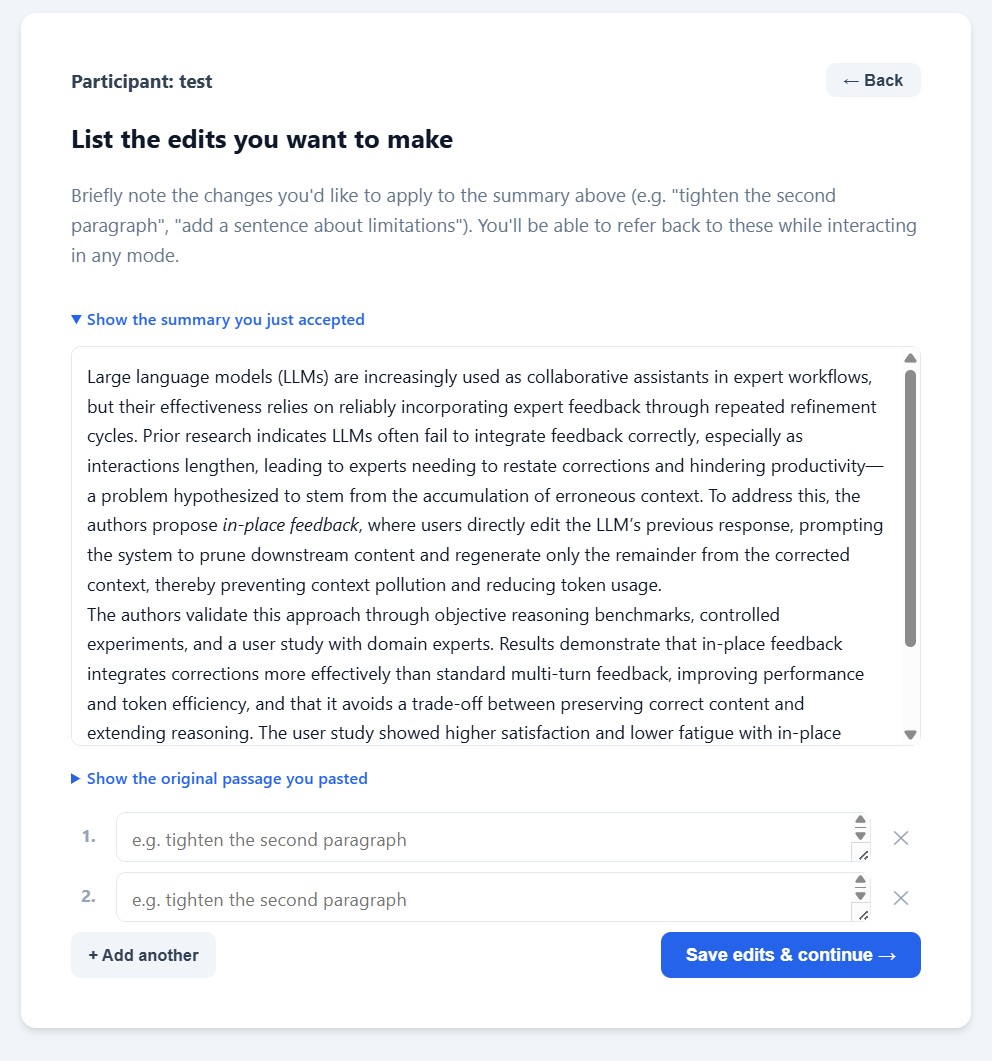}
    \caption{Screenshot of the edit decision page. Participants record the edits they plan to make. This serves as a memo to help them stay organized during the subsequent interaction.}
    \label{fig:user_study_edit_decision_page}
\end{figure*}

\begin{figure*}[h]
    \centering
    \includegraphics[width=.8\linewidth]{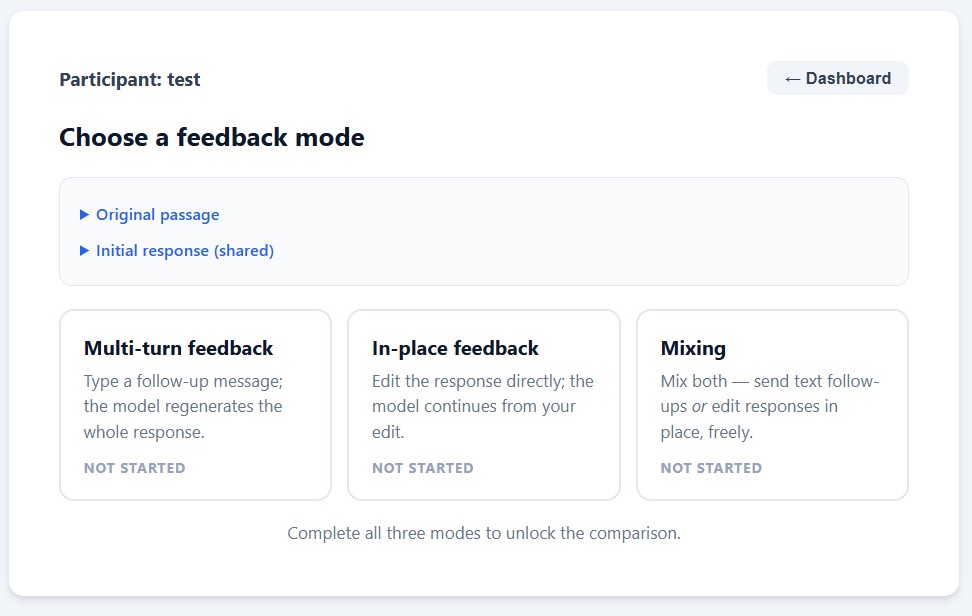}
    \caption{Screenshot of the mode selection page, where participants choose the editing mode for the upcoming task.}
    \label{fig:user_study_mode_selection_page}
\end{figure*}

\begin{figure*}[h]
    \centering
    \includegraphics[width=.75\linewidth]{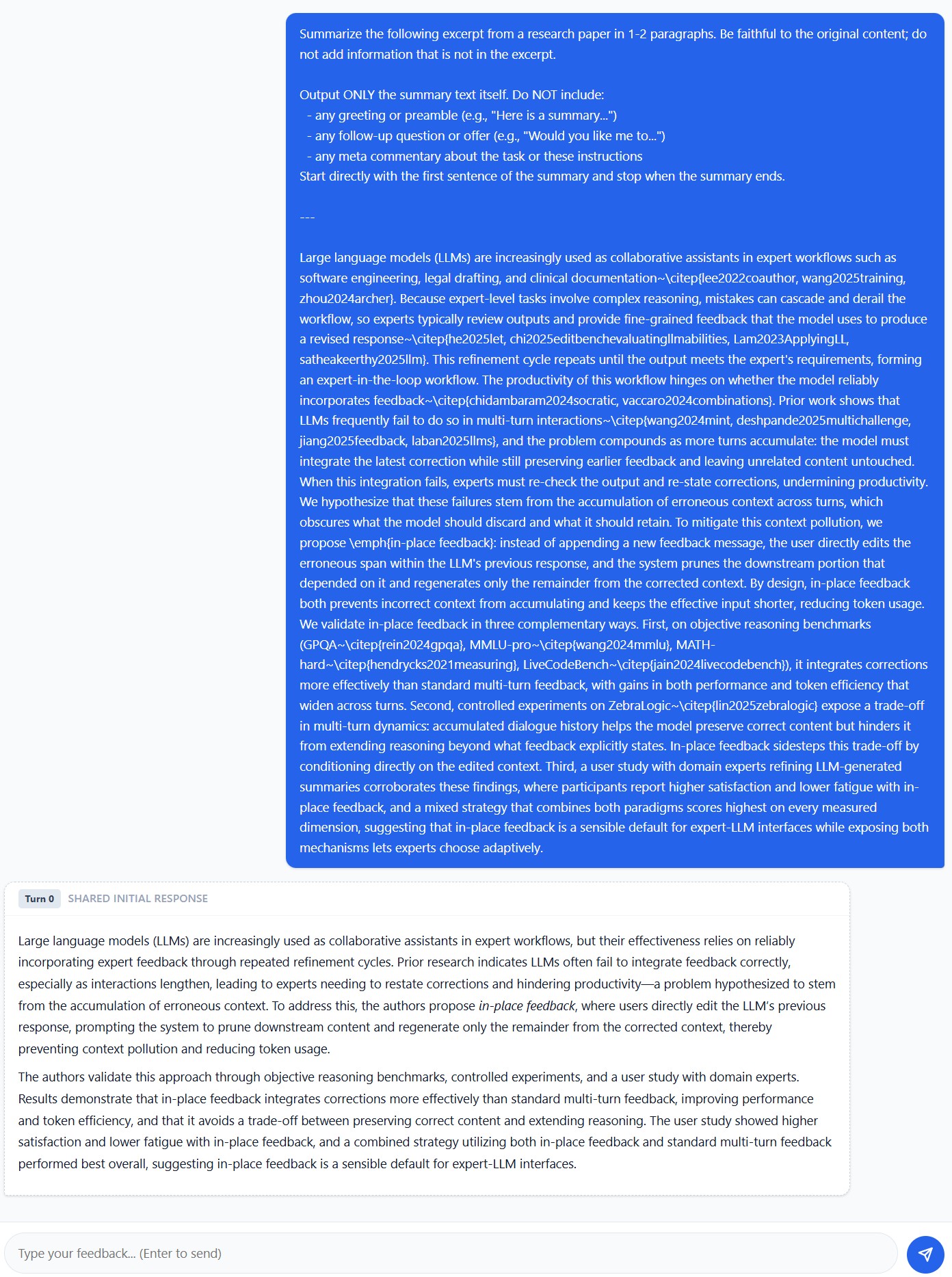}
    \caption{Screenshot of the multi-turn mode interface, used when participants provide feedback through standard multi-turn feedback.}
    \label{fig:user_study_multi_turn_page}
\end{figure*}

\begin{figure*}[h]
    \centering
    \includegraphics[width=.75\linewidth]{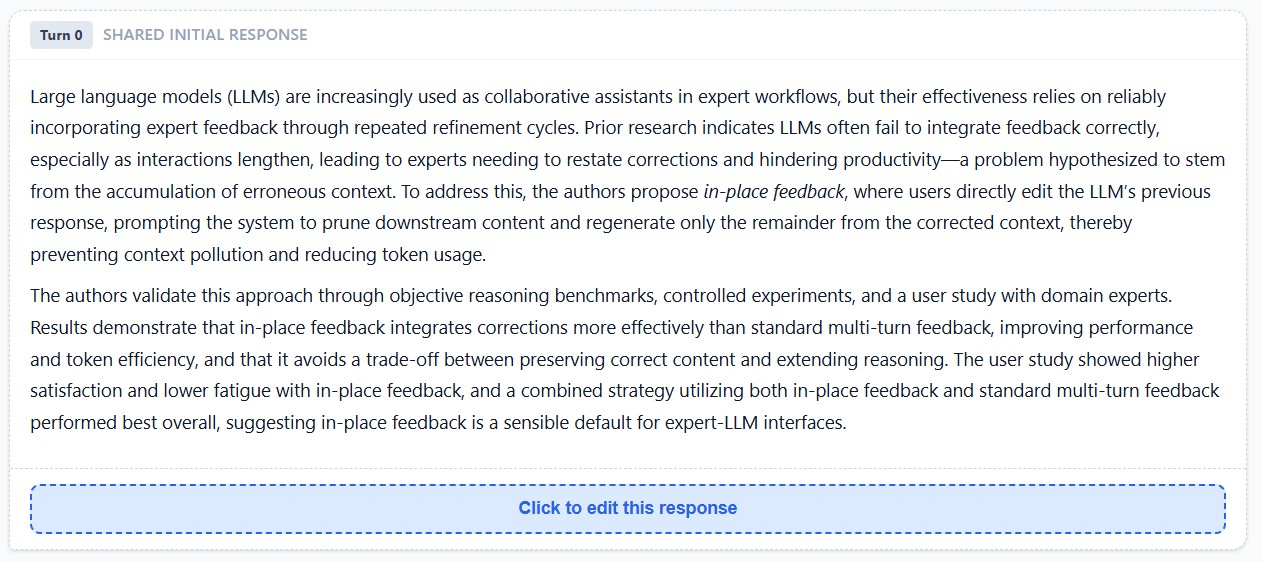}
    \caption{Screenshot of a model response in the in-place feedback mode. Clicking the button below the response opens an editor that allows participants to perform an in-place edit on the response.}
    \label{fig:user_study_in_place_msg}
\end{figure*}

\begin{figure*}[h]
    \centering
    \includegraphics[width=.8\linewidth]{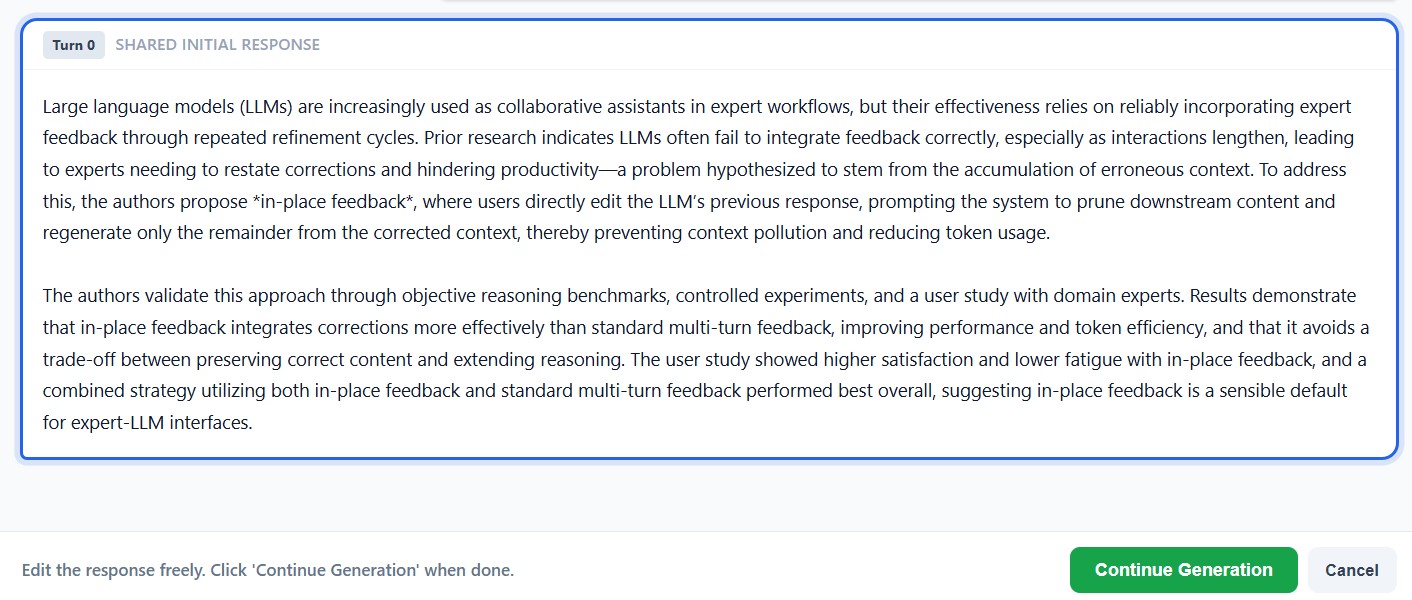}
    \caption{Screenshot of the in-place feedback interface. Participants directly modify the model's response and click \texttt{Continue Generation} to resume generation from the edited context.}
    \label{fig:user_study_in_place_msg_clicked}
\end{figure*}

\begin{figure*}[h]
    \centering
    \includegraphics[width=.8\linewidth]{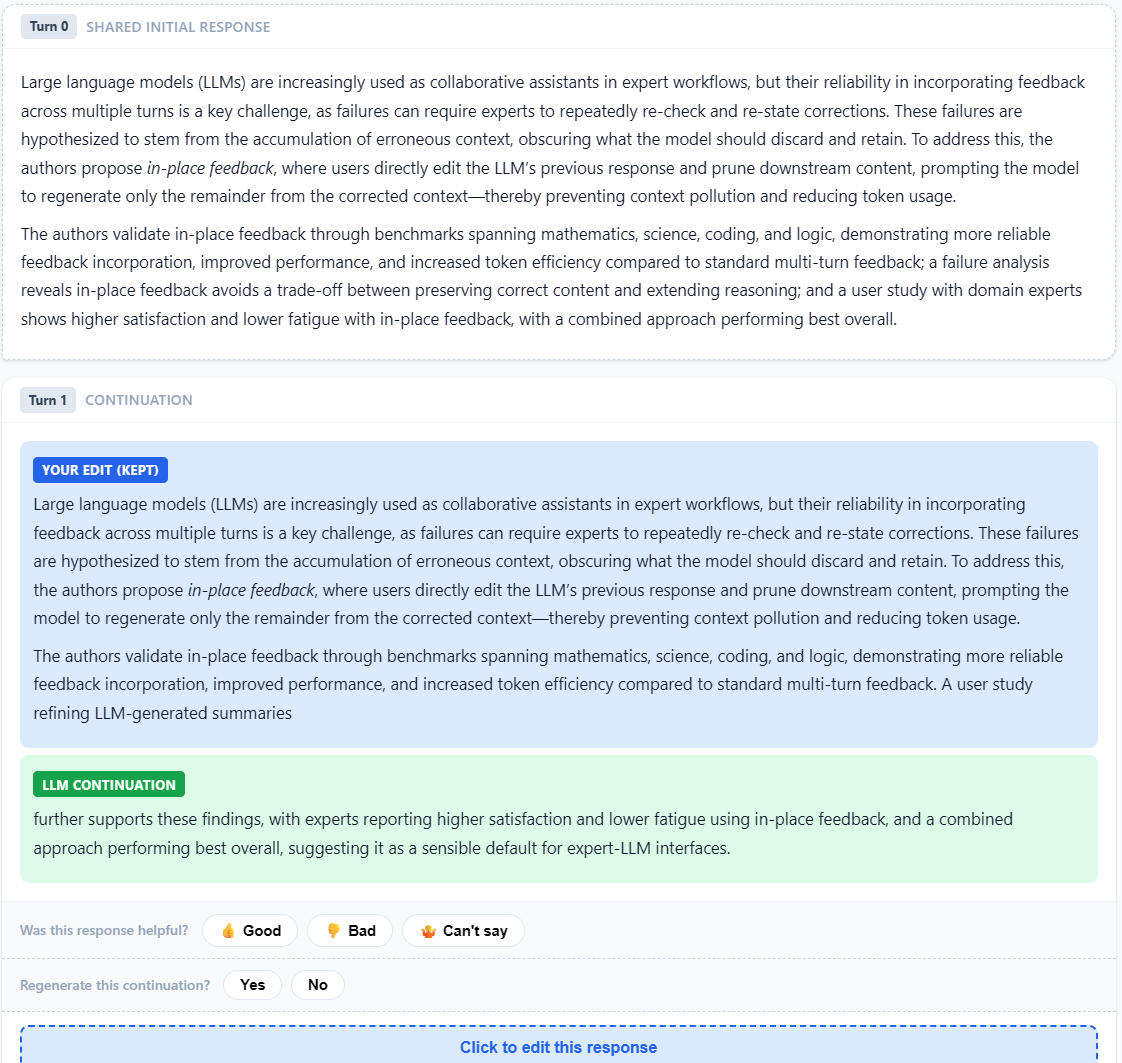}
    \caption{Screenshot of an example response produced under in-place feedback. The blue span indicates the participant's in-place edit, and the green span indicates the continuation subsequently generated by the model.}
    \label{fig:user_study_in_place_response}
\end{figure*}

\cref{fig:user_study_summarization_page,fig:user_study_edit_decision_page,fig:user_study_mode_selection_page,fig:user_study_multi_turn_page,fig:user_study_in_place_msg,fig:user_study_in_place_msg_clicked,fig:user_study_in_place_response} are examples of our user study interface.

% \begin{figure*}[h]
%     \centering
%     \includegraphics[width=.6\linewidth]{figures_user_study/after_edit_page.jpg}
%     \caption{Screenshot of the post-task evaluation page, consisting of rating items for the current mode and a text box for open-ended comments.}
%     \label{fig:user_study_after_edit_page}
% \end{figure*}

% \begin{figure*}[h]
%     \centering
%     \includegraphics[width=.6\linewidth]{figures_user_study/mode_comparison_page.jpg}
%     \caption{Screenshot of the mode comparison page, where participants compare the two interaction modes across multiple dimensions and provide open-ended feedback.}
%     \label{fig:user_study_mode_comparison_page}
% \end{figure*}